\documentclass{article}

\usepackage{times}
\usepackage[utf8]{inputenc} 
\usepackage[T1]{fontenc}    
\usepackage{url}            
\usepackage{amsfonts,amsmath,amssymb}       
\usepackage{nicefrac}       
\usepackage{multirow}
\usepackage{algorithm,algorithmic}
\usepackage[normalem]{ulem}
\usepackage{arydshln}
\usepackage{caption}
\usepackage{enumitem}
\usepackage{wrapfig}
\usepackage{microtype}
\usepackage{graphicx}
\usepackage{booktabs} 
\usepackage{cutwin}
\usepackage[pangram]{blindtext}
\usepackage[calc]{adjustbox}
\usepackage{subcaption}
\usepackage{color}
\usepackage[dvipsnames]{xcolor}

\usepackage{amsthm}

\definecolor{Red}{rgb}{0.6,0,0}
\definecolor{Blue}{rgb}{0,0,0.8}
\definecolor{Green}{rgb}{0,0.4,0.7}
\definecolor{airforceblue}{rgb}{0.36, 0.54, 0.66}
\definecolor{ao(english)}{rgb}{0.0, 0.5, 0.0}
\definecolor{azure(colorwheel)}{rgb}{0.0, 0.5, 1.0}
\definecolor{crimson}{rgb}{0.86, 0.08, 0.24}
\definecolor{darkcerulean}{rgb}{0.03, 0.27, 0.49}
\definecolor{cobalt}{rgb}{0.0, 0.28, 0.67}
\definecolor{rosegold}{rgb}{0.72, 0.43, 0.47}
\definecolor{orange-red}{rgb}{1.0, 0.27, 0.0}
\definecolor{mountainmeadow}{rgb}{0.19, 0.73, 0.56}
\definecolor{malachite}{rgb}{0.04, 0.85, 0.32}
\definecolor{darkblue}{rgb}{0.0, 0.0, 0.55}
 
 \definecolor{redfigure}{RGB}{240, 87, 87}
 \definecolor{bluefigure}{RGB}{64,105,225}
\definecolor{greenfigure}{RGB}{34,178,170}

\usepackage{hyperref}
\hypersetup{colorlinks=true}
\hypersetup{linktoc=all}
\hypersetup{citecolor=MidnightBlue}
\hypersetup{linkcolor=BrickRed}
\hypersetup{urlcolor=MidnightBlue}
\usepackage[all]{hypcap}
\usepackage[noabbrev, nameinlink, capitalize]{cleveref}
\newcommand{\eat}[1]{}

\newcommand{\lump}{\textsc{Lump }}

\newcommand*\rowlegend[{1}]{\rotatebox[origin=lt]{90}{#1}}%
\newcommand{\specialcell}[2][c]{\begin{tabular}[#1]{@{}c@{}}#2\end{tabular}}
\newcommand{\rb}[1]{\raisebox{1.7ex}[0pt]{#1}}
\newcommand{\bm}[1]{\boldsymbol{#1}} 

\usepackage{tikz}
\usetikzlibrary{positioning, shapes.geometric}

\renewcommand{\algorithmiccomment}[1]{\bgroup\hfill$\triangleright$~#1\egroup}

\usepackage{iclr_2022}
\iclrfinalcopy
\title{Representational Continuity for \\ Unsupervised Continual Learning}

\author{%
    Divyam Madaan$^{1}$\thanks{Corresponding author.~~~{$^\dagger$} The work was done while the author was an intern at Microsoft Research.}~~~
    Jaehong Yoon$^{2,3~\color{Red}{\dagger}}$~~
    Yuanchun Li$^{5,6}$~~
    Yunxin Liu$^{5,6}$~~
    Sung Ju Hwang$^{2,4}$\\
	New York University$^{1}$\hspace{0.15in}
	KAIST$^{2}$\hspace{0.15in}
	Microsoft Research$^{3}$\hspace{0.15in}
	AITRICS$^{4}$\\
	Institute for AI Industry Research (AIR)$^{5}$\hspace{0.15in}
	Tsinghua University$^{6}$\\
	\texttt{divyam.madaan@nyu.edu, \{jaehong.yoon,sjhwang82\}@kaist.ac.kr}\\
	\texttt{liyuanchun@air.tsinghua.edu.cn, liuyunxin@air.tsinghua.edu.cn}
}
\begin{document}
\maketitle

\begin{abstract}
Continual learning (CL) aims to learn a sequence of tasks without forgetting the previously acquired knowledge. However, recent CL advances are restricted to supervised continual learning (SCL) scenarios. Consequently, they are not scalable to real-world applications where the data distribution is often biased and unannotated. In this work, we focus on \emph{unsupervised continual learning (UCL)}, where we learn the feature representations on an unlabelled sequence of tasks and show that reliance on annotated data is not necessary for continual learning. We conduct a systematic study analyzing the learned feature representations and show that unsupervised visual representations are surprisingly more robust to catastrophic forgetting, consistently achieve better performance, and generalize better to out-of-distribution tasks than SCL. Furthermore, we find that UCL achieves a smoother loss landscape through qualitative analysis of the learned representations and learns meaningful feature representations. Additionally, we propose {\bf L}ifelong {\bf U}nsupervised {\bf M}ixu{\bf p} (\textsc{Lump}), a simple yet effective technique that interpolates between the current task and previous tasks' instances to alleviate catastrophic forgetting for unsupervised representations. We release our code \href{https://github.com/divyam3897/UCL}{online}.
\end{abstract}
\section{Introduction}
Recently continual learning~\citep{ThrunS1995} has gained a lot of attention in the deep learning community due to its ability to continually learn on a sequence of non-stationary tasks~\citep{KumarA2012icml,LiZ2016eccv} and close proximity to the human learning process~\citep{Flesch18comparing}. However, the inability of the learner to prevent forgetting of the knowledge learnt from the previous tasks has been a long-standing problem~\citep{mccloskey1989catastrophic, goodfellow2013empirical}. To address this problem, a large body of methods~\citep{rusu2016progressive, zenke17si, YoonJ2018iclr, li2019learn, Aljundi2019GradientBS, buzzega2020dark} have been proposed; however, all these methods focus on the supervised learning paradigm, but obtaining high-quality labels is expensive and often impractical in real-world scenarios. In contrast, CL for unsupervised representation learning has received limited attention in the community. Although~\cite{rao19curl} instantiated a continual unsupervised representation learning framework (\textsc{Curl}), it is not scalable for high-resolution tasks, as it is composed of MLP encoders/decoders and a simple MoG generative replay. This is evident in their limited empirical evaluation using digit-based gray-scale datasets. 

Meanwhile, a set of directions have shown huge potential to tackle the representation learning problem without labels~\citep{he2019moco, chen20simclr, grill20byol, chen2020big, chen2020exploring, zbontar2021barlow} by aligning contrastive pairs of training instances or maximizing the similarity between two augmented views of each image.  However, a common assumption for existing methods is the availability of a large amount of unbiased and unlabelled datasets to learn the feature representations. We argue that this assumption is not realistic for most of the real-time applications, including self-driving cars~\citep{DBLP:journals/corr/BojarskiTDFFGJM16}, medical applications~\citep{kelly2019key} and conversational agents~\citep{li2020compositional}. The collected datasets are often limited in size during the initial training phase~\citep{finn2017model}, and datasets/tasks change continuously with time. 

To accommodate such continuous shifts in data distributions, representation learning models need to increment the knowledge without losing the representations learned in the past.
\begin{figure*}[t!]
\centering
\vspace{-0.3in}
\resizebox{\linewidth}{!}{%
\begin{tabular}{cc}
     \hspace{-0.2in}
     \includegraphics[height=3.5cm]{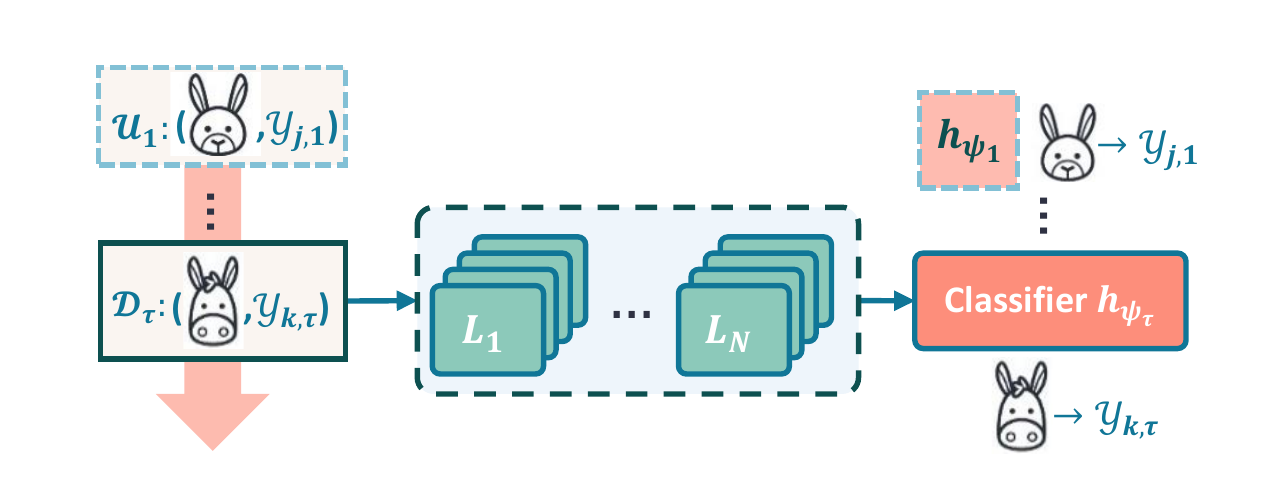}&
     \includegraphics[height=3.5cm]{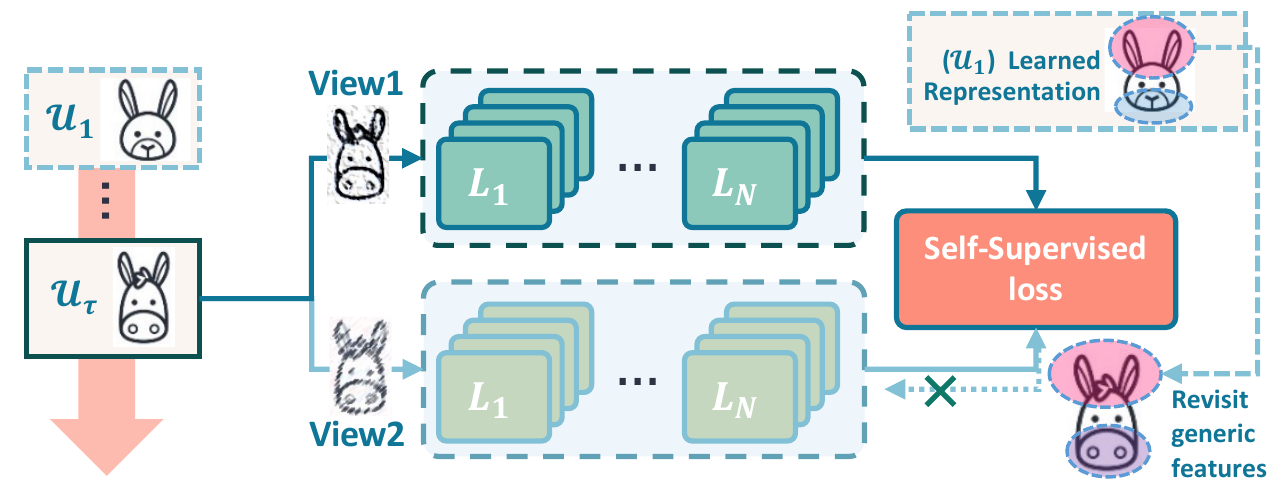}\\
     \hspace{-0.2in}{\footnotesize \textsc{Supervised Continual Learning (SCL)}} &{\footnotesize \textsc{Unsupervised Continual Learning (UCL)}} \\
\end{tabular}}
\vspace{-0.1in}
\caption{\footnotesize \textbf{Illustration of supervised and unsupervised continual learning.} The objective of SCL is to learn the ability to classify labeled images in the current task while preserving the past tasks' knowledge, where the tasks are non-iid to each other. On the other hand, UCL aims to learn the representation of images without the presence of labels and the model learns general-purpose representations during sequential training.}
\label{fig:concept}
\vspace{-0.2in}
\end{figure*}
With this motivation, we attempt to bridge the gap between unsupervised representation learning and continual learning to address the challenge of learning the representations on a sequence of tasks. Specifically, we focus on \emph{unsupervised continual learning (UCL)}, where the goal of the continual learner is to learn the representations from a stream of unlabelled data instances without forgetting (see \cref{fig:concept}). To this end, we extend various existing SCL strategies to the unsupervised continual learning framework and analyze the performance of current state-of-the-art representation learning techniques: \emph{SimSiam}~\citep{chen2020exploring} and \emph{BarlowTwins}~\citep{zbontar2021barlow} for UCL. Surprisingly, we observe that the unsupervised representations are comparatively more robust to catastrophic forgetting across all datasets and simply finetuning on the sequence of tasks can outperform various state-of-the-art continual learning alternatives. Furthermore, we show that UCL generalize better to various out of distribution tasks and outperforms SCL for few-shot training scenarios (\cref{sec:quantitative}).

We demystify the robustness of unsupervised representations by investigating the feature similarity, measured by centered kernel alignment (CKA)~\citep{kornblith2019similarity} between two independent UCL and SCL models and between UCL and SCL models. We notice that two unsupervised model representations have a relatively high feature similarity compared to two supervised representations. Furthermore, in all cases, two models have high similarity in lower layers indicating that they learn similar low-level features.
Further, we measure the $\ell_2$ distance between model parameters~\citep{neyshabur2021transferred} and visually compare the feature representations learned by different SCL and UCL strategies. We observe that UCL obtains human perceptual feature patterns for previous tasks, demonstrating their effectiveness to alleviate catastrophic forgetting (\cref{sec:qualitative}). We conjecture that this is due to their characteristic ability to learn general-purpose features~\citep{doersch2021crosstransformers}, which makes them transfer better and comparatively more robust to catastrophic forgetting.

To gain further insights, we visualize the loss landscape~\citep{li2017visualizing} of the UCL and SCL models and observe that UCL obtains a flatter and smoother loss landscape than SCL.
Additionally, we propose a simple yet effective technique coined {\bf L}ifelong {\bf U}nsupervised {\bf M}ixu{\bf p} (\textsc{Lump}), which utilizes mixup~\citep{zhang2017mixup} for unlabelled training instances. In particular, \lump interpolates between the current task examples and examples from previous instances to minimize catastrophic forgetting. We emphasize that \lump is easy to implement, does not require additional hyperparameters, and simply trains on the interpolated instances. To this end, \lump requires little, or no modification to existing rehearsal-based methods effectively minimizes catastrophic forgetting even with uniformly selecting the examples from replay buffer. We show that \lump with UCL outperforms the state-of-the-art supervised continual learning methods across multiple experimental settings with significantly lower catastrophic forgetting. In summary, our contributions are as follows:

\begin{itemize}
    \item We attempt to bridge the gap between continual learning and representation learning and tackle the two crucial problems of continual learning with unlabelled data and representation learning on a sequence of tasks.
    \item Systematic quantitative analysis shows that UCL achieves better performance over SCL with significantly lower catastrophic forgetting on Sequential CIFAR-10, CIFAR-100, and Tiny-ImageNet. Additionally, we evaluate out-of-distribution tasks and few-shot training demonstrating the expressive power of unsupervised representations. 
    \item We provide visualization of the representations and loss landscapes, which show that UCL learns discriminative, human perceptual patterns and achieves a flatter and smoother loss landscape. Furthermore, we propose {\bf L}ifelong {\bf U}nsupervised {\bf M}ixu{\bf p} (\textsc{Lump}) for UCL, which effectively alleviates catastrophic forgetting and provides better qualitative interpretations. 
\end{itemize}

\section{Related work}

{\bf Continual learning.}
We can partition the existing continual learning methods into three categories. The \emph{regularization} approaches~\citep{LiZ2016eccv, zenke17si, schwarz2018progress, ahn2019uncertainty} impose a regularization constraint to the objective that mitigates catastrophic forgetting. The \emph{architectural} approaches~\citep{rusu2016progressive, YoonJ2018iclr, li2019learn}  avoid this problem by including task-specific parameters and allowing the expansion of the network during continual learning. The ~\emph{rehearsal} approaches~\citep{rebuffi2017icarl, rolnick19er, Aljundi2019GradientBS} allocate a small memory buffer to store and replay the examples from the previous task. However, all these methods are confined to supervised learning, which limits their application in real-life problems. \cite{rao19curl, smith21unsupervised} tackled the problem of continual unsupervised representation learning; however, their methods are restricted to simple low-resolution tasks and not scalable to large-scale continual learning datasets.

{\bf Representational learning.} A large number of works have addressed the unsupervised learning problem in the standard machine learning framework. Specifically, contrastive learning frameworks~\citep{he2019moco, chen20simclr, grill20byol, chen2020big, chen2020mocov2} that learn the representations by measuring the similarities of positive and negative pairs have gained a lot of attention in the community. However, all these methods require large batches and negative sample pairs, which restrict the scalability of these networks.~\cite{chen2020exploring} tackled these limitations and proposed \emph{SimSiam}, that use standard Siamese networks~\citep{bromley94siamese} with the stop-gradient operation to prevent the collapsing of Siamese networks to a constant. Recently, ~\cite{zbontar2021barlow} formulated an objective that pushes the cross-correlation matrix between the embeddings of distorted versions of a sample closer to the identity matrix. However, all these methods assume the presence of large datasets for pre-training, which is impractical in real-world applications. In contrast, we tackle the problem of incremental representational learning and learn the representations sequentially while maximizing task adaptation and minimizing catastrophic forgetting.
\section{Preliminaries}

\subsection{Problem Setup}
We consider the continual learning setting, where we learn on a continuum of data consisting of $T$ tasks $\mathcal{T}_{1:T} = \left(\mathcal{T}_1 \ldots \mathcal{T}_T\right)$. In supervised continual learning, each task consists a task descriptor $\tau \in \{1\ldots T\}$ and its corresponding dataset $\mathcal{D}_\tau=\{\left(\bm{x}_{i,\tau}, y_{i, \tau}\right)_{i=1}^{n_\tau}\}$ with $n_\tau$ examples. Each input-pair $(\bm{x}_{i, \tau}, y_{i, \tau}) \in \mathcal{X}_{\tau} \times \mathcal{Y}_\tau$, where \((\mathcal{X}_{\tau}, \mathcal{Y}_\tau)\) is an unkown data distribution. Let us consider a network $f_{\Theta}: \mathcal{X}_\tau \rightarrow \mathbb{R}^D$ parametrized by $\Theta = \{\bm{w}_l\}_{l=1}^{l=L}$, where  $\mathbb{R}^D$ and $L$ denote $D$-dimensional embedding space and number of layers respectively. The classifier is denoted by $h_{\psi}: \mathbb{R}^D \rightarrow \mathcal{Y}_\tau$. The network error using cross entropy loss (CE) for SCL with finetuning can be formally defined as:
    \begin{equation}
        \mathcal{L}_{\textsc{SCL}}^{\textsc{Finetune}} =  \textsc{CE}\left(h_\psi\left(f_\Theta\left(\bm{x}_{i,\tau}\right), \tau\right), y_{i,\tau}\right).
        \label{eq:scl_objective}
    \end{equation}

In this work, we assume the absence of label supervision during training and focus on unsupervised continual learning. In particular, each task consists of \(\mathcal{U}_\tau=\{\left(\bm{x}_{i,\tau}\right)_{i=1}^{n_\tau}\}\), \(\bm{x}_{i, \tau} \in \mathcal{X}_\tau\) with $n_\tau$ examples. Our aim is to learn the representations $f_{\Theta}: \mathcal{X}_\tau \rightarrow \mathbb{R}^D$ on a sequence of tasks while preserving the knowledge of the previous tasks. We introduce the representation learning framework and propose \lump in \cref{sec:ucl} to learn unsupervised representations while effectively mitigating catastrophic forgetting.

\subsection{Learning Protocol and Evaluation Metrics}
Currently, the traditional continual learning strategies follow the standard training protocol, where we learn the network representations $f_{\Theta}: \mathcal{X}_\tau \rightarrow \mathcal{Y}_\tau$ on a sequence of tasks. In contrast, our goal is to learn the feature representations $f_{\Theta}: \mathcal{X}_\tau \rightarrow \mathbb{R}^D$, so we follow a two-step learning protocol to obtain the model predictions. First, we pre-train the representations on a sequence of tasks \(T_{1:T}=\left(\mathcal{T}\ldots\mathcal{T}_T\right)\) to obtain the representations. Next,  we evaluate the quality of our pre-trained representations using a K-nearest neighbor (KNN) classifier~\citep{wu18knn} following the setup in~\cite{chen20simclr, chen2020exploring, zbontar2021barlow}.

To validate knowledge transfer of the learned representations, we adopt the metrics from the SCL literature~\citep{chaudhry2019continual, mirzadeh2020understanding}. Let \(a_{\tau,i}\) denote the test accuracy of task $i$ after learning task $\mathcal{T}_\tau$ using a KNN on frozen pre-trained representations on task \(\mathcal{T}_\tau\). More formally, we can define the metrics to evaluate the continually learned representations as follow:
\begin{enumerate}[leftmargin=1.2em, partopsep=0em]
    \item {\bf Average accuracy} is the average test accuracy of all the tasks completed until the continual learning of task \(\tau\):        $A_\tau = \frac{1}{\tau}\sum_{i=1}^{\tau}a_{\tau,i}$
    \item {\bf Average Forgetting} is the average performance decrease of each task between its maximum accuracy and accuracy at the completion of training: $F~=~\frac{1}{T-1}\sum_{i=1}^{T-1}\max_{{\tau\in \{1,\ldots,T\}}}\left(a_{\tau,i} - a_{T,i}\right)$
\end{enumerate}
\section{Unsupervised Continual Learning}\label{sec:ucl}
\subsection{Continuous representation learning with sequential tasks} \label{sec:rep_framework}
To learn feature representations, contrastive learning~\citep{chen20simclr, chen2020big, he2019moco} maximizes the similarity of representations between the images of the same views (positive pairs) and minimizes the similarity between images of different views (negative pairs). However, these methods require large batches, negative sample pairs~\citep{chen20simclr, chen2020big}, or architectural modifications~\citep{he2019moco, chen2020mocov2}, or non-differentiable operators~\citep{caron2021unsupervised}, which makes their application difficult for continual learning scenarios. In this work, we focus on SimSiam~\citep{chen2020exploring} and BarlowTwins~\citep{zbontar2021barlow}, which tackle these limitations and achieve state-of-the-art performance on standard representation learning benchmarks.

{\bf SimSiam~\citep{chen2020exploring}} uses a variant of Siamese networks~\citep{bromley94siamese} for learning input data representations. It consists of an encoder network \(f_\Theta\), which is composed of a backbone network, and is shared across a projection MLP and prediction MLP head $h(\cdot)$. Specifically, SimSiam minimizes the cosine-similarity between the output vectors of the projector and the predictor MLP across two different augmentations for an image. Initially, we consider \textsc{Finetune}, which is a a naive CL baseline that minimizes the cosine-similarity between the projector output \((z_{i, \tau}^1 = f_\Theta(x_{i, \tau}^1))\) and the predictor output \((p_{i, \tau}^2 = h(f_\Theta(x_{i, \tau}^2))\) on a sequence of tasks as follows:
\begin{align}\label{eq:lossimsiam}
    \mathcal{L}_{\textsc{UCL}}^{\textsc{Finetune}} = \frac{1}{2}D(p_{i, \tau}^1, \texttt{stopgrad}(z_{i, \tau}^2)) + \frac{1}{2}D(&p_{i, \tau}^2, \texttt{stopgrad}(z_{i, \tau}^1)), \\ \nonumber
    \text{where}~~ D(p_{i, \tau}^1, z_{i, \tau}^2) &= - \frac{p_{i, \tau}^1}{\lVert p_{i, \tau}^2 \rVert_2}\cdot\frac{z_{i, \tau}^2}{\lVert z_{i, \tau}^2\rVert_2},
\end{align}

 $x_{i, \tau}^1$ and $x_{i, \tau}^2$ are two randomly augmented views of an input example \(x_{i, \tau} \in \mathcal{T}_\tau\) and \(\lVert \cdot \rVert_2\) denotes the $\ell_2$-norm. Note that, the \(\texttt{stopgrad}\) is a crucial component in SimSiam to prevent the trivial solutions obtained by Siamese networks. Due to its simplicity and effectiveness, we chose Simsiam to analyze the performance of unsupervised representations for continual learning.

{\bf BarlowTwins~\citep{zbontar2021barlow}} minimizes the redundancy between the embedding vector components of the distorted versions of an instance while conserving the maximum information inspired from~\cite{barlow1961possible}. In particular, the objective function eliminates the SimSiam \(\texttt{stopgrad}\) component and instead makes the cross-correlation matrix computed between the outputs of two identical networks closer to the identity matrix. Let $\mathcal{C}$ be the cross-correlation matrix between the outputs of two Siamese branches along the batch dimension and $Z_1$ and $Z_2$ denote the batch embeddings of the distorted views for all images of a batch from the current task $(x_{\tau} \in \mathcal{U}_\tau)$. The objective function for UCL with finetuning and BarlowTwins can then be defined as:
\begin{align}
 \mathcal{L}_{\textsc{UCL}}^{\textsc{Finetune}} = \sum_i  (1-\mathcal{C}_{ii})^2 + ~~\lambda \cdot \sum_{i}\sum_{j \neq i} {\mathcal{C}_{ij}}^2,
~~\text{where}~~\mathcal{C}_{ij} = \frac{
\sum_\mathcal{B} z^1_{\mathcal{B},i} z^2_{\mathcal{B},j}}
{\sqrt{\sum_\mathcal{B} {(z^1_{\mathcal{B},i})}^2} \sqrt{\sum_\mathcal{B} {(z^2_{\mathcal{B},j})}^2}}.
\label{eq:lossBarlow}
\end{align}

$\lambda$ is a positive constant trading off the importance of the invariance and redundancy reduction terms of the loss, $i$ and $j$ denote the network's output vector dimensions. Similar to SimSiam, BarlowTwins is simple, easy to implement, and can be applied to existing continual learning strategies with little or no modification.

\subsection{Preserving representational continuity: A view of existing SCL methods} \label{sec:scl_variants}
Learning feature representations from labelled instances on a sequence of tasks has been long studied in continual learning. However, the majority of these learning strategies are not directly applicable to UCL. To compare with the regularization-based strategies, we extend \emph{Synaptic Intelligence~(SI)}~\citep{zenke17si} to UCL and consider the online per-synapse consolidation during the entire training trajectory of the unsupervised representations. For architectural-based strategies, we investigate \emph{Progressive Neural Networks (PNN)}~\citep{rusu2016progressive} and learn the feature representations progressively using the representations learning frameworks proposed in \cref{sec:rep_framework}.

We also formulate \emph{Dark Experience Replay (DER)}~\citep{buzzega2020dark} for UCL. DER for SCL alleviates catastrophic forgetting by matching the network logits across a sequence of tasks during the optimization trajectory. Notably, the loss for SCL-DER can be defined as follow:
\begin{equation}
        \mathcal{L}_{\textsc{SCL}}^{\textsc{Der}} =  \mathcal{L}_{\textsc{SCL}}^{\textsc{Finetune}} +~\alpha \cdot \mathbb{E}_{(x,p) \sim \mathcal{M}}\big[ \left\lVert \mathrm{softmax}(p) - \mathrm{softmax}(h_{\psi}(x_{i, \tau}))\right\rVert^2_2\big],
\end{equation}
where $p = h_{{\psi}_\tau(x)}$, $\mathcal{L}_{\textsc{SCL}}^{\textsc{Finetune}}$ denotes the cross-entropy loss on the current task (see \Cref{eq:scl_objective}) and random examples are selected using reservoir sampling from the replay-buffer $\mathcal{M}$.  Since, we do not have access to the labels for UCL, we cannot minimize the aforementioned objective. 

Instead, we utilize the output of the projected output by the backbone network to preserve the knowledge of the past tasks over the entire training trajectory. In particular, DER for UCL consists of a combination of two terms. The first term learns the representations using SimSiam from \cref{eq:lossimsiam} or BarlowTwins from \cref{eq:lossBarlow} and the second term minimizes the Euclidean distance between the projected outputs to minimize catastrophic forgetting. More formally, UCL-DER minimizes the following loss:
\begin{equation}
    \mathcal{L}_{\textsc{UCL}}^{\textsc{Der}} =  \mathcal{L}_{\textsc{UCL}}^{\textsc{Finetune}} +~\alpha \cdot \mathbb{E}_{(x) \sim \mathcal{M}}\big[ \left\lVert f_{\Theta_{\tau}}(x) - f_{\Theta}(x_{i, \tau})\right\rVert^2_2\big]
\end{equation}

However, the performance of the rehearsal-based methods is sensitive to the choice of $\alpha$ and often requires supervised training setup, task identities, and boundaries. To tackle this issue, we propose Lifelong Unsupervised Mixup in the subsequent subsection, which interpolates between the current and past task instances to mitigate catastrophic forgetting effectively.

\subsection{Lifelong unsupervised mixup}
The standard Mixup~\citep{zhang2017mixup} training constructs virtual training examples based on the principle of Vicinal Risk Minimization~\cite{}. In particular, let $(x_i, y_i)$ and $(x_j, y_j)$ denote two random feature-target pairs sampled from the training data distribution and let $(\tilde{x}, \tilde{y})$ denote the interpolated feature-target pair in the vicinity of these examples; mixup then minimizes the following objective:
  \begin{align}
        \mathcal{L}^{\textsc{Mixup}}(\tilde{x}, \tilde{y})&=  \textsc{CE}\left(h_\psi\left(f_\Theta\left(\tilde{x}\right)\right), \tilde{y}\right), \\ \nonumber~~\text{where}~~\tilde{x}
   &= \lambda \cdot x_i + (1-\lambda) \cdot x_j ~~\text{and}~~\tilde{y} = \lambda \cdot y_i +
   (1-\lambda) \cdot y_j.
    \end{align}

$\lambda \sim \text{Beta}(\alpha, \alpha)$, for $\alpha \in (0, \infty)$. In this work, we focus on lifelong self-supervised learning and propose Lifelong Unsupervised Mixup (\textsc{Lump}) that utilizes mixup for UCL by incorporating the instances stored in the replay-buffer from the previous tasks into the vicinal distribution. In particular, \lump interpolates between the examples of the current task ($x_{i, \tau}) \in \mathcal{U}_{\tau}$ and random examples selected using uniform sampling from the replay buffer, which encourages the model to behave linearly across a sequence of tasks. More formally, \lump minimizes the objective in \cref{eq:lossimsiam} and \cref{eq:lossBarlow} on the following interpolated instances $\tilde{x}_{i, \tau}$ for the current task $\tau$:
\begin{equation}
\tilde{x}_{i, \tau} = \lambda \cdot x_{i, \tau} + (1 - \lambda) \cdot x_{j, \mathcal{M}},  
\end{equation}
where $x_{j, \mathcal{M}} \sim \mathcal{M}$ denotes the example selected using uniform sampling from replay buffer $\mathcal{M}$. The interpolated examples not only augments the past tasks' instances in the replay buffer but also approximates a regularized loss minimization~\citep{zhang2021how}. During UCL, \lump enhances the robustness of learned representation by revisiting the attributes of the past task that are similar to the current task.
Recently, \cite{kim2020mixco, lee2021imix, verma2021towards, shen2022mix} also employed mixup for contrastive learning. Our work is different from these existing works in that our objective is different, and we focus on unsupervised continual learning. 
 To this end, \lump successively mitigates catastrophic forgetting and learns discriminative \& human-perceptual features over the current state-of-the-art SCL strategies (see \cref{tab:main_table} and \Cref{fig:feature_maps}).

\section{Experiments}

 \subsection{Experimental setup}
 
 {\bf Baselines.} We compare with multiple supervised and unsupervised continual learning baselines across different categories of continual learning methods.
 
\begin{enumerate}[leftmargin=1.2em, partopsep=0em]
     \item {\bf Supervised continual learning.} \textsc{Finetune} is a vanilla supervised learning method trained on a sequence of tasks without regularization or episodic memory and \textsc{Multitask} optimizes the model on complete data. For regularization-based CL methods, we compare against \textsc{SI}~\citep{zenke17si} and \textsc{AGEM}~\citep{chaudhry2018efficient}. We include \textsc{PNN}~\citep{rusu2016progressive} for architecture-based methods. Lastly, we consider \textsc{GSS}~\citep{Aljundi2019GradientBS} that populates the replay-buffer using solid-angle minimization and \textsc{DER}~\citep{buzzega2020dark} matches the network logits sampled through the optimization trajectory for rehearsal during continual learning. 
     \item  {\bf Unsupervised continual learning.} We consider the unsupervised variants of various SCL baselines to show the utility of the unsupervised representations for sequential learning. Specifically, we use \textsc{SimSiam}~\citep{chen2020exploring} and \textsc{BarlowTwins}~\citep{zbontar2021barlow}, which are the state-of-the-art representational learning techniques for learning the unsupervised continual representations. We compare with \textsc{Finetune} and \textsc{Multitask} following the supervised learning baselines, and
 \textsc{Si}~\citep{zenke17si}, \textsc{Pnn}~\citep{rusu2016progressive} for unsupervised regularization and architecture CL methods respectively. For rehearsal-based method, we compare with the UCL variant of \textsc{Der}~\citep{buzzega2020dark} described in \cref{sec:scl_variants}
 \end{enumerate}
 
{\bf Datasets.} 
We compare the performance of SCL and UCL on various continual learning benchmarks using single-head ResNet-18~\citep{he2016deep} architecture. {\bf Split CIFAR-10}~\citep{alex12cifar} consists of two random classes out of the ten classes for each task.
{\bf Split~CIFAR-100}~\citep{alex12cifar} consists of five random classes out of the 100 classes for each task. 
{\bf Split Tiny-ImageNet} is a variant of the ImageNet dataset~\citep{deng2009imagenet} containing five random classes out of the 100 classes for each task with the images sized 64 $\times$ 64 pixels.

{\bf Training and evaluation setup.} We follow the hyperparameter setup of \cite{buzzega2020dark} for all the SCL strategies and tune them for the UCL representation learning strategies. All the learned representations are evaluated with KNN classifier~\citep{wu18knn} across three independent runs. Further, we use the hyper-parameters obtained by SimSiam for training UCL strategies with BarlowTwins to analyze the sensitivity of UCL to hyper-parameters and for a fair comparison between different methods. We train all the UCL methods for 200 epochs and evaluate with the KNN classifier~\citep{wu18knn}. We provide the hyper-parameters in detail in \cref{tab:hyperparameters}.

\subsection{Quantitative results}\label{sec:quantitative}
{\bf Evaluation on SimSiam.} 
\cref{tab:main_table} shows the evaluation results for supervised and unsupervised representations learnt by SimSiam~\citep{chen2020exploring} across various continual learning strategies. In all cases, continual learning with unsupervised representations achieves significantly better performance than supervised representations with substantially lower forgetting. For instance, \textsc{Si} with UCL obtains better performance and $68\%$, $54\%$, and $44\%$ lower forgetting relative to the best-performing SCL strategy on Split CIFAR-10, Split CIFAR-100, and Split Tiny-ImageNet, respectively.
Surprisingly, \textsc{Finetune} with UCL achieves higher performance and significantly lower forgetting in comparison to all SCL strategies except \textsc{Der}. Furthermore, \textsc{Lump} improves upon the UCL strategies: $2.8\%$ and $5.9\%$ relative increase in accuracy and $15\%$ and $57.1\%$ relative decrease in forgetting on Split CIFAR-100 and Split Tiny-ImageNet, respectively.

{\bf Evaluation on BarlowTwins.} To verify that unsupervised representations are indeed more robust to catastrophic forgetting, we train BarlowTwins~\citep{zbontar2021barlow} on a sequence of tasks. We notice that the representations learned with BarlowTwins substantially improve the accuracy and forgetting over SCL: $71.4\%$, $69.7\%$ and $73.2\%$ decrease in forgetting with \textsc{Finetune} on Split CIFAR-10, Split CIFAR-100 and Split Tiny-ImageNet respectively. Similarly, we observe that \textsc{Si}, and \textsc{Der} are more robust to catastrophic forgetting; however, PNN underperforms on complicated tasks since feature accumulation using adaptor modules is insufficient to construct useful representations for current task adaptation. Interestingly, representations learnt with BarlowTwins achieve lower forgetting for \textsc{Finetune}, \textsc{Der} and \textsc{Lump} than SimSiam with comparable accuracy across all the datasets. 

\begin{table}[t]
\caption{\small {\bf Accuracy and forgetting} of the learnt representations on Split CIFAR-10, Split CIFAR-100 and Split Tiny-ImageNet on Resnet-18 architecture with KNN classifier~\citep{wu18knn}. All the values are measured by computing mean and standard deviation across three trials. The best and second-best results are highlighted in {\bf bold} and \underline{underline} respectively. \label{tab:main_table}}
\vspace{-0.1in}
\setlength{\tabcolsep}{3pt} %
\resizebox{\textwidth}{!}{
\begin{tabular}{ll@{\hspace{6pt}}ccccccc}
\toprule
& {\textsc{Method}}&\multicolumn{2}{c}{\textsc{Split CIFAR-10}} &\multicolumn{2}{c}{\textsc{Split CIFAR-100}}&\multicolumn{2}{c}{\textsc{Split Tiny-ImageNet}}\\
\midrule
& & \textsc{Accuracy} & \textsc{Forgetting} & \textsc{Accuracy} & \textsc{Forgetting} & \textsc{Accuracy} & \textsc{Forgetting}\\
\midrule
& \multicolumn{7}{c}{\textsc{Supervised Continual Learning}} \\
\midrule
& \textsc{Finetune} & 
{82.87} \scriptsize($\pm$ 0.47) & {14.26} \scriptsize($\pm$ 0.52) &
{61.08} \scriptsize($\pm$ 0.04) & {31.23} \scriptsize($\pm$ 0.41) &
{53.10} \scriptsize($\pm$ 1.37) & {33.15} \scriptsize($\pm$ 1.22) \\

& \textsc{PNN}~{\small \citep{rusu2016progressive}} & 
{82.74} \scriptsize($\pm$ 2.12) & $-$ &
{66.05} \scriptsize($\pm$ 0.86) & $-$ &
{64.38} \scriptsize($\pm$ 0.92) & $-$ \\

& \textsc{Si}~{\small \citep{zenke17si}} & 
{85.18} \scriptsize($\pm$ 0.65) & {11.39} \scriptsize($\pm$ 0.77) & 
{63.58} \scriptsize($\pm$ 0.37) & {27.98} \scriptsize($\pm$ 0.34) & 
{44.96} \scriptsize($\pm$ 2.41) & {26.29} \scriptsize($\pm$ 1.40) \\

& \textsc{A-gem}~{\small \citep{chaudhry2018efficient}} & 
{82.41} \scriptsize($\pm$ 1.24) & {13.82} \scriptsize($\pm$ 1.27) &
{59.81} \scriptsize($\pm$ 1.07) & {30.08} \scriptsize($\pm$ 0.91) & 
{60.45} \scriptsize($\pm$ 0.24) & {24.94} \scriptsize($\pm$ 1.24) \\

& \textsc{Gss}~{\small \citep{Aljundi2019GradientBS}} & 
{89.49} \scriptsize($\pm$ 1.75) & {~~7.50} \scriptsize($\pm$ 1.52) & 
{70.78} \scriptsize($\pm$ 1.67) & {21.28} \scriptsize($\pm$ 1.52) & 
{70.96} \scriptsize($\pm$ 0.72) & {14.76} \scriptsize($\pm$ 1.22) \\

& \textsc{Der}~{\small \citep{buzzega2020dark}} & 
\underline{91.35 \scriptsize($\pm$ 0.46)} & {~~5.65} \scriptsize($\pm$ 0.35) & 
{79.52} \scriptsize($\pm$ 1.88) & {12.80} \scriptsize($\pm$ 1.47) & 
{68.03} \scriptsize($\pm$ 0.85) & {17.74} \scriptsize($\pm$ 0.65) \\

\cmidrule{2-9}
& \textsc{Multitask} & {97.77} \scriptsize($\pm$ 0.15) & $-$ & {93.89} \scriptsize($\pm$ 0.78) & $-$ & {91.79} \scriptsize($\pm$ 0.46) & $-$ \\
\midrule
& \multicolumn{7}{c}{\textsc{Unsupervised Continual Learning}} \\
\midrule
\parbox[t]{2mm}{\multirow{6}{*}{\rotatebox[origin=c]{90}{\textsc{SimSiam}}}}
& \textsc{Finetune} & 
{90.11} \scriptsize($\pm$ 0.12) & {~~5.42} \scriptsize($\pm$ 0.08) & 
{75.42} \scriptsize($\pm$ 0.78) & {10.19} \scriptsize($\pm$ 0.37) & 
{71.07} \scriptsize($\pm$ 0.20) & {9.48} \scriptsize($\pm$ 0.56) \\

& \textsc{PNN}~{\small \citep{rusu2016progressive}} & 
{90.93} \scriptsize($\pm$ 0.22) & $-$ &
{66.58} \scriptsize($\pm$ 1.00) & $-$ &
{62.15} \scriptsize($\pm$ 1.35) & $-$ \\

& \textsc{Si}~{\small \citep{zenke17si}} & 
{\bf 92.75 \scriptsize($\pm$ 0.06)} & \underline{~~1.81 \scriptsize($\pm$ 0.21)} & 
{80.08} \scriptsize($\pm$ 1.30) & {~~5.54} \scriptsize($\pm$ 1.30) & 
\underline{72.34 \scriptsize($\pm$ 0.42)} & {8.26} \scriptsize($\pm$ 0.64) \\

& \textsc{Der}~{\small \citep{buzzega2020dark}} & 
{91.22} \scriptsize($\pm$ 0.30) & {~~4.63} \scriptsize($\pm$ 0.26) & 
{77.27} \scriptsize($\pm$ 0.30) & {~~9.31} \scriptsize($\pm$ 0.09) & 
{71.90} \scriptsize($\pm$ 1.44) & {8.36} \scriptsize($\pm$ 2.06) \\

& \textsc{Lump} & 
{91.00} \scriptsize($\pm$ 0.40) & {~~2.92} \scriptsize($\pm$ 0.53) &
{\bf 82.30 \scriptsize($\pm$ 1.35)} & \underline{~~4.71 \scriptsize($\pm$ 1.52)} & 
\bf{76.66 \scriptsize($\pm$ 2.39)} & \underline{3.54 \scriptsize($\pm$ 1.04)} \\

\cmidrule{2-9}
& \textsc{Multitask} & {95.76} \scriptsize($\pm$ 0.08) & $-$ & {86.31} \scriptsize($\pm$ 0.38) & $-$  & {82.89} \scriptsize($\pm$ 0.49) & $-$ \\

\midrule 

\parbox[t]{2mm}{\multirow{6}{*}{\rotatebox[origin=c]{90}{\textsc{BarlowTwins}}}}
& \textsc{Finetune} & 
{87.72} \scriptsize($\pm$ 0.32) & {~~4.08} \scriptsize($\pm$ 0.56) & 
{71.97} \scriptsize($\pm$ 0.54) & {~~9.45} \scriptsize($\pm$ 1.01) & 
{66.28} \scriptsize($\pm$ 1.23) & {8.89} \scriptsize($\pm$ 0.66) \\

& \textsc{PNN}~{\small \citep{rusu2016progressive}} & 
{87.52} \scriptsize($\pm$ 0.33) & $-$ &
{57.93} \scriptsize($\pm$ 2.98) & $-$ &
{48.70} \scriptsize($\pm$ 2.59) & $-$ \\

& \textsc{Si}~{\small \citep{zenke17si}} & 
{90.21} \scriptsize($\pm$ 0.08) & {~~2.03} \scriptsize($\pm$ 0.22) & 
{75.04} \scriptsize($\pm$ 0.63) & {~~7.43} \scriptsize($\pm$ 0.67) & 
{56.96} \scriptsize($\pm$ 1.48) & {17.04} \scriptsize($\pm$ 0.89) \\

& \textsc{Der}~{\small \citep{buzzega2020dark}} & 
{88.67} \scriptsize($\pm$ 0.24) & {~~2.41} \scriptsize($\pm$ 0.26) & 
{73.48} \scriptsize($\pm$ 0.53) & {~~7.98} \scriptsize($\pm$ 0.29) &
{68.56} \scriptsize($\pm$ 1.47) & {~~7.87} \scriptsize($\pm$ 0.44) \\

& \textsc{Lump} & 
{90.31} \scriptsize($\pm$ 0.30) & {\bf ~~1.13 \scriptsize($\pm$ 0.18)} &
\underline{80.24 \scriptsize($\pm$ 1.04)} & {\bf ~~3.53 \scriptsize($\pm$ 0.83)} &
{72.17} \scriptsize($\pm$ 0.89) & {\bf ~~2.43 \scriptsize($\pm$ 1.00)} \\

\cmidrule{2-9}
& \textsc{Multitask} & {95.48} \scriptsize($\pm$ 0.14) & $-$ & {87.16} \scriptsize($\pm$ 0.52) & $-$ &  {82.42} \scriptsize($\pm$ 0.74) & $-$\\
\bottomrule
\end{tabular}}
\vspace{-0.2in}
\end{table}

{\bf Evaluation on Few-shot training.} 
\cref{fig:fewshot} compares the effect of few-shot training on UCL and SCL, where each task has a limited number of training instances. Specifically, we conduct the experimental evaluation using $100$, $200$, $500$, and $2500$ training instances for each task in split CIFAR-100 dataset. Surprisingly, we observe that the gap in average accuracy between SCL and UCL methods widens with a decrease in the number of training instances. Note that UCL decreases the accuracy by $15.78\%p$ on average with lower forgetting when the number of training instances decreases from $2500$ to $100$; whereas, SCL obtains a severe $32.21\%p$ deterioration in accuracy.
We conjecture that this is an outcome of the discriminative feature embeddings learned by UCL, which discriminates all the images in the dataset and captures more than class-specific information as also observed in \citet{doersch2021crosstransformers}. Furthermore, \lump improves the performance over all the baselines with a significant margin across all few-shot experiments.

\begin{table*}[t!]
    \centering
    \begin{minipage}[t]{0.43\linewidth}
     \resizebox{1.05\linewidth}{!}{%
    \begin{tabular}{cc}
        \includegraphics[width=0.64\linewidth]{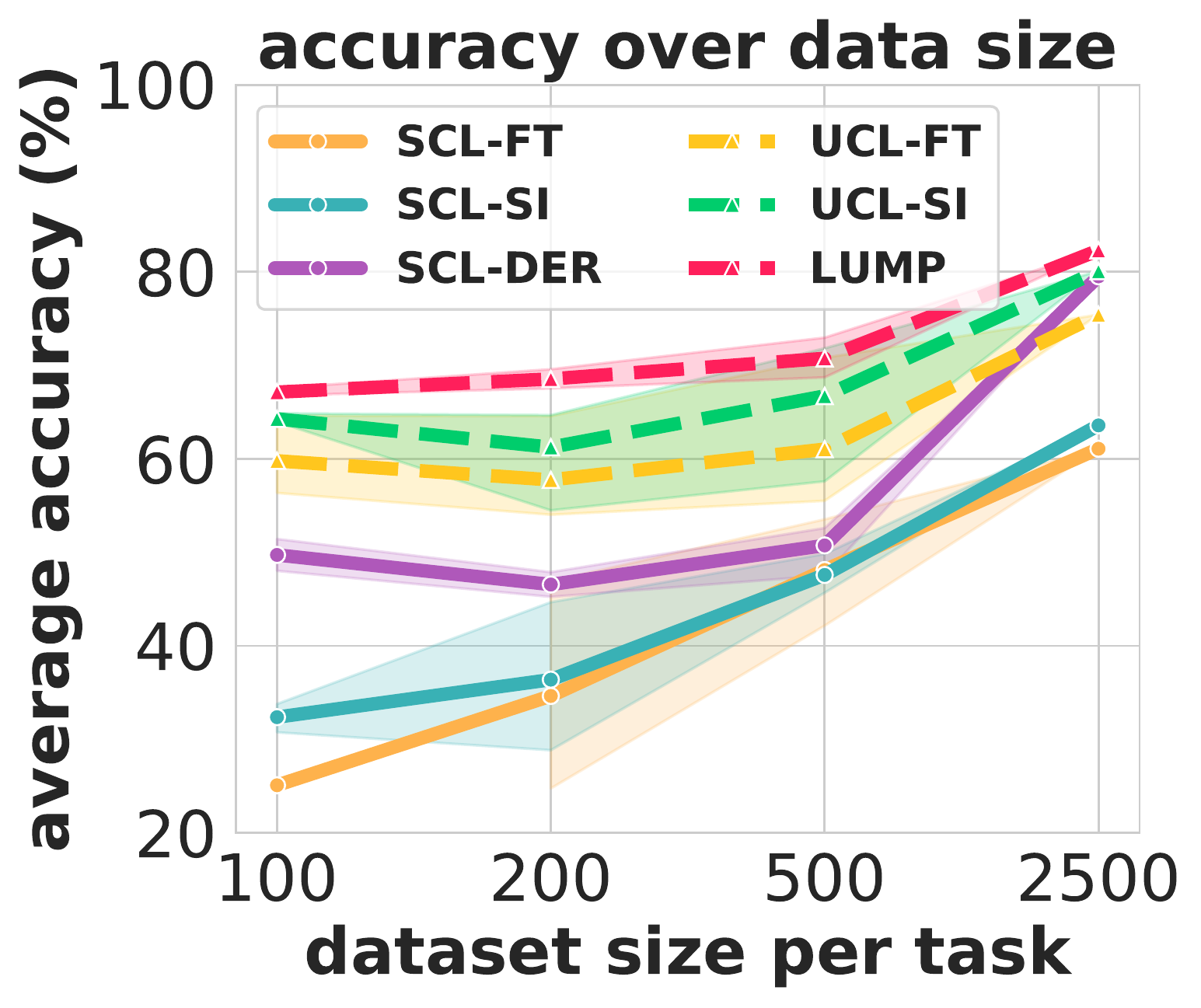} & \hspace{-0.2in}
        \includegraphics[width=0.62\linewidth]{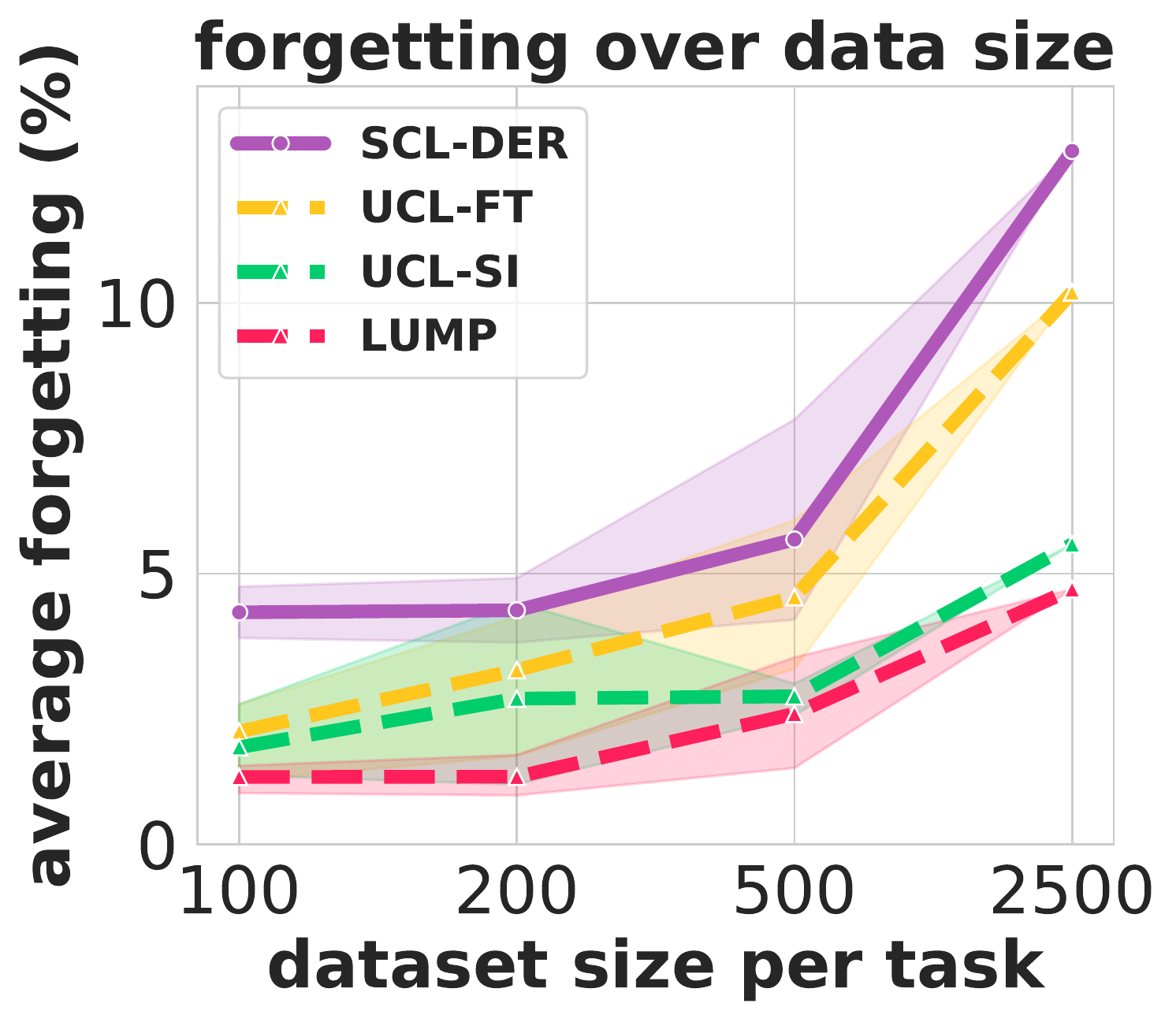}\\
    \end{tabular}}
            \vspace{-0.15in}
            \captionof{figure}{\small \textbf{Evaluation on Few-shot training} for Split CIFAR-100 across different number of training instances per task. The results are measured across three independent trials. \label{fig:fewshot}}
        \end{minipage}%
        \hfill%
           \begin{minipage}[t]{0.55\linewidth}
             \resizebox{\linewidth}{!}{%
           \begin{tabular}{cccc}
    \includegraphics[height=3.05cm]{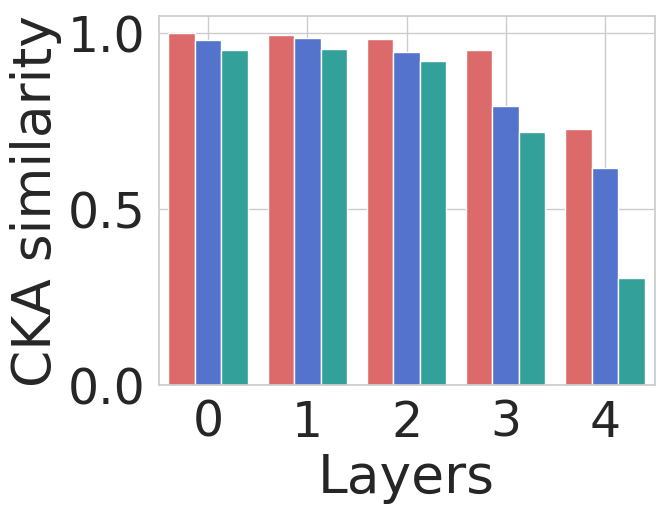} &
 \includegraphics[height=3.05cm]{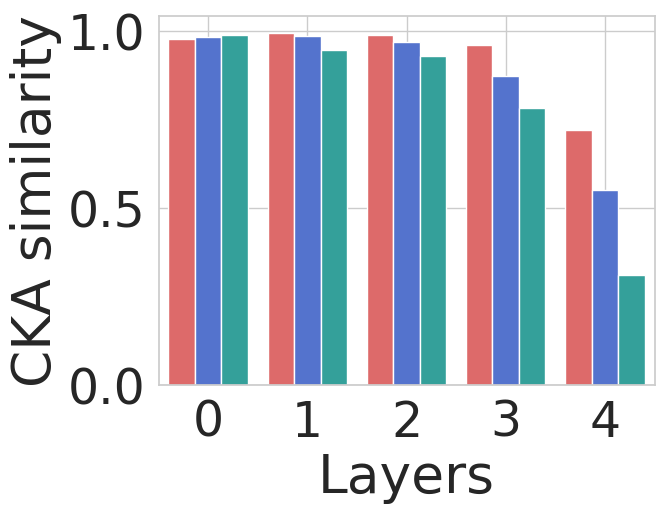} &
\includegraphics[height=3.05cm]{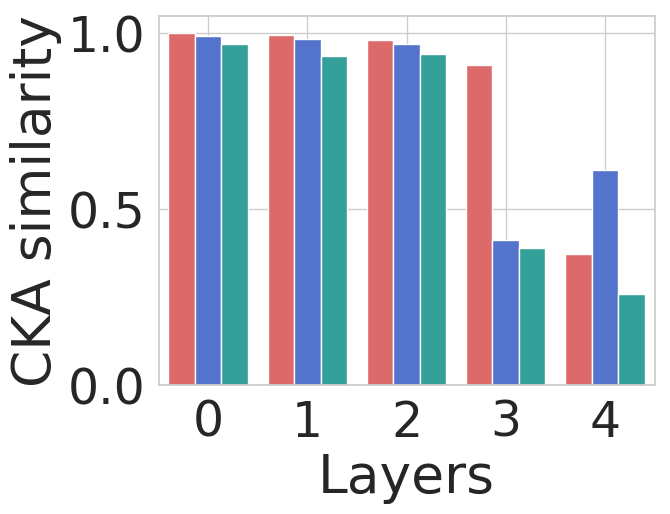} \\
 \multicolumn{1}{c}{\textbf{(a)} \textsc{Finetune} } & \multicolumn{1}{c}{\textbf{(b)} \textsc{Si}} & \multicolumn{1}{c}{\textbf{(c)} \textsc{Der}} \vspace{0.05in}\\
    \end{tabular}}
    \vspace{-0.15in}
    \captionof{figure}{{\bf CKA Feature similarity} between two independent UCL models (\textcolor{redfigure}{\bf red}), two independent SCL models (\textcolor{bluefigure}{\bf blue}), and UCL and SCL model (\textcolor{greenfigure}{\bf green}) for different strategies on Split CIFAR-100 test distribution. \label{fig:cka}}
    \end{minipage}
\vspace{-0.3in}
\end{table*}
{\bf Evaluation on OOD datasets.} We evaluate the learnt representations on various out-of-distribution (OOD) datasets in \cref{tab:ood_table} to measure their generalization to unseen data distributions. In particular, we conduct the OOD evaluation on MNIST~\citep{lecun1998mnist}, Fashion-MNIST (FMNIST)~\citep{xiao2017fashion}, SVHN~\citep{netzer2011reading}, CIFAR-10 and CIFAR-100~\citep{alex12cifar} using a KNN classifier~\citep{wu18knn}. We observe that unsupervised representations outperform the supervised representations in all cases across all the datasets. In particular, the UCL representations learned with Simsiam, and \textsc{Si} on Split-CIFAR-10 improves the absolute performance over the best-performing SCL strategy by 4.58\%, 6.09\%, 15.26\%, and 17.07\% on MNIST, FMNIST, SVHN, and CIFAR-100 respectively. Further, \textsc{Lump} trained on Split-CIFAR-100 outperforms \textsc{Si} across all datasets and obtains comparable performance with Split CIFAR-10 dataset. 
 
\subsection{Qualitative analysis}\label{sec:qualitative}
{\bf Similarity in feature and parameter space.} We analyze the similarity between the representations learnt between (i) Two independent UCL models, (ii) Two independent SCL models (iii) SCL and UCL models using centered kernel alignment (CKA)~\citep{kornblith2019similarity} in \cref{fig:cka}, which provides a score between 0 and 1 measuring the similarity between a pair of hidden representations. For two representations $\Theta_1: \mathcal{X} \rightarrow \mathbb{R}^{d_1}$ and $\Theta_2: \mathcal{X} \rightarrow \mathbb{R}^{d_1}$, $\text{CKA}(\Theta_1, \Theta_2) = \frac{||\text{Cov}( \Theta_1(x), \Theta_2(x) )||_F^2} {||\text{Cov}(\Theta_1(x))||_F \cdot ||\text{Cov}(\Theta_2(x))||_F}$, where covariances are with respect to the test distribution.
Additionally, we measure the $\ell_2$ distance~\citep{neyshabur2021transferred} between the parameters of two independent UCL models (see \cref{tab:l2_u}) and two independent SCL models (see \cref{tab:l2_s}). First, we observe that the representations learned by two independent UCL methods have a high feature similarity and lower $\ell_2$ distance compared to the two independent SCL methods, demonstrating UCL representations' robustness.
Second, we note that the representations between any two independent models are highly similar in the lower layers indicating that they learn similar high-level features, including edges and shapes; however, the features are dissimilar for the higher modules. Lastly, we see that the representations between a UCL and SCL model are similar in the lower layers but diverge in the higher layers across all CL strategies.

\begin{table}[t]
\caption{\small {\bf Comparison of accuracy} on out of distribution datasets using a KNN classifier~\citep{wu18knn} on pre-trained SCL and UCL representations. We consider MNIST~\citep{lecun1998mnist}, Fashion-MNIST (FMNIST)~\citep{xiao2017fashion}, SVHN~\citep{netzer2011reading} as out of distribution for Split CIFAR-100 and Split CIFAR-10. All the values are measured by computing mean and standard deviation across three trials. The best and second-best results are highlighted in {\bf bold} and \underline{underline} respectively. \label{tab:ood_table}}
\vspace{-0.1in}
\setlength{\tabcolsep}{3pt} %
\resizebox{\textwidth}{!}{
\begin{tabular}{ll@{\hspace{6pt}}ccccccccccc}
\toprule
& {\textsc{In-class}}&\multicolumn{4}{c}{\textsc{Split CIFAR-10}} &\multicolumn{4}{c}{\textsc{Split CIFAR-100}} \\
\midrule
& \textsc{Out-of-class} & MNIST & FMNIST & SVHN & CIFAR-100 &  MNIST & FMNIST & SVHN & CIFAR-10\\
\midrule
& \multicolumn{9}{c}{\textsc{Supervised Continual Learning}} \\
\midrule
& \textsc{Finetune} & 
{86.42} \scriptsize($\pm$ 1.11) & {74.47} \scriptsize($\pm$ 0.84) & {41.00} \scriptsize($\pm$ 0.85) & {17.42} \scriptsize($\pm$ 0.96) &
{75.02} \scriptsize($\pm$ 3.97) & {62.37} \scriptsize($\pm$ 3.20) & 
{38.05} \scriptsize($\pm$ 0.73) & {39.18} \scriptsize($\pm$ 0.83) &\\


& \textsc{Si}~{\small \citep{zenke17si}} & 
{87.08} \scriptsize($\pm$ 0.79) & {76.41} \scriptsize($\pm$ 0.81) &
{42.62} \scriptsize($\pm$ 1.31) & {19.14} \scriptsize($\pm$ 0.91) &
{79.96} \scriptsize($\pm$ 2.63) & {63.71} \scriptsize($\pm$ 1.36) & 
{40.92} \scriptsize($\pm$ 1.64) & {40.41} \scriptsize($\pm$ 1.71) &  \\
 
& \textsc{A-gem}~{\small \citep{chaudhry2018efficient}} & 
{86.07} \scriptsize($\pm$ 1.94) & {74.74} \scriptsize($\pm$ 3.21) & {37.77} \scriptsize($\pm$ 3.49) & {16.11} \scriptsize($\pm$ 0.38) &
{77.56} \scriptsize($\pm$ 3.21) & {64.16} \scriptsize($\pm$ 2.29) & 
{37.48} \scriptsize($\pm$ 1.73) & {37.91} \scriptsize($\pm$ 1.33)  \\

& \textsc{Gss}~{\small \citep{Aljundi2019GradientBS}} & 
{70.36} \scriptsize($\pm$ 3.54) & {69.20} \scriptsize($\pm$ 2.51) &
{33.11} \scriptsize($\pm$ 2.26) & {18.21} \scriptsize($\pm$ 0.39) &
{76.54} \scriptsize($\pm$ 0.46) & {65.31} \scriptsize($\pm$ 1.72) & 
{35.72} \scriptsize($\pm$ 2.37) & {49.41} \scriptsize($\pm$ 1.81)  \\

& \textsc{Der}~{\small \citep{buzzega2020dark}} & 
{80.32} \scriptsize($\pm$ 1.91) & {70.49} \scriptsize($\pm$ 1.54) & 
{41.48} \scriptsize($\pm$ 2.76) & {17.72} \scriptsize($\pm$ 0.25) &
{87.71} \scriptsize($\pm$ 2.23) & {75.97} \scriptsize($\pm$ 1.29) & 
\underline{50.26 \scriptsize($\pm$ 0.95)} & {59.07} \scriptsize($\pm$ 1.06) & \\


\cmidrule(l{3pt}r{3pt}){2-2} \cmidrule(l{3pt}r{3pt}){3-7} \cmidrule(l{3pt}r{3pt}){8-12}
 & \textsc{multitask} & {88.79} \scriptsize($\pm$ 1.13) & {79.50} \scriptsize($\pm$ 0.52) & 
{41.26} \scriptsize($\pm$ 1.95) & {27.68} \scriptsize($\pm$ 0.66) &
 { 92.29} \scriptsize($\pm$ 3.37) & {86.12} \scriptsize($\pm$ 1.87) & { 54.94} \scriptsize($\pm$ 1.77) & { 54.04} \scriptsize($\pm$ 3.68) \\
 
\midrule
& \multicolumn{9}{c}{\textsc{Unsupervised Continual Learning}} \\
\midrule
\parbox[t]{2mm}{\multirow{5}{*}{\rotatebox[origin=c]{90}{\textsc{SimSiam}}}}
& \textsc{Finetune} & 
{89.23} \scriptsize($\pm$ 0.99) & {80.05} \scriptsize($\pm$ 0.34) & 
{49.66} \scriptsize($\pm$ 0.81) & {34.52} \scriptsize($\pm$ 0.12) &
{85.99} \scriptsize($\pm$ 0.86) & {76.90} \scriptsize($\pm$ 0.11) & 
{50.09} (\bf \scriptsize $\pm$ 1.41) & {57.15} \scriptsize($\pm$ 0.96) \\


& \textsc{Si}~{\small \citep{zenke17si}} & 
{\bf 93.72} (\bf \scriptsize $\pm$ 0.58)  & {\bf 82.50} (\bf \scriptsize $\pm$ 0.51) & {\bf 57.88} (\bf \scriptsize $\pm$ 0.16) &  {\bf 36.21} (\bf \scriptsize $\pm$ 0.69)  &
{91.50} \scriptsize($\pm$ 1.26) & {80.57} \scriptsize($\pm$ 0.93) & 
{\bf 54.07} (\bf \scriptsize $\pm$ 2.73) & {60.55} (\bf \scriptsize $\pm$ 2.54) \\

& \textsc{Der}~{\small \citep{buzzega2020dark}} & 
{88.35} \scriptsize($\pm$ 0.82) & 
{79.33} \scriptsize($\pm$ 0.62) & {48.83} \scriptsize($\pm$ 0.55)) & {30.68} \scriptsize($\pm$ 0.36)  &
{87.96} \scriptsize($\pm$ 2.04) & {76.21} \scriptsize($\pm$ 0.63) & 
{47.70} \scriptsize($\pm$ 0.94) & {56.26} \scriptsize($\pm$ 0.16)  \\

& \textsc{Lump} & 
\underline{{91.03} \scriptsize($\pm$ 0.22)} & {80.78} \scriptsize($\pm$ 0.88) & {45.18} \scriptsize($\pm$ 1.57) & {31.17} \scriptsize($\pm$ 1.83)  &
{\bf 91.76} (\bf \scriptsize $\pm$ 1.17) & {\bf 81.61} (\bf \scriptsize $\pm$ 0.45) & 
{50.13} \scriptsize($\pm$ 0.71) & {\bf 63.00} (\bf \scriptsize $\pm$ 0.53) \\

\cmidrule(l{3pt}r{3pt}){2-2} \cmidrule(l{3pt}r{3pt}){3-7} \cmidrule(l{3pt}r{3pt}){8-12}
 & \textsc{Multitask} &
 {90.69} \scriptsize($\pm$ 0.13) & {80.65} \scriptsize($\pm$ 0.42) & 
{47.67} \scriptsize($\pm$ 0.45) & {39.55} \scriptsize($\pm$ 0.18) &
 { 90.35} \scriptsize($\pm$ 0.24) & {81.11} \scriptsize($\pm$ 1.86) & { 52.20} \scriptsize($\pm$ 0.61) & {70.19} \scriptsize($\pm$ 0.15)  \\
 \midrule
 \parbox[t]{2mm}{\multirow{5}{*}{\rotatebox[origin=c]{90}{\footnotesize \textsc{BarlowTwins}}}}
& \textsc{Finetune} & 
{86.86} \scriptsize($\pm$ 1.62) & {78.37} \scriptsize($\pm$ 0.74) & 
{44.64} \scriptsize($\pm$ 2.39) & {28.03} \scriptsize($\pm$ 0.52) &
{76.08} \scriptsize($\pm$ 2.86) & {76.82} \scriptsize($\pm$ 0.83) & 
{42.95} \scriptsize($\pm$ 0.90) & {53.12} \scriptsize($\pm$ 0.13) & \\


& \textsc{Si}{\small ~\citep{zenke17si}} & 
{90.31} \scriptsize($\pm$ 0.69) & {80.58} \scriptsize($\pm$ 0.68) & 
{49.18} \scriptsize($\pm$ 0.51) & \underline{{31.80} \scriptsize($\pm$ 0.4)} &
{85.24} \scriptsize($\pm$ 0.99) & {78.82} \scriptsize($\pm$ 0.67) & 
{45.18} \scriptsize($\pm$ 1.37) & {53.99} \scriptsize($\pm$ 0.56) \\

& \textsc{Der}~{\small \citep{buzzega2020dark}} & 
{85.15} \scriptsize($\pm$ 2.19) & {77.96} \scriptsize($\pm$ 0.59) & 
{45.68} \scriptsize($\pm$ 0.93) & {27.83} \scriptsize($\pm$ 0.86) &
 {78.08} \scriptsize($\pm$ 1.95)  & {76.67} \scriptsize($\pm$ 0.68) & 
{44.58} \scriptsize($\pm$ 1.01) & {53.24} \scriptsize($\pm$ 0.82) \\

& \textsc{Lump} & 
{88.73} \scriptsize($\pm$ 0.54) & \underline{{81.69} \scriptsize($\pm$ 0.45)} & 
\underline{{51.53} \scriptsize($\pm$ 0.41)} & {31.53} \scriptsize($\pm$ 0.36) &
\underline{90.22} \scriptsize($\pm$ 1.39) & \underline{81.28} \scriptsize($\pm$ 0.91) & 
{50.24} \scriptsize($\pm$ 0.95) & \underline{60.76} \scriptsize($\pm$ 0.87)  \\

\cmidrule(l{3pt}r{3pt}){2-2} \cmidrule(l{3pt}r{3pt}){3-7} \cmidrule(l{3pt}r{3pt}){8-12}
 & \textsc{Multitask} & {88.63} \scriptsize($\pm$ 1.38) & {79.49} \scriptsize($\pm$ 0.29) & {49.24} \scriptsize($\pm$ 2.44) &
{36.33} \scriptsize($\pm$ 0.29) & {86.98} \scriptsize($\pm$ 1.70) & {79.40} \scriptsize($\pm$ 1.10) & {50.19} \scriptsize($\pm$ 0.81) & {49.50} \scriptsize($\pm$ 0.38) \\
\bottomrule
\end{tabular}}
\vspace{-0.1in}
\end{table}

\begin{figure*}[t!]
\small
\begin{minipage}[t]{1.05\linewidth}
\resizebox{\linewidth}{!}{%
\begin{tabular}{c cccc ccc}
\hspace{-0.2in}
\rb{\includegraphics[width=0.13\textwidth]{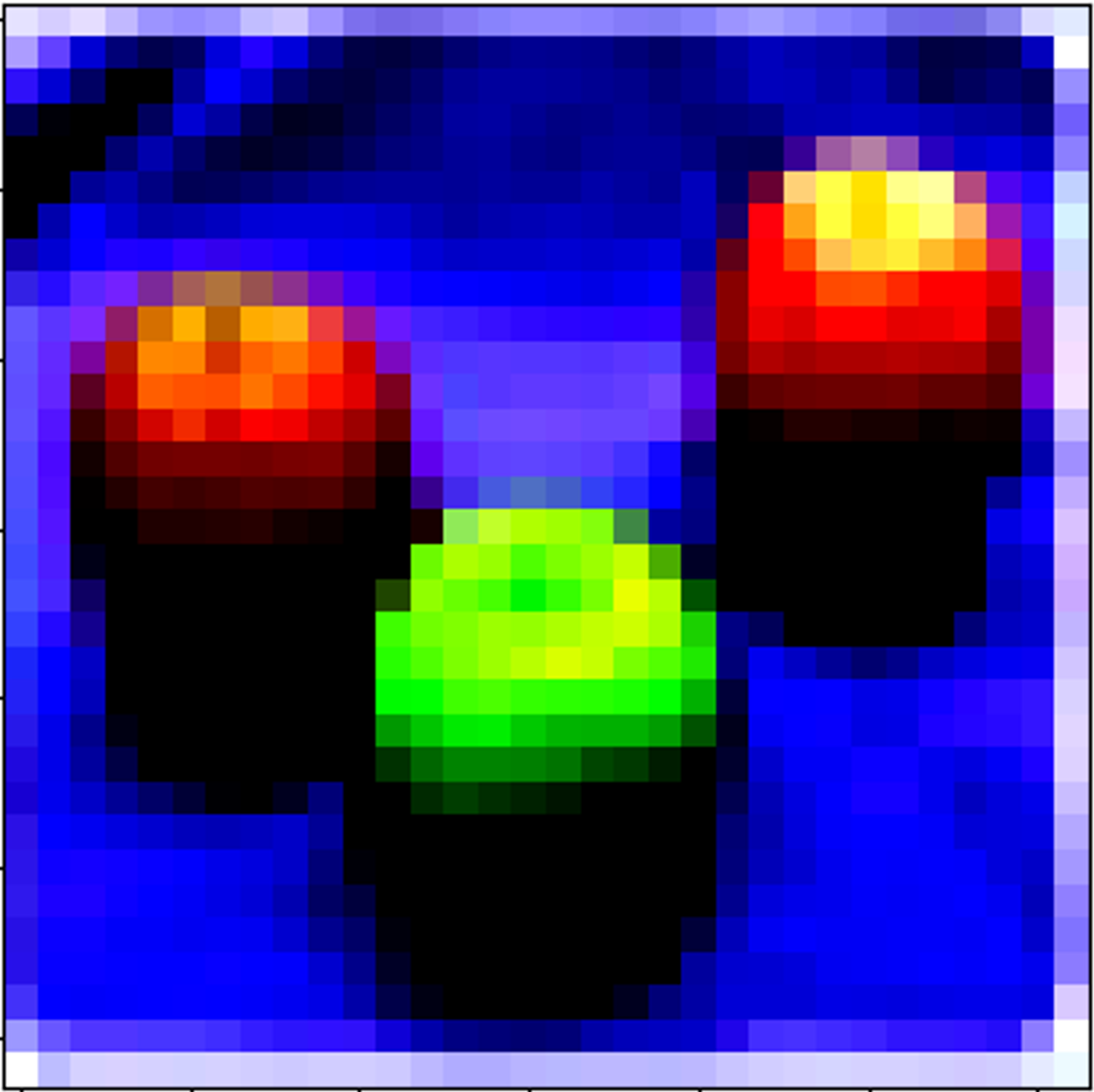}}&
\includegraphics[width=0.17\textwidth]{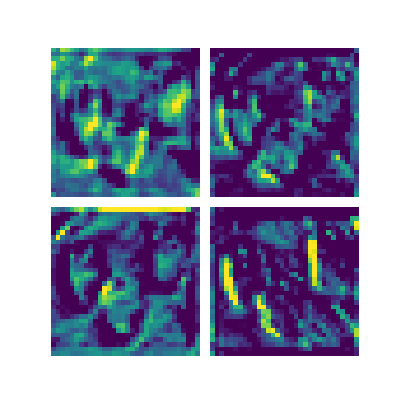} & \hspace{-0.3in}
\includegraphics[width=0.17\textwidth]{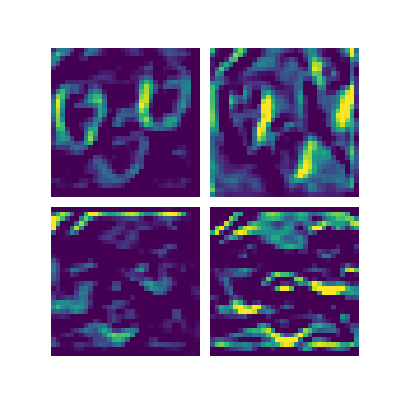} & \hspace{-0.3in}
\includegraphics[width=0.17\textwidth]{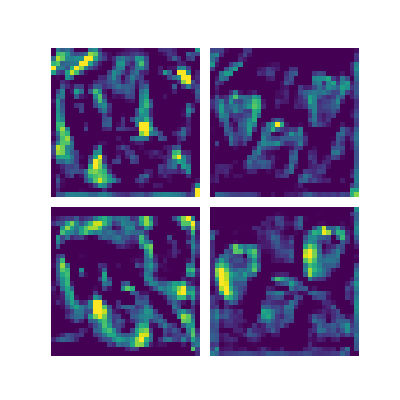} & \hspace{-0.3in}
\includegraphics[width=0.17\textwidth]{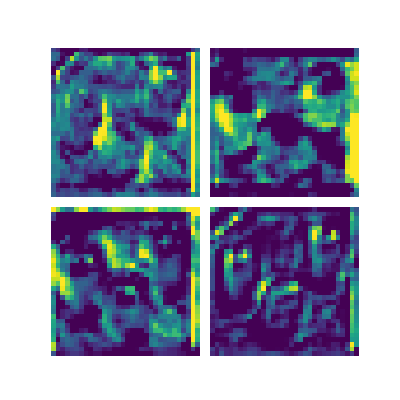} & \hspace{-0.2in}
\includegraphics[width=0.17\textwidth]{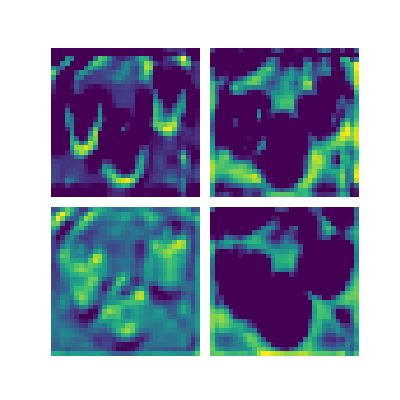} & \hspace{-0.3in}
\includegraphics[width=0.17\textwidth]{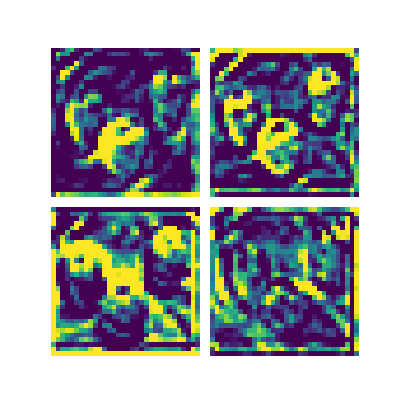} & \hspace{-0.3in}
\includegraphics[width=0.17\textwidth]{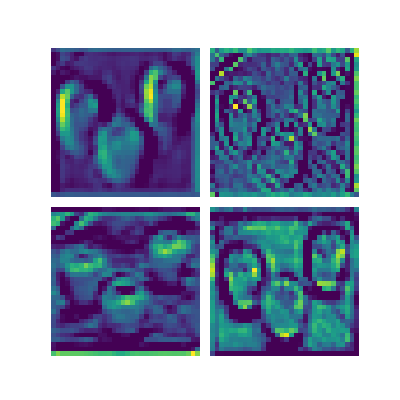} \vspace{-0.1in}\\
\hspace{-0.2in} \textsc{Apple} ($\mathcal{T}_0$) & \specialcell{\textsc{Scl-Finetune}\\{\scriptsize Acc: $54.7\pm 0.2$}} & \hspace{-0.2in} \hspace{-0.2in} \specialcell{\textsc{Scl-Si}\\{\scriptsize Acc: $58.9\pm 0.2$}} & \hspace{-0.3in} \specialcell{\textsc{Scl-Gss}\\{\scriptsize Acc: $78.4\pm 1.8$}} & \hspace{-0.3in} \specialcell{\textsc{Scl-Der}\\{\scriptsize Acc: $73.1\pm 0.4$}} & \hspace{-0.2in} \specialcell{\textsc{Ucl-Finetune}\\{\scriptsize Acc: $70.8\pm 0.4$}} & \hspace{-0.3in} \specialcell{\textsc{Ucl-Si}\\{\scriptsize Acc: $76.4\pm 1.6$}}& \hspace{-0.3in} \specialcell{\textsc{Lump (Ours)}\\{\scriptsize Acc: $76.6\pm 2.7$}}\\
\end{tabular}}
\end{minipage}

\begin{minipage}[t]{1.05\linewidth}
\resizebox{\linewidth}{!}{%
\begin{tabular}{c cccc ccc}
\hspace{-0.2in}
\rb{\includegraphics[width=0.13\textwidth]{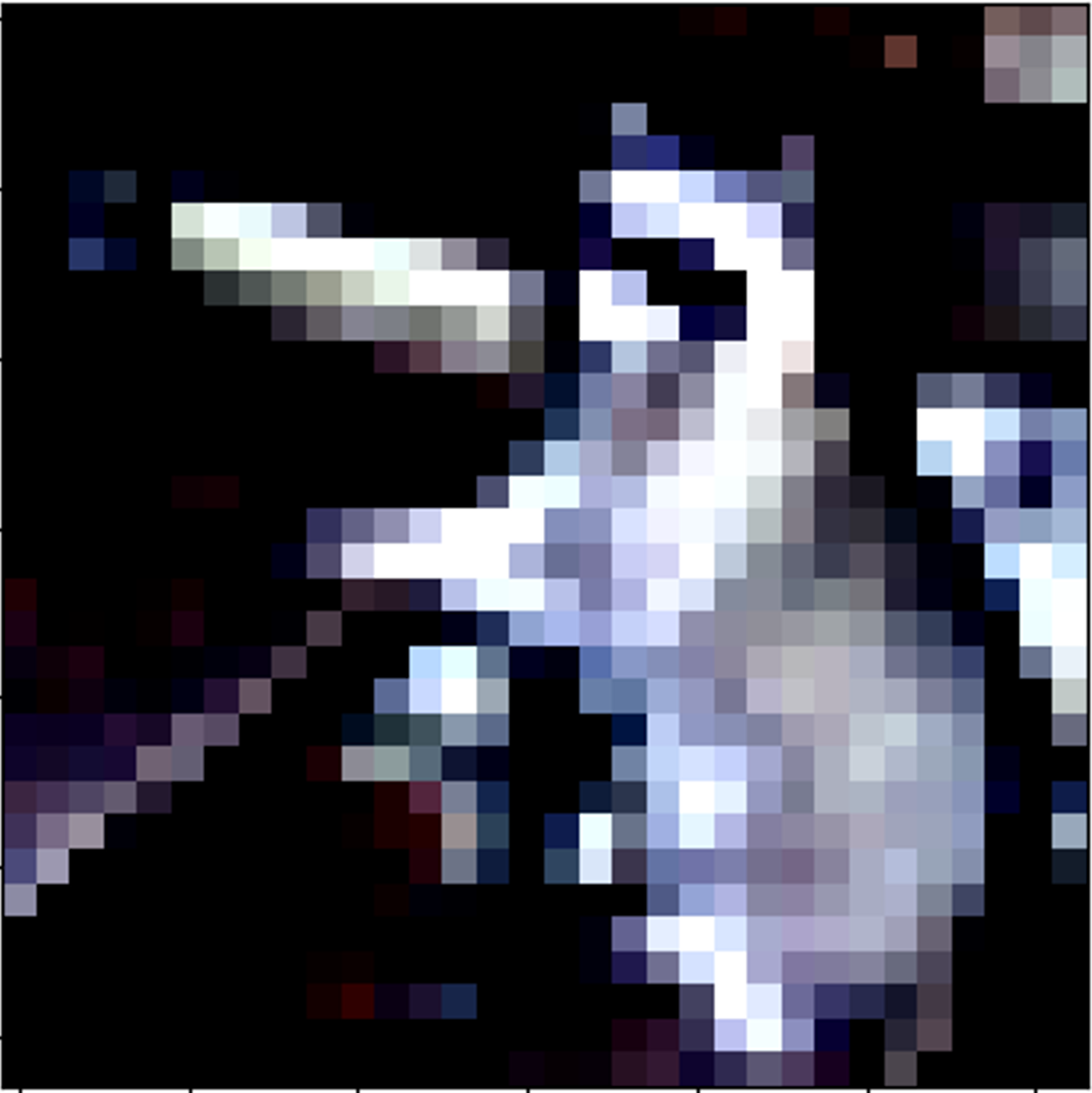}}&
\includegraphics[width=0.17\textwidth]{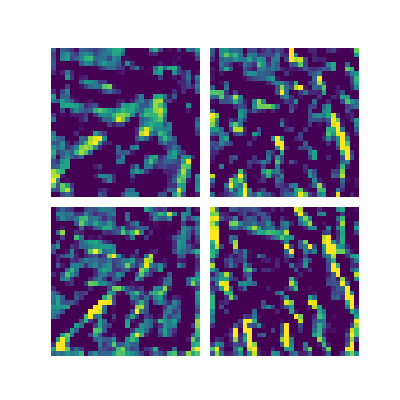} & \hspace{-0.3in}
\includegraphics[width=0.17\textwidth]{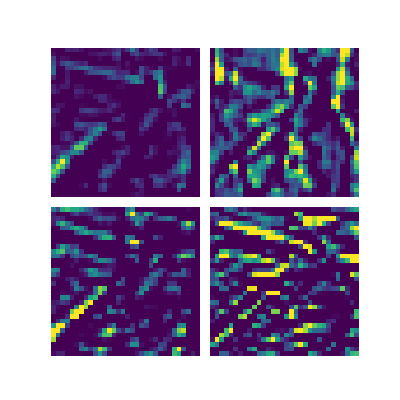} & \hspace{-0.3in}
\includegraphics[width=0.17\textwidth]{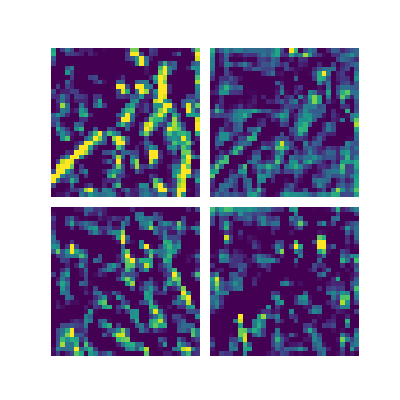} & \hspace{-0.3in}
\includegraphics[width=0.17\textwidth]{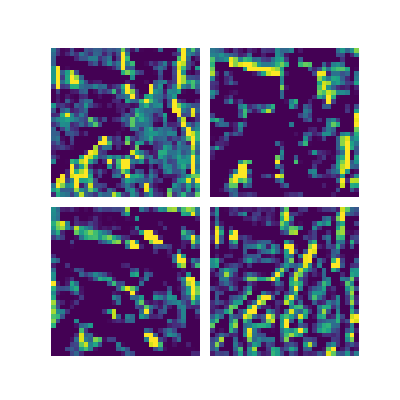} & \hspace{-0.2in}
\includegraphics[width=0.17\textwidth]{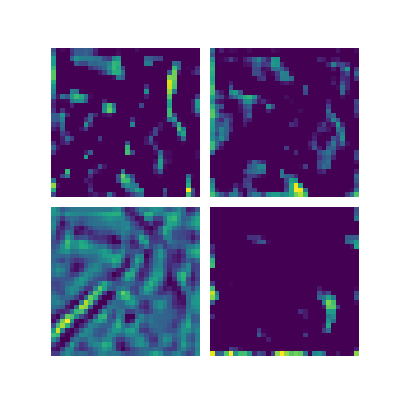} & \hspace{-0.3in}
\includegraphics[width=0.17\textwidth]{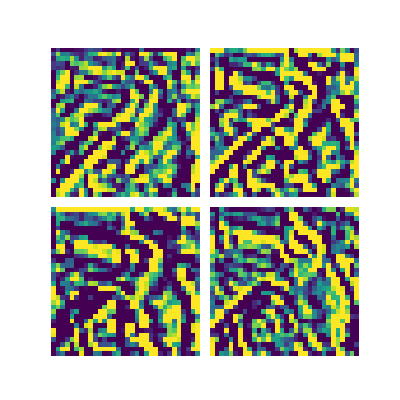} & \hspace{-0.3in}
\includegraphics[width=0.17\textwidth]{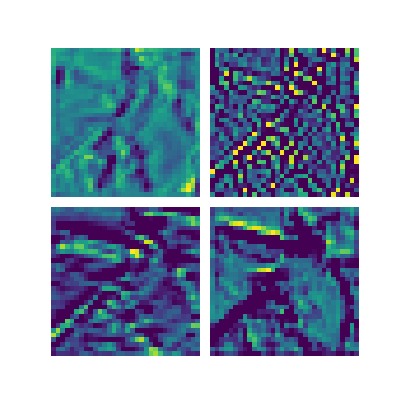} \vspace{-0.12in}\\
\hspace{-0.2in} {\scriptsize \textsc{Raccoon} ($\mathcal{T}_{13}$)} & \specialcell{\textsc{Scl-Finetune}\\{\scriptsize Acc: $50.6\pm 1.4$}} & \hspace{-0.2in} \hspace{-0.2in} \specialcell{\textsc{Scl-Si}\\{\scriptsize Acc: $48.4\pm 1.0$}} & \hspace{-0.3in} \specialcell{\textsc{Scl-Gss}\\{\scriptsize Acc: $59.9\pm 2.2$}} & \hspace{-0.3in} \specialcell{\textsc{Scl-Der}\\{\scriptsize Acc: $76.4\pm 2.2$}} & \hspace{-0.2in} \specialcell{\textsc{Ucl-Finetune}\\{\scriptsize Acc: $74.6\pm 0.5$}} & \hspace{-0.3in} \specialcell{\textsc{Ucl-Si}\\{\scriptsize Acc: $78.0\pm 1.6$}}& \hspace{-0.3in} \specialcell{\textsc{Lump (Ours)}\\{\scriptsize Acc: $80.8\pm 0.5$}}\\
\end{tabular}}
\vspace{-0.15in}
\captionof{figure}{\small {\bf Visualization of feature maps} for the second block representations learnt by SCL and UCL strategies  \\(with Simsiam) for ResNet-18 architecture after the completion of CL for Split CIFAR-100 dataset ($n=20$). \label{fig:feature_maps}}
\end{minipage}
\label{fig:visualize}
\end{figure*}

{\bf Visualization of feature space.} Next, we visualize the learned features to dissect further the representations learned by UCL and SCL strategies. \cref{fig:feature_maps} shows the visualization of the latent feature maps for tasks $\mathcal{T}_0$ and $\mathcal{T}_{13}$ after the completion of continual learning. For $\mathcal{T}_0$, we observe that the SCL methods are prone to catastrophic forgetting, as the features appear noisy and do not have coherent patterns. In contrast, the features learned by UCL strategies are perceptually relevant and robust to catastrophic forgetting, with \lump learning the most distinctive features. Similar to $\mathcal{T}_0$, we observe that the UCL features are more relevant and distinguishable than SCL for $\mathcal{T}_{13}$. Note that we randomly selected the examples and feature maps for all visualizations.

\begin{figure*}[h!]
\begin{minipage}[t]{1.02\linewidth}
\vspace{-0.1in}
\resizebox{\linewidth}{!}{%
\begin{tabular}{cccccc}
\small
\hspace{-0.15in}\rowlegend{~~~~$\mathcal{T}_0$}
    \includegraphics[width=0.15\textwidth]{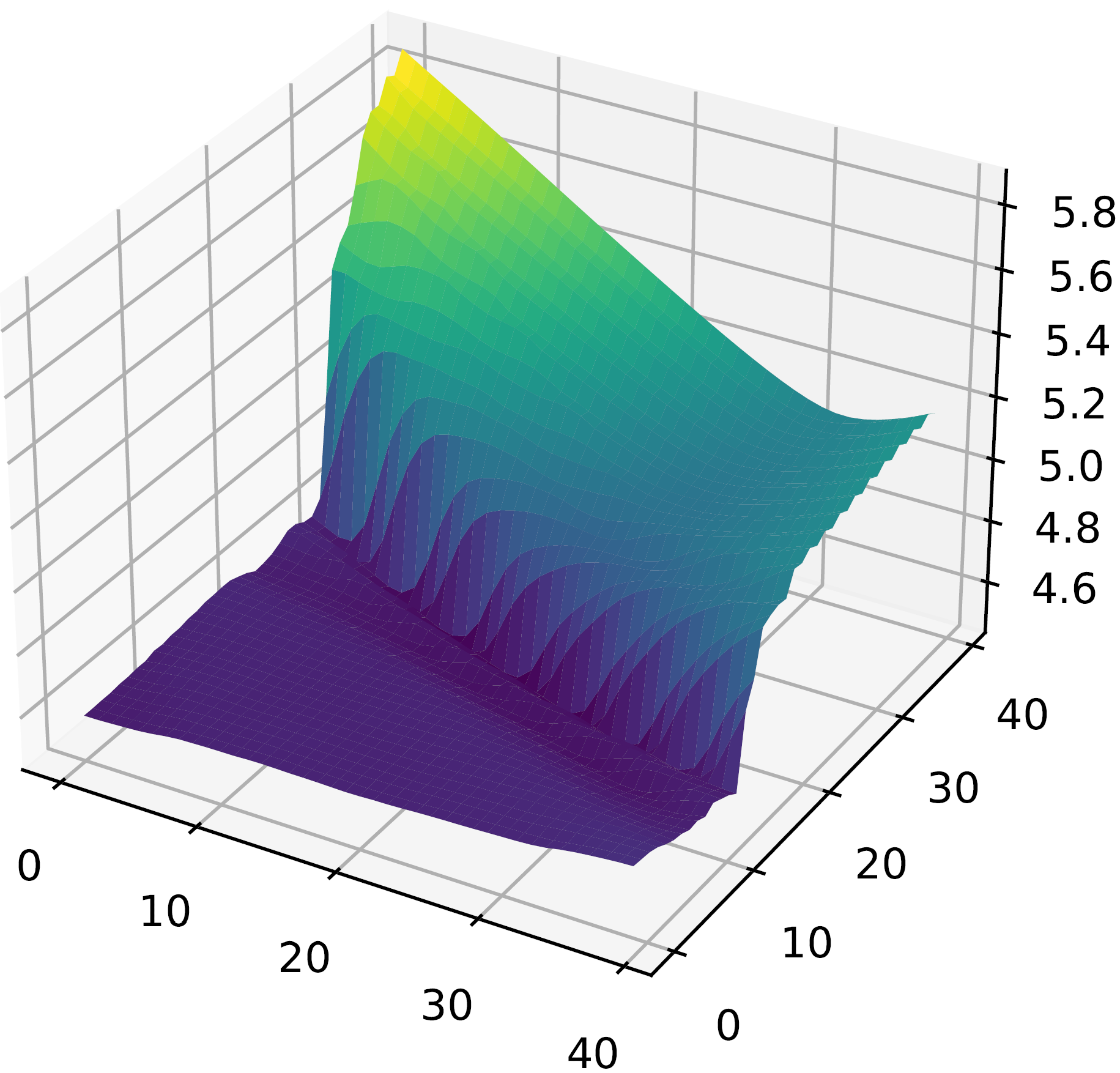} & \hspace{-0.15in}
    \includegraphics[width=0.15\textwidth]{figures/images/sclsi-t0.png} & \hspace{-0.15in}
    \includegraphics[width=0.15\textwidth]{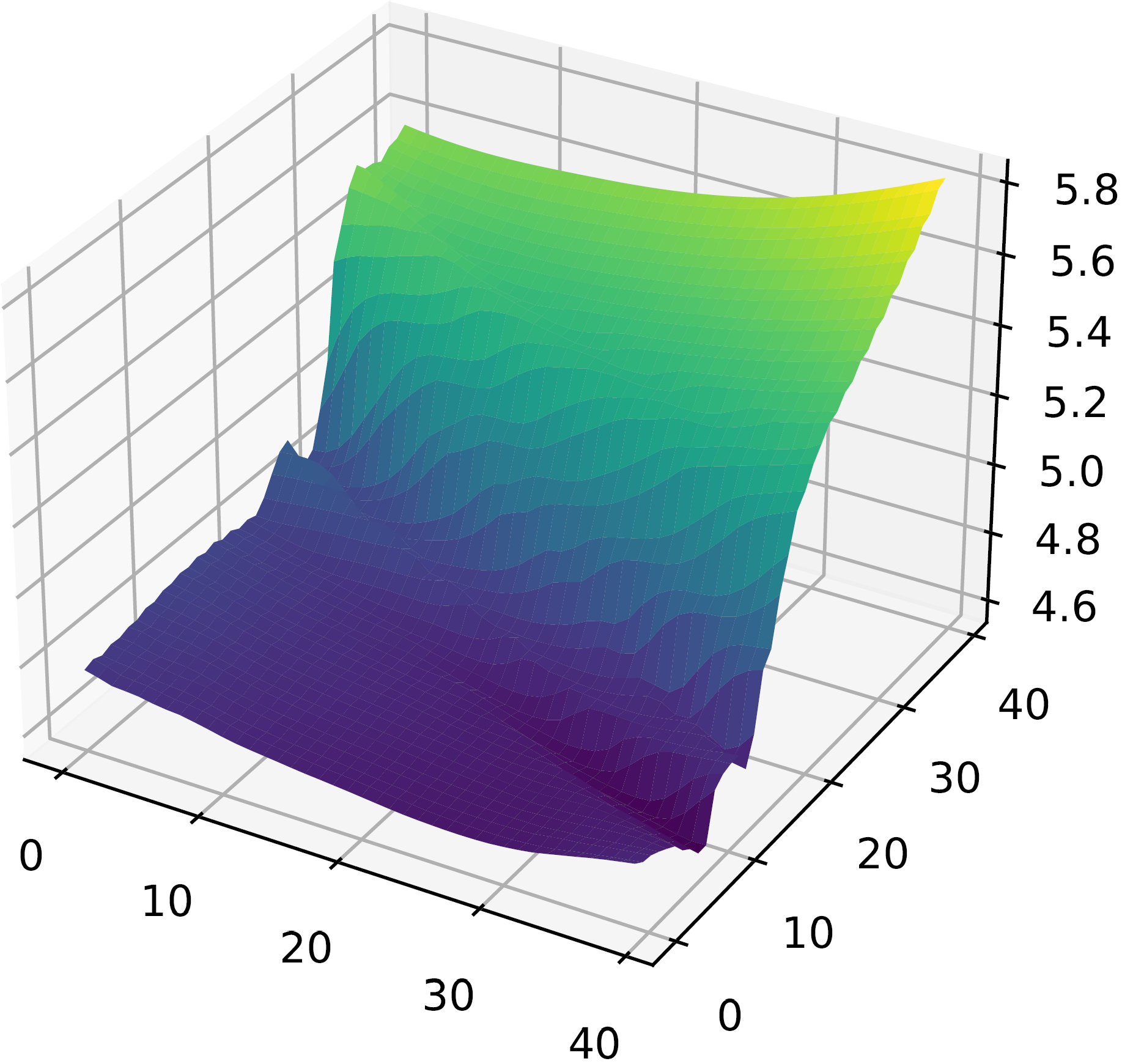} & \hspace{-0.15in}
    \includegraphics[width=0.15\textwidth]{figures/images/uclfinetune-t0.png} & \hspace{-0.15in}
    \includegraphics[width=0.15\textwidth]{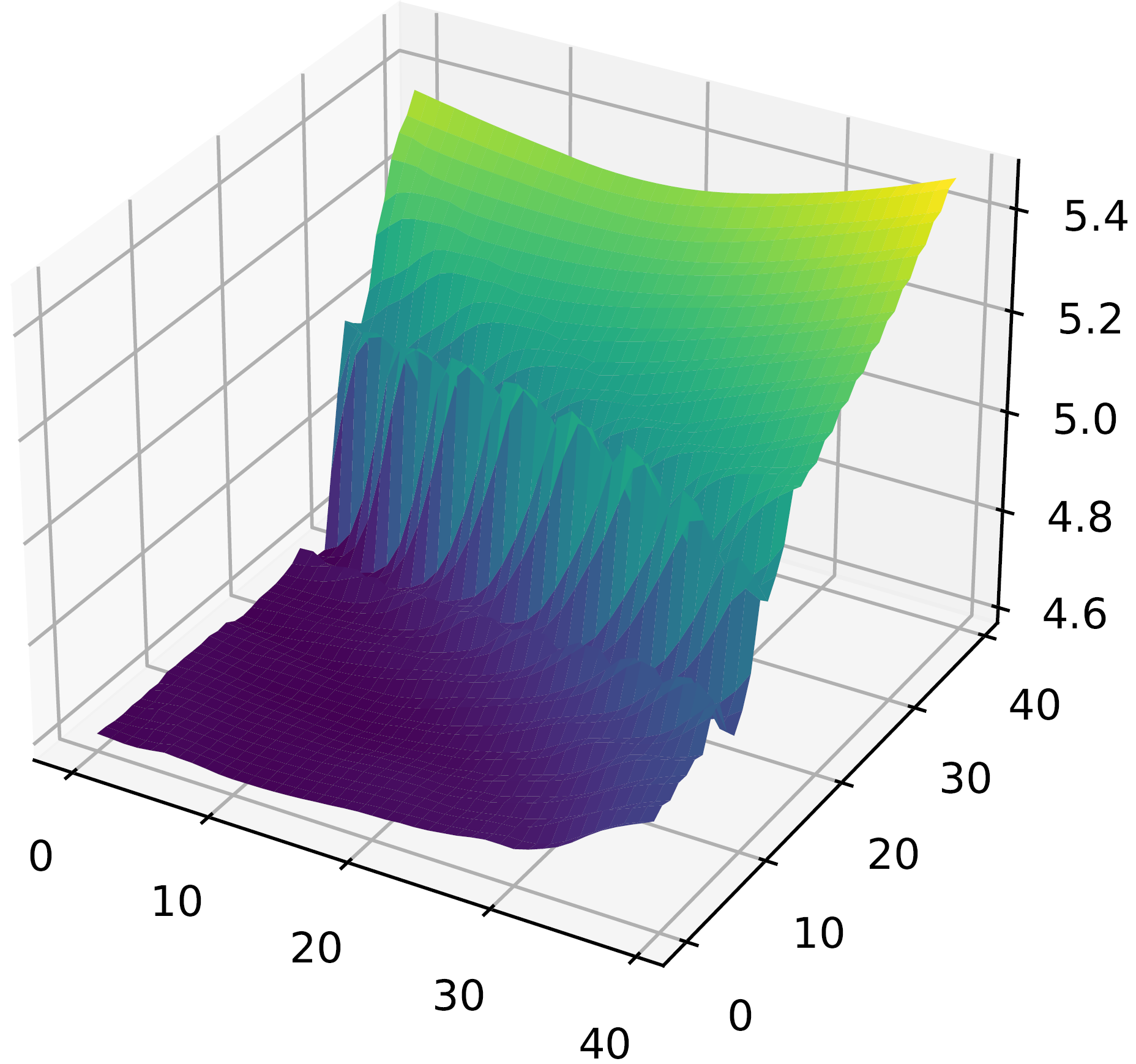} & \hspace{-0.15in}
    \includegraphics[width=0.15\textwidth]{figures/images/uclmixup-t0.png} \\
\hspace{-0.15in}\rowlegend{~~~~$\mathcal{T}_{19}$}    
    \includegraphics[width=0.15\textwidth]{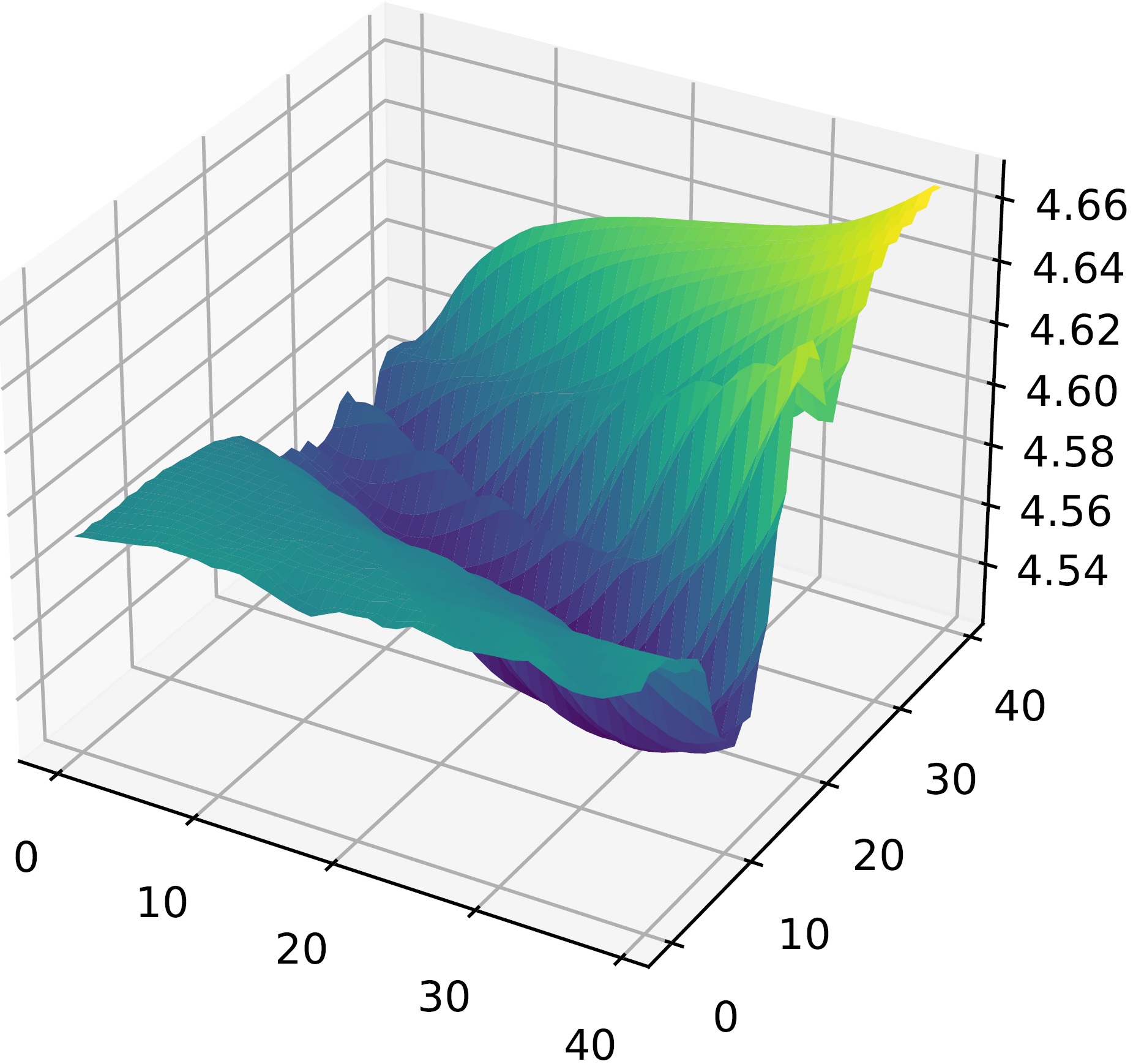} & \hspace{-0.15in}
    \includegraphics[width=0.15\textwidth]{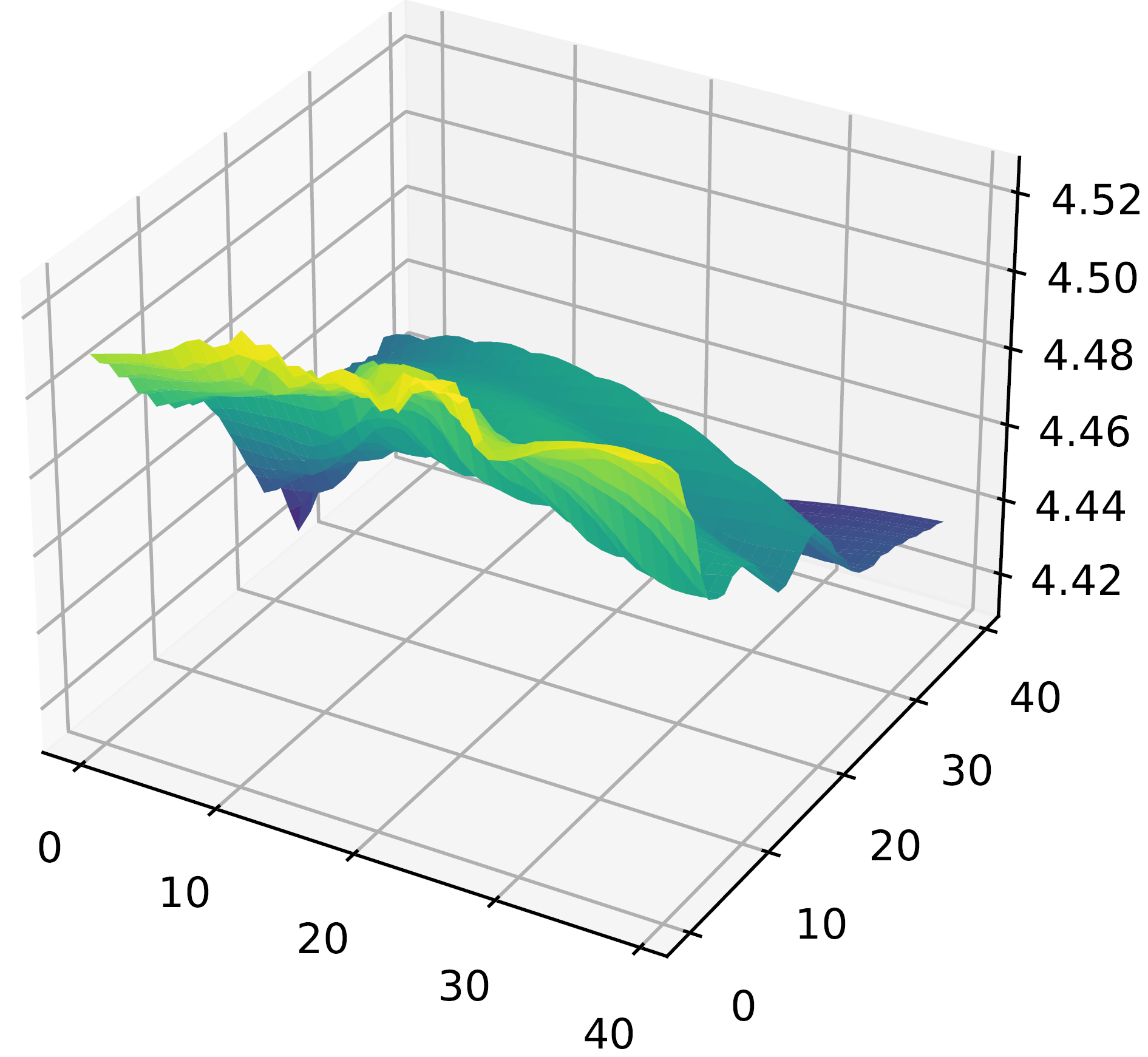} & \hspace{-0.15in}
    \includegraphics[width=0.15\textwidth]{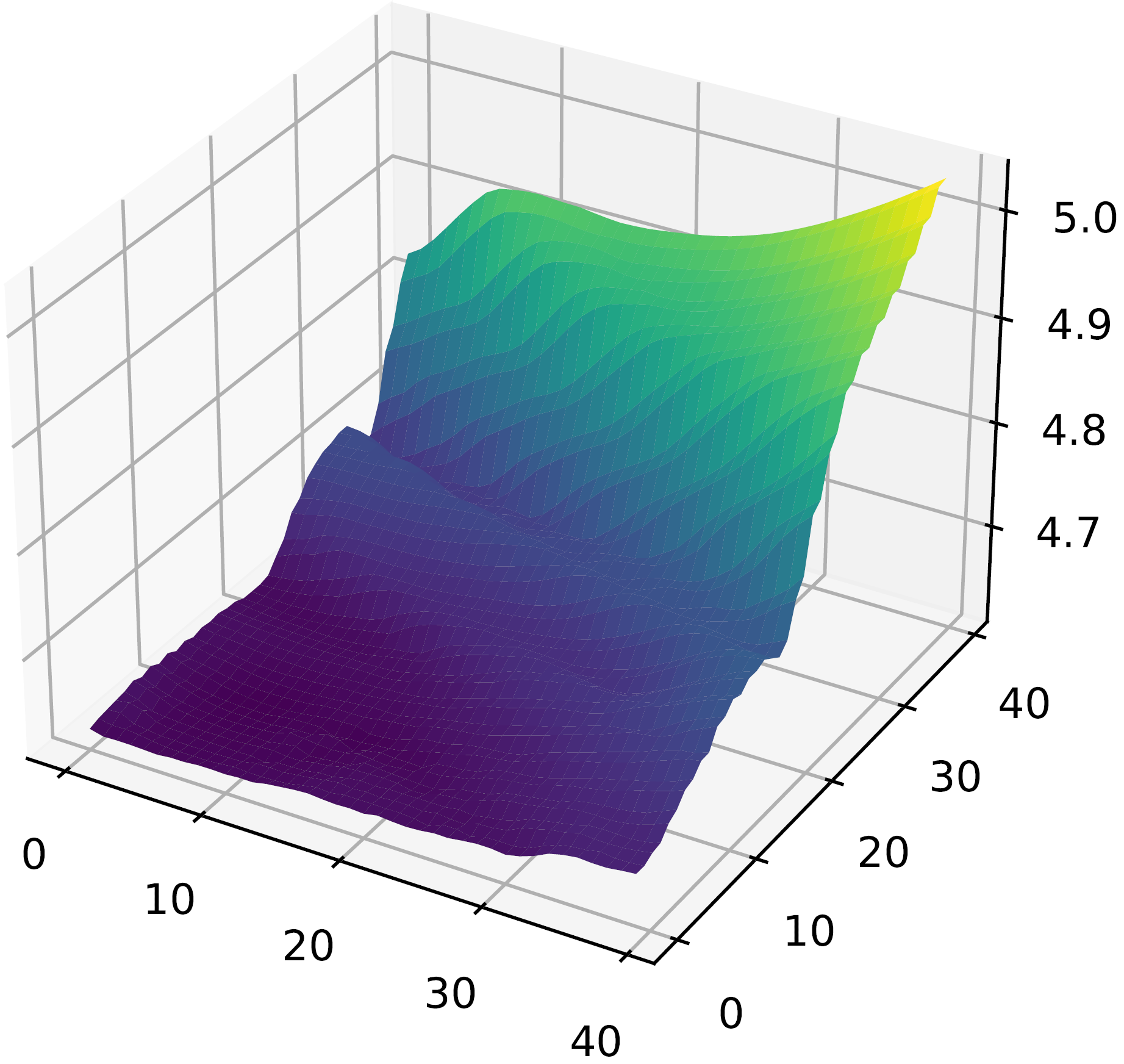} & \hspace{-0.15in}
    \includegraphics[width=0.15\textwidth]{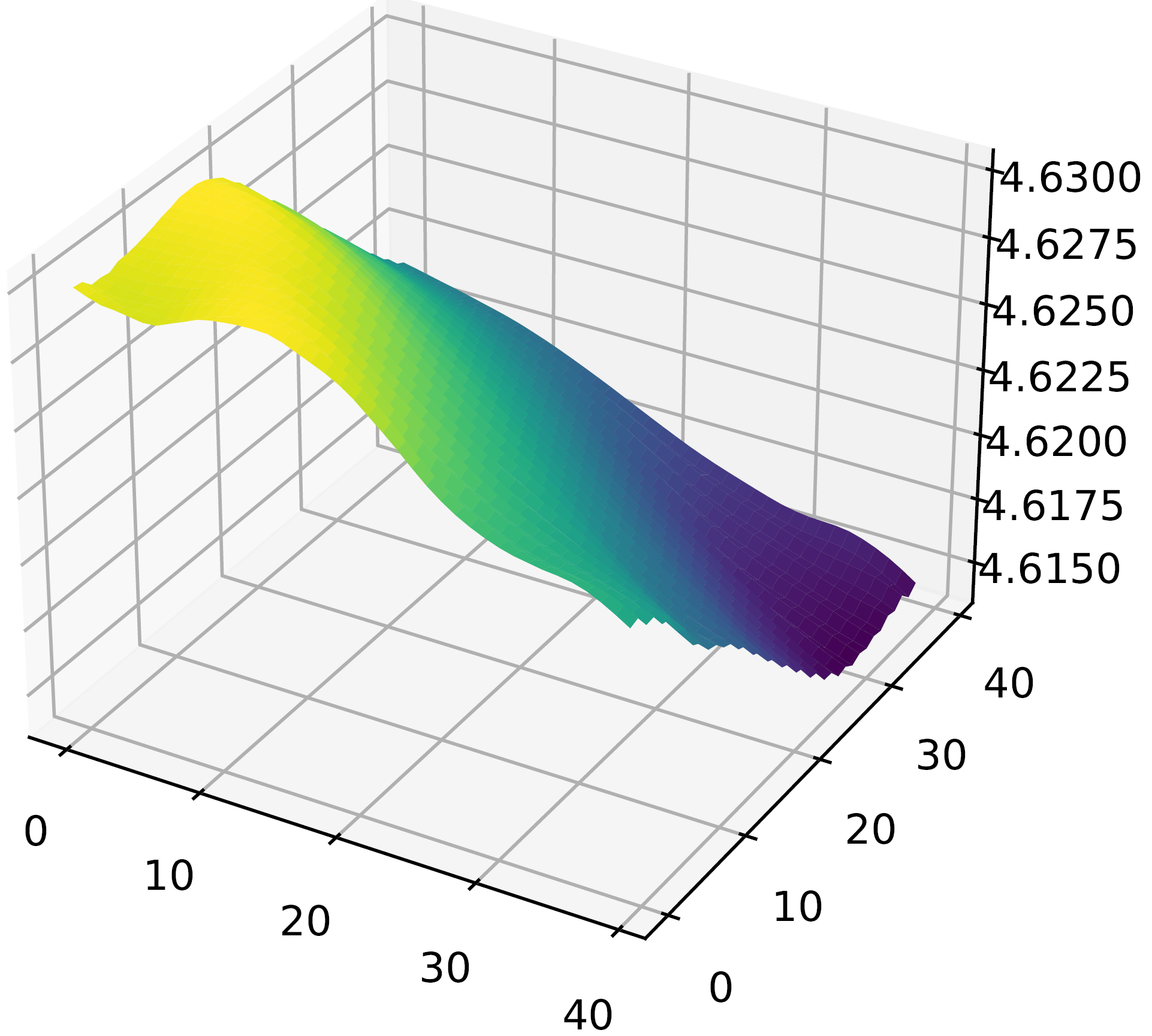} & \hspace{-0.15in}
    \includegraphics[width=0.15\textwidth]{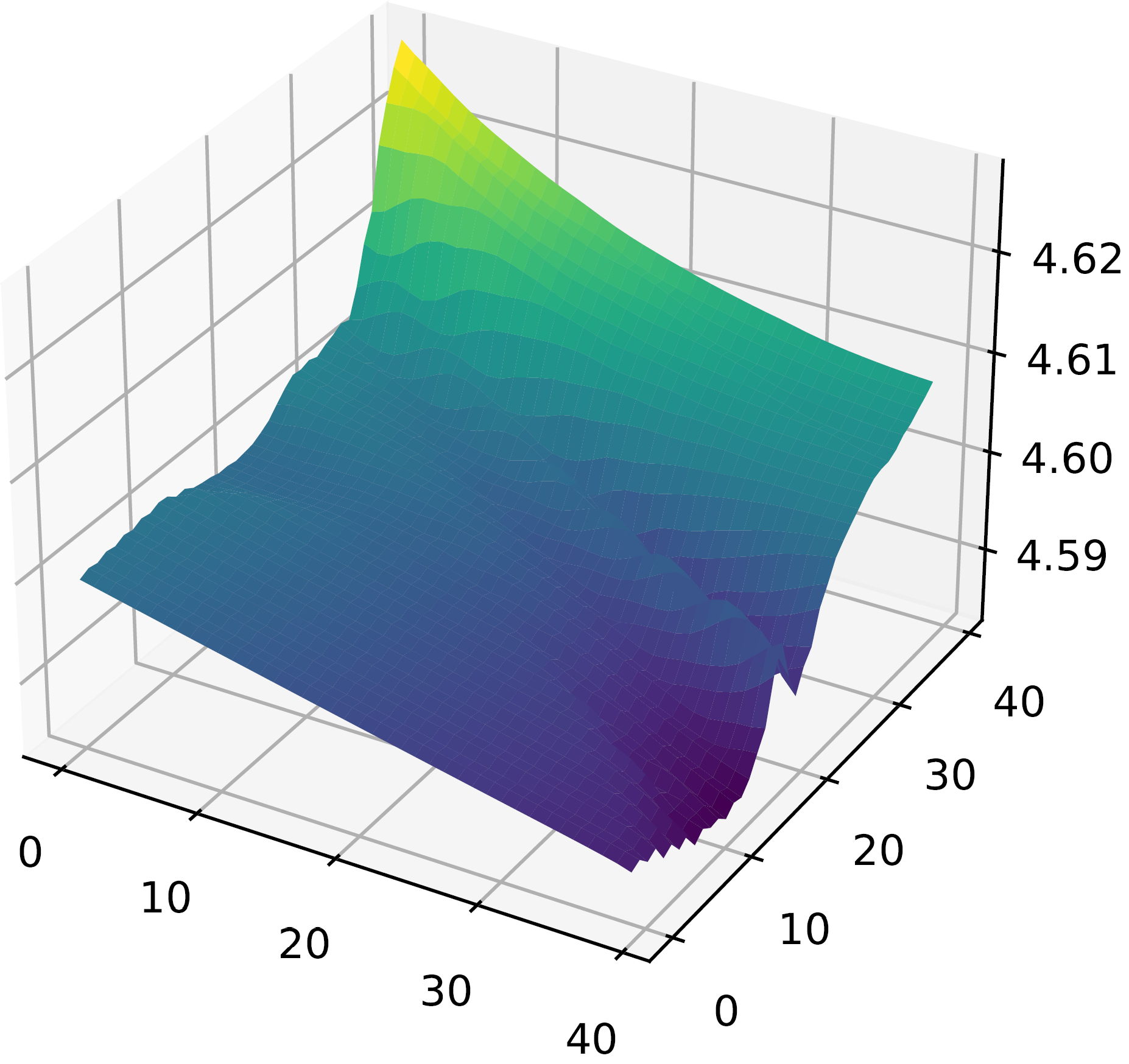} & \hspace{-0.15in}
    \includegraphics[width=0.15\textwidth]{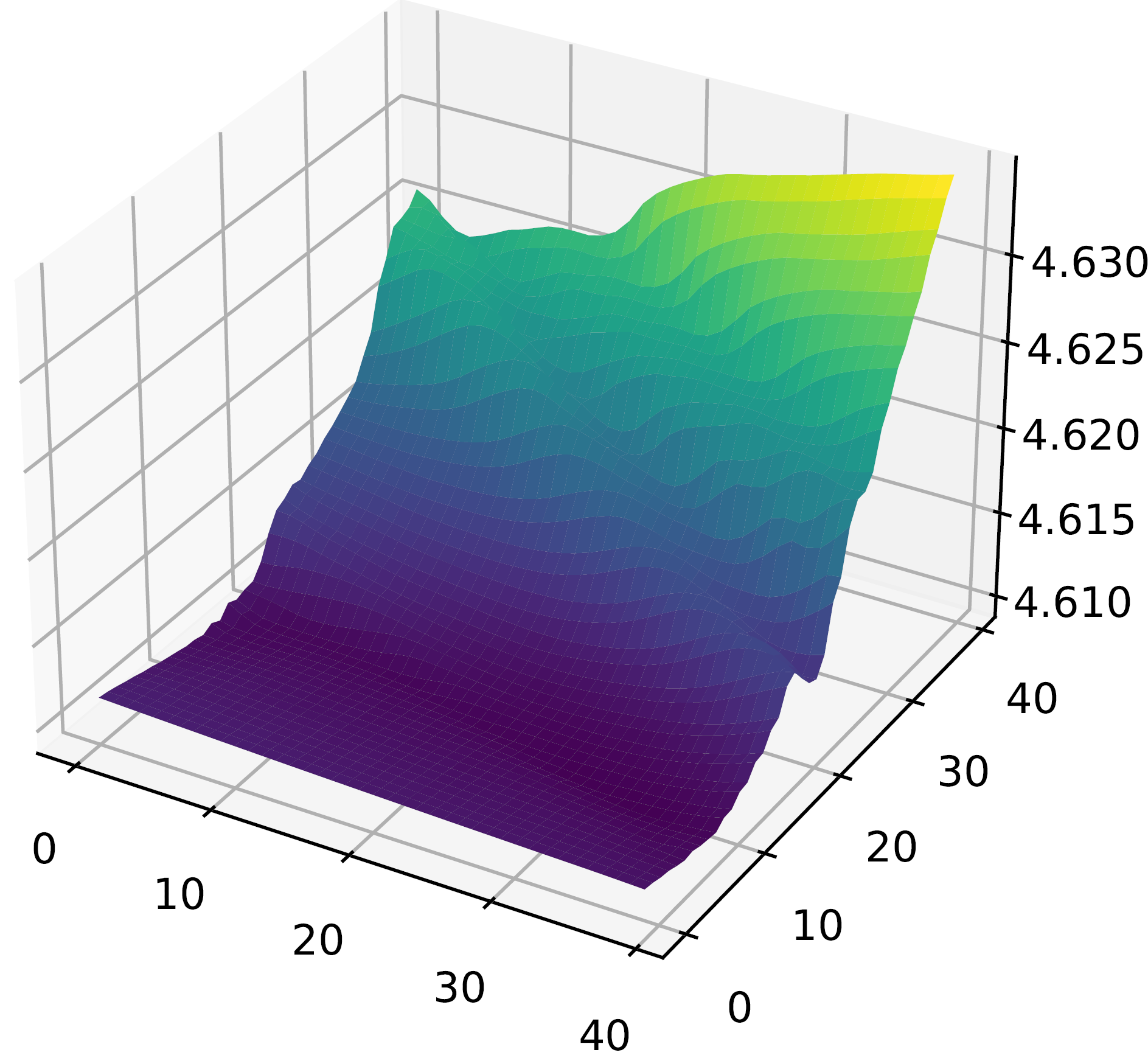} \\
\hspace{-0.15in}\specialcell{\scriptsize \textsc{Scl-Finetune}\\{\scriptsize Acc:$54.73\pm 0.25$}} & \hspace{-0.15in} \specialcell{\scriptsize \textsc{Scl-Si}\\{\scriptsize Acc: $58.59\pm 0.20$}} & \hspace{-0.15in} \specialcell{\scriptsize \textsc{Scl-Der}\\{\scriptsize Acc: $73.13\pm 0.38$}} & \hspace{-0.15in}
\specialcell{\scriptsize \textsc{Ucl-Finetune}\\{\scriptsize Acc: $70.80\pm 0.40$}} & \hspace{-0.15in} \specialcell{\scriptsize \textsc{Ucl-Si}\\{\scriptsize Acc: $76.39\pm 1.56$}} &  \hspace{-0.15in} 
\specialcell{\scriptsize \textsc{Lump (Ours)}\\{\scriptsize Acc:$ 76.60\pm 2.70$}}\\
\end{tabular}}
\vspace{-0.1in}
\captionof{figure}{\small {\bf Loss landscape visualization} of $\mathcal{T}_0$ after the completion of training on task $\mathcal{T}_0$ ({\bf top}) and $\mathcal{T}_{19}$ ({\bf bottom}) for Split CIFAR-100 dataset on ResNet-18 architecture. We use Simsiam for UCL methods. \label{fig:loss_vis}}
\vspace{-0.2in}
\end{minipage}
\end{figure*}

{\bf Loss landscape visualization. } To gain further insights, we visualize the loss landscape of task $\mathcal{T}_0$ after the completion of training on task $\mathcal{T}_0$ and $\mathcal{T}_{19}$ for various UCL and SCL strategies in \cref{fig:loss_vis}. We measure the cross-entropy loss for all methods with a randomly initialized linear classifier for a fair evaluation of two different directions. We use the visualization tool from \cite{li2017visualizing} that searches the task loss surface by repeatedly adding random perturbations to model weights. We observe that the loss landscape after $\mathcal{T}_0$ looks quite similar across all the strategies since the forgetting does not exist yet. However, after training $\mathcal{T}_{19}$, there is a clear difference with the UCL strategies obtaining a flatter and smoother loss landscape because UCL methods are more stable and robust to the forgetting, which hurts the loss landscapes of past tasks for SCL. It is important to observe that \lump obtains a smoother landscape than other UCL strategies, demonstrating its effectiveness. We defer further analyses for feature and loss landscape visualization to \cref{sec:additional_experiments}.

\begin{table*}[t!]
\begin{minipage}[t!]{0.49\linewidth}
        \captionof{table}{\footnotesize $\ell_2$ distance between UCL parameters after completion of training. \label{tab:l2_u}}
        \vspace{-0.1in}
    \resizebox{\linewidth}{!}{%
    \begin{tabular}{lcccc} 
    \toprule
     \textsc{Model}  & \textsc{Finetune} & \textsc{Si} & \textsc{Der} & \textsc{MultiTask}\\ \midrule
       \textsc{Finetune} & {60.00} \scriptsize($\pm$ 1.70) &  &  & \\
       \textsc{Si}  & {76.46} \scriptsize($\pm$ 0.48) &  {92.35} \scriptsize($\pm$ 0.61) & &  \\ 
        \textsc{der} & {55.60} \scriptsize($\pm$ 1.42)  & {75.54} \scriptsize($\pm$ 0.97) &  {48.76} \scriptsize($\pm$ 1.54) & \\ 
\textsc{Multitask} & \small{61.32} \scriptsize($\pm$ 0.59) & \small{79.95} \scriptsize($\pm$ 0.40) &  \small{57.90} \scriptsize($\pm$ 0.86) &  \small{61.42} \scriptsize($\pm$ 0.78 )\\ 
 \bottomrule
    \end{tabular}}
\end{minipage}
\hfill
\begin{minipage}[t!]{0.49\linewidth}
 \captionof{table}{\footnotesize $\ell_2$ distance between SCL paraneters after completion of training. \label{tab:l2_s}}
               \vspace{-0.1in}
    \resizebox{\linewidth}{!}{%
    \begin{tabular}{lcccc} 
    \toprule
     \textsc{Model}  & \textsc{Finetune} & \textsc{Si} & \textsc{Der} & \textsc{MultiTask}\\ \midrule
       \textsc{Finetune} & {183.31} \scriptsize($\pm$ 0.10) &  &  & \\
       \textsc{Si}  & {206.16} \scriptsize($\pm$ 0.28) &  {226.05} \scriptsize($\pm$ 0.13) & &  \\ 
        \textsc{der} & {202.61} \scriptsize($\pm$ 0.46)  & {224.78} \scriptsize($\pm$ 0.75) &  {219.06} \scriptsize($\pm$ 0.27) & \\ 
\textsc{Multitask} & \small{258.12} \scriptsize($\pm$ 0.26) & \small{277.30} \scriptsize($\pm$ 0.69) &  \small{271.48} \scriptsize($\pm$ 0.45) &  \small{314.84} \scriptsize($\pm$ 0.92)\\ 
 \bottomrule
\end{tabular}}
\end{minipage}
\vspace{-0.15in}
\end{table*}

\section{Discussion and Conclusion}
\vspace{-0.1in}
This work attempts to bridge the gap between unsupervised representation learning and continual learning. In particular, we establish the following findings for unsupervised continual learning.

{\bf Surpassing supervised continual learning.} Our empirical evaluation across various CL strategies and datasets shows that UCL representations are more robust to catastrophic forgetting than SCL representations. Furthermore, we notice that UCL generalizes better to OOD tasks and achieves stronger performance on few-shot learning tasks. We propose \emph{Lifelong unsupervised mixup (\textsc{Lump})}, which interpolates the unsupervised instances between the current task and past task and obtains higher performance with lower catastrophic forgetting across a wide range of tasks.

{\bf Dissecting the learned representations.} We conduct a systematic analysis to understand the differences between the representations learned by UCL and SCL strategies. By investigating the similarity between the representations, we observe that UCL and SCL strategies have high similarities in the lower layers but are dissimilar in the higher layers. We also show that UCL representations learn coherent and discriminative patterns and smoother loss landscape than SCL.

{\bf Limitations and future work.}
In this work, we do not consider the high-resolution tasks for CL. We intend to evaluate the forgetting of the learnt representations on ImageNet~\citep{deng2009imagenet} in future work, since UCL shows lower catastrophic forgetting and representation learning has made significant progress on ImageNet over the past years. In follow-up work, we intend to conduct further analysis to understand the behavior of UCL and develop sophisticated methods to continually learn unsupervised representations under various setups, such as class-incremental or task-agnostic CL.

\section*{Acknowledgements}
We thank the anonymous reviewers for their insightful comments and suggestions.
This work was supported by Microsoft Research Asia, the Engineering Research Center Program through the National Research Foundation of Korea (NRF) funded by the Korean Government MSIT (NRF-2018R1A5A1059921), Institute of Information \& communications Technology Planning \& Evaluation (IITP) grant funded by the Korea government (MSIT)  (No.2019-0-00075, Artificial Intelligence Graduate School Program (KAIST) and 2021-0-01696). Any opinions, findings, and conclusions or recommendations expressed in this material are those of the authors and do not necessarily reflect the views of the funding agencies.

\section*{Author Contributions}
Divyam Madaan conceived of the presented idea, developed the experimental framework, carried out OOD evaluation, CKA visualization and took the lead in writing the manuscript. Jaehong Yoon performed the hyperparameter search, carried out the visualization of loss landscape and feature maps and performed the few-shot training analysis. Yuanchun Li, Yunxin Liu, and Sung Ju Hwang supervised the project.

\bibliography{references}
\bibliographystyle{iclr_2022}

\newpage
\appendix
\renewcommand\thefigure{\thesection.\arabic{figure}}    
\renewcommand\thetable{\thesection.\arabic{table}}
\section{Supplementary Material}
{\bf Organization.} In the supplementary material, we provide the implementation details followed by the hyper-parameter configurations in \cref{sec:experimental_details}. Further, we show the other experiments we conducted and additional visualizations and results in \cref{sec:additional_experiments}.

\subsection{Experimental Details} \label{sec:experimental_details}

{\bf Implementations.} We use the DER~\citep{buzzega2020dark} open-source codebase\footnote{\url{https://github.com/aimagelab/mammoth}} for all the experiments. In particular, we reproduce all their experimental results for supervised continual learning and use various models with their set of hyper-parameters as our baselines. We follow the original representations for SimSiam\footnote{\url{https://github.com/facebookresearch/simsiam}} and BarlowTwins\footnote{\url{https://github.com/facebookresearch/barlowtwins}} for unsupervised continual learning. We verify our implementation by reproducing the reported results on CIFAR-10 in the original paper, where we train the representations on the complete CIFAR-10 dataset and evaluate on the test-set using KNN classifier~\citep{wu18knn}. In particular, \citep{wu18knn} stores the features for each instance in the task-level training set in a discrete memory bank. The optimal feature-level embeddings are then learned by instance-level discrimination, which maximally scatters the features of the training samples. Following prior works in representation learning, we use the task-level training set without any augmentation in the task-incremental setup for the supervised and unsupervised KNN evaluation.

{\bf Hyperparameter configurations.} We use the tuned hyper-parameters reported by \cite{buzzega2020dark} for all the SCL experiments. On the other hand, we tune the hyper-parameters
for continual learning strategies for UCL. We provide the hyper-parameters setup for UCL for different datasets in \cref{tab:hyperparameters}. We train all the UCL methods with a batch size of $256$ for $200$ epochs, while training the SCL methods with a batch size of $32$ for $50$ epochs following \cite{buzzega2020dark}. We observed that training the SCL methods further lead to a degredation in performance for all the methods. We use the same set of augmentations for both SCL and UCL except that we use \texttt{RandomResizedCrop} with scale in $[0.2, 1.0]$ for UCL~\citep{wu18knn, chen2020exploring} and \texttt{RandomCrop} for SCL. For rehearsal-based methods,  we use the buffer size $200$ for Split CIFAR-10, Split CIFAR-100 and $256$ for Split Tiny-ImageNet dataset. We use a learning rate of $0.03$ for SGD optimizer with weight decay $5$e-$4$ and momentum $0.9$.

\begin{table}[H]
\caption{{\bf Hyperparameter configurations} for all the datasets on ResNet-18 architecture. \label{tab:hyperparameters}}
\vspace{-0.1in}
\centering\begin{tabular}{llll}
\toprule
\textsc{Method} & \textsc{Split CIFAR-10} & \textsc{Split CIFAR-100} & \textsc{Seq. Tiny-ImageNet} \\
\midrule
\textsc{Si}      & ~~ \textit{c}~$:100$  ~~ \textit{$\xi:1$} & ~~ \textit{c}~$:0.1$  ~~ \textit{$\xi:1$}&~~ \textit{c}~$:0.01$  ~~ \textit{$\xi:1$} \\
\textsc{Pnn}     &~~  \textit{wd}~$:64$ & ~~  \textit{wd}~$:12$ & ~~  \textit{wd}~$:8$ \\
\textsc{DER}     &~~  \textit{$\alpha:0.1$}  & ~~ \textit{$\alpha:0.1$} & ~~ \textit{$\alpha:0.01$} \\
\textsc{Lump}   &  ~~ $\lambda:0.1$ & ~~ $\lambda:0.1$ & ~~ $ \lambda:0.4$\\
\bottomrule
\end{tabular}
\end{table}

\subsection{Additional Experiments} \label{sec:additional_experiments}
\begin{figure*}[t!]
\begin{minipage}[t]{1.05\linewidth}
\resizebox{\linewidth}{!}{%
\begin{tabular}{c cccc}
& $\mathcal{T}_0$ & $\mathcal{T}_{17}$ & $\mathcal{T}_{18}$ & $\mathcal{T}_{19}$\\ 
\begin{tabular}{c}
\textsc{Scl-Finetune}\\{ Acc: $54.73\pm 0.25$}\\~\\~\\~\\~\\~\\~\\~\\~
\end{tabular}&
\includegraphics[width=0.25\textwidth]{figures/images/sclfinetune-t0.png} &
\includegraphics[width=0.25\textwidth]{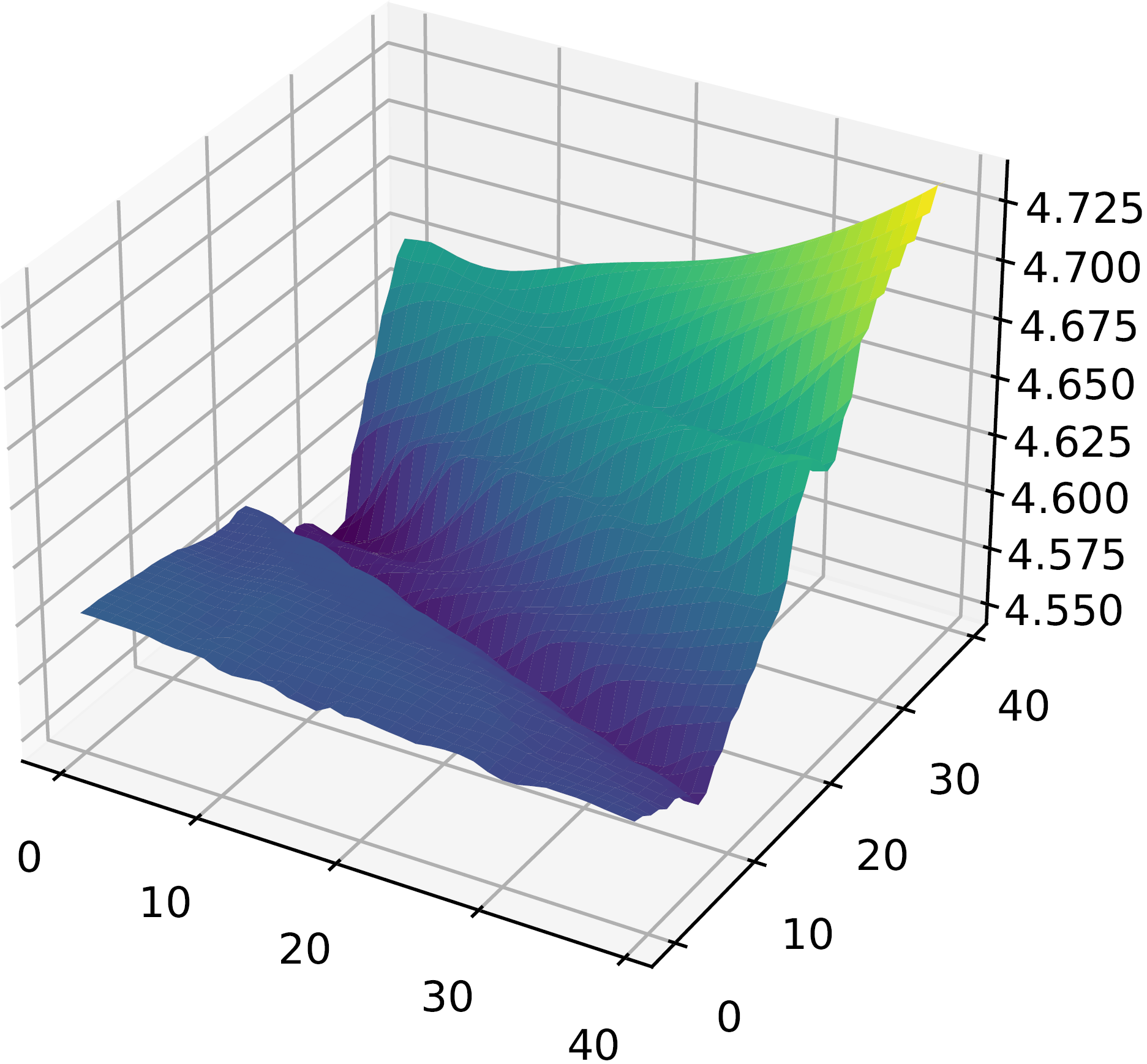} &
\includegraphics[width=0.25\textwidth]{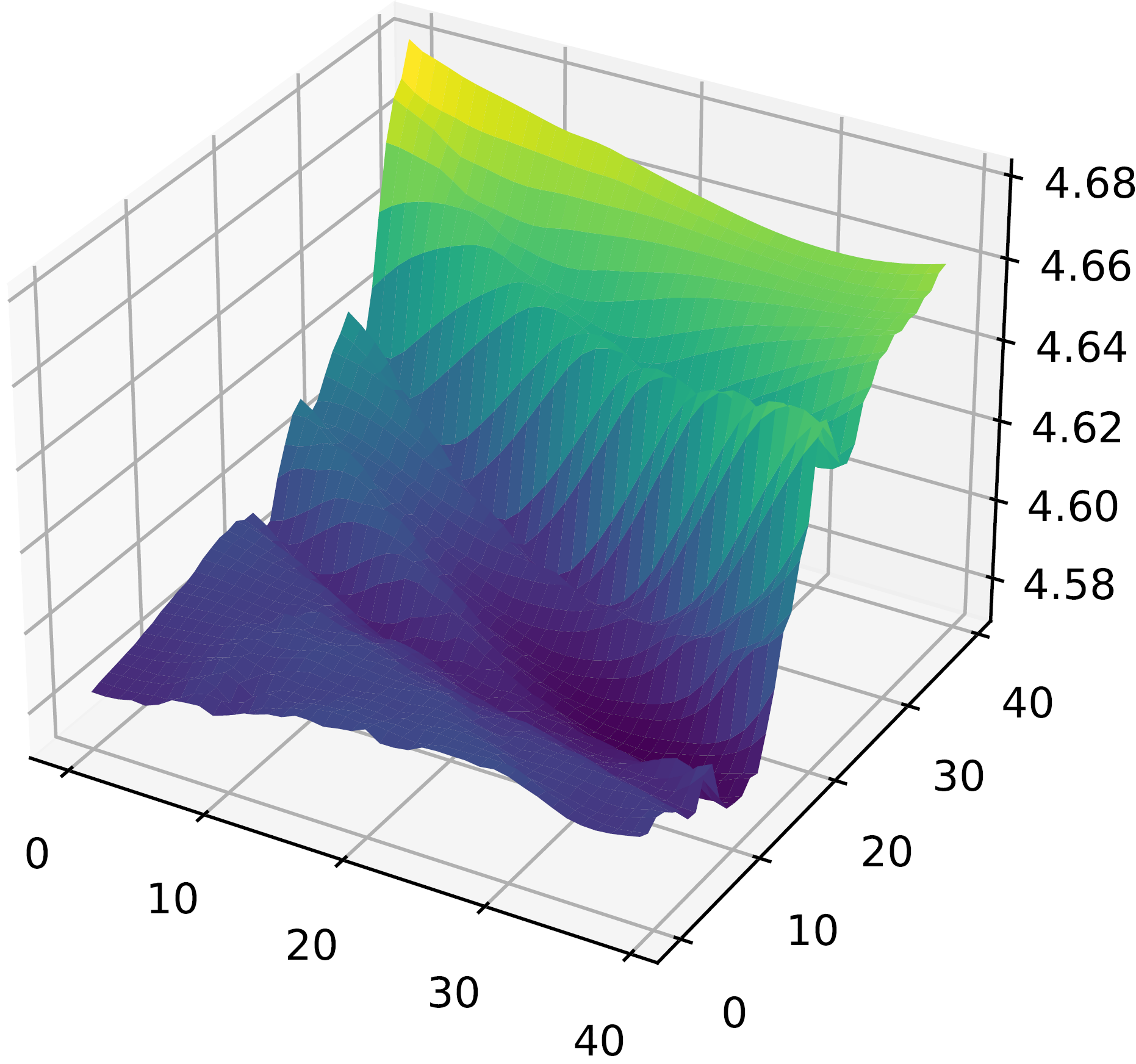} &
\includegraphics[width=0.25\textwidth]{figures/images/sclfinetune-t19.png} \vspace{-0.65in}\\
\begin{tabular}{c}
\textsc{Scl-Si}\\{ Acc: $58.59\pm 0.20$}\\~\\~\\~\\~\\~\\~\\~\\~
\end{tabular}&
\includegraphics[width=0.25\textwidth]{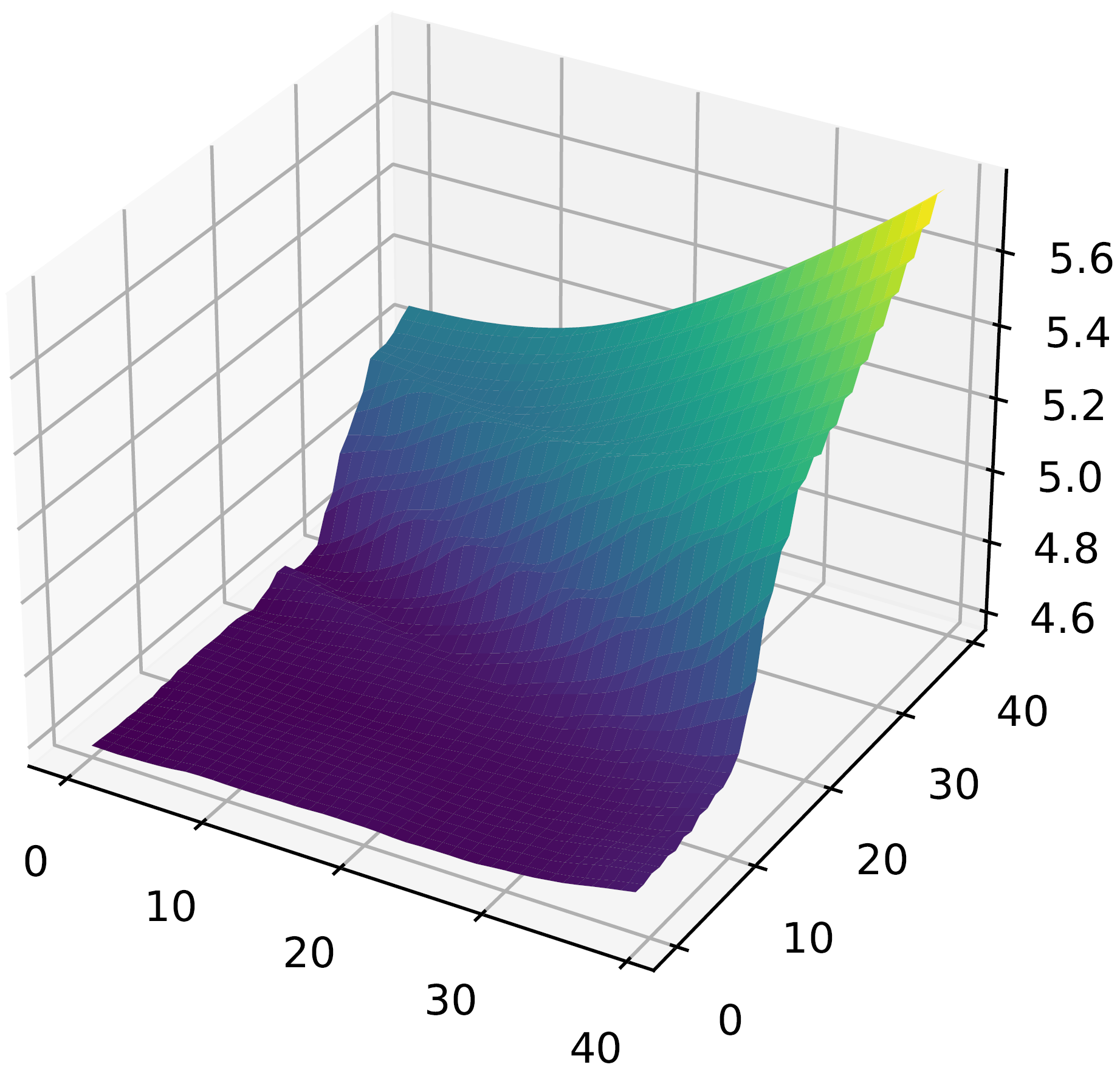} &
\includegraphics[width=0.25\textwidth]{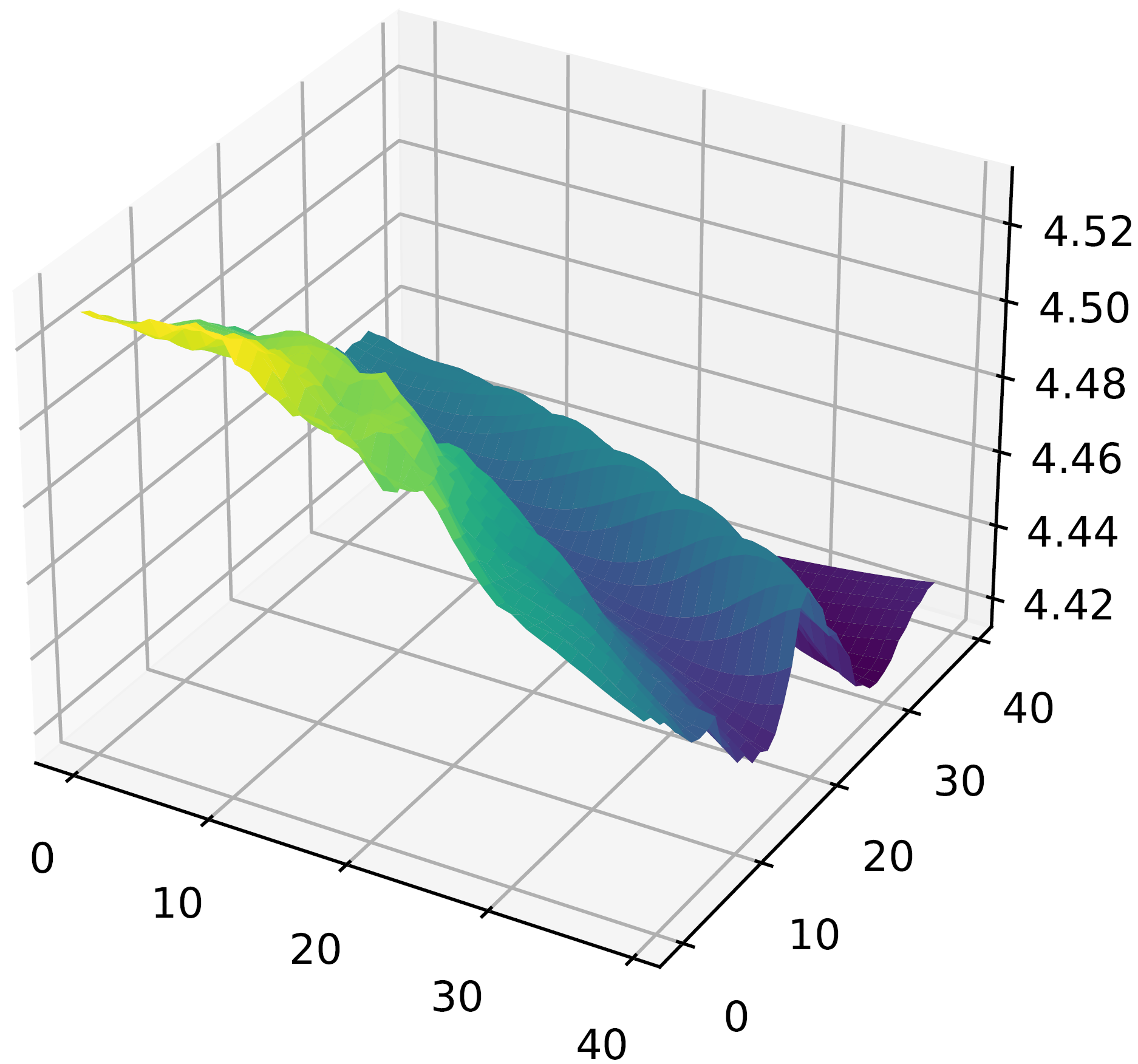} &
\includegraphics[width=0.25\textwidth]{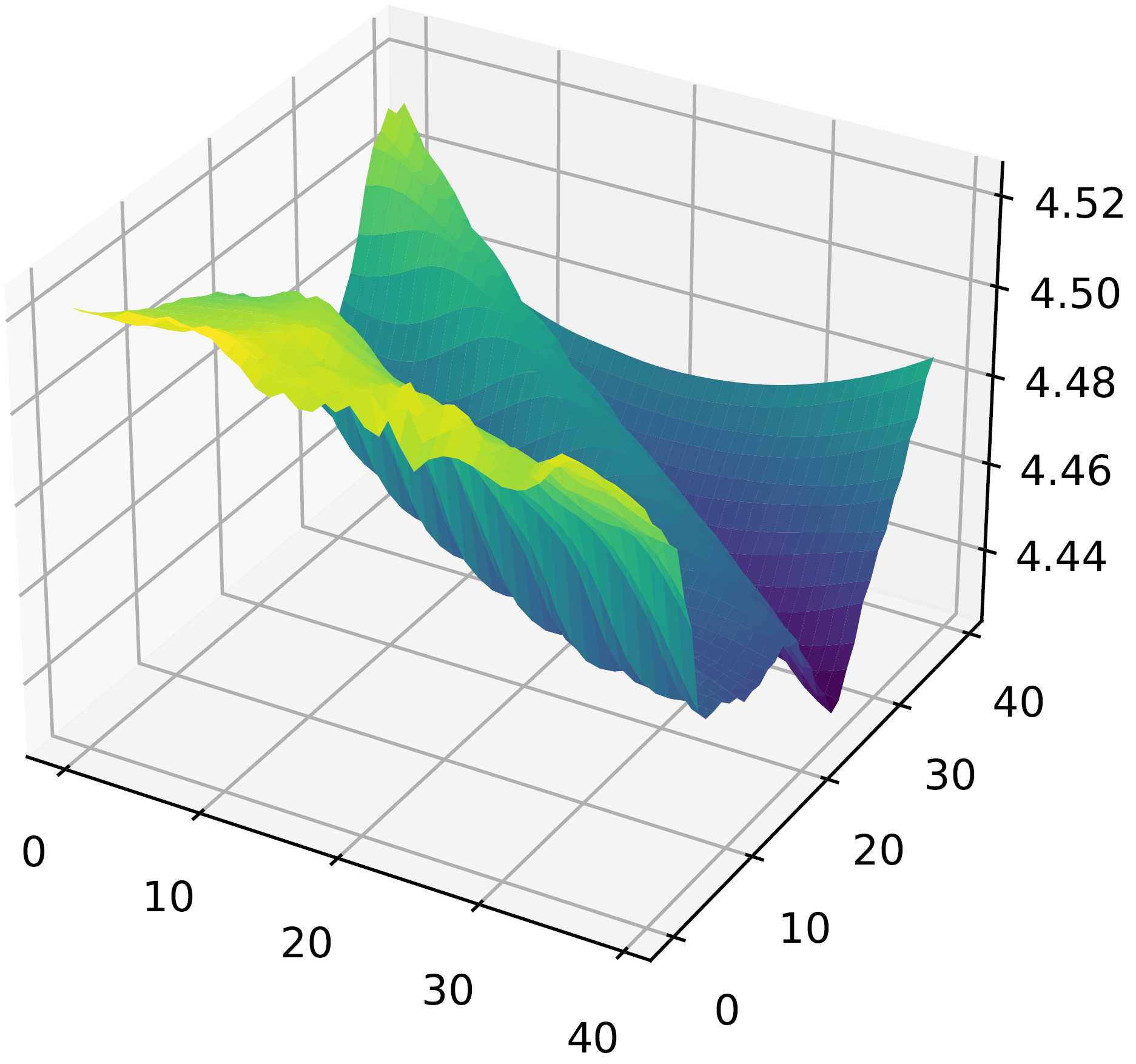} &
\includegraphics[width=0.25\textwidth]{figures/images/sclsi-t19.png} \vspace{-0.65in}\\
\begin{tabular}{c}
\textsc{Scl-Der}\\{ Acc: $73.13\pm 0.38$}\\~\\~\\~\\~\\~\\~\\~\\~
\end{tabular}&
\includegraphics[width=0.25\textwidth]{figures/images/sclder-t0.png} &
\includegraphics[width=0.25\textwidth]{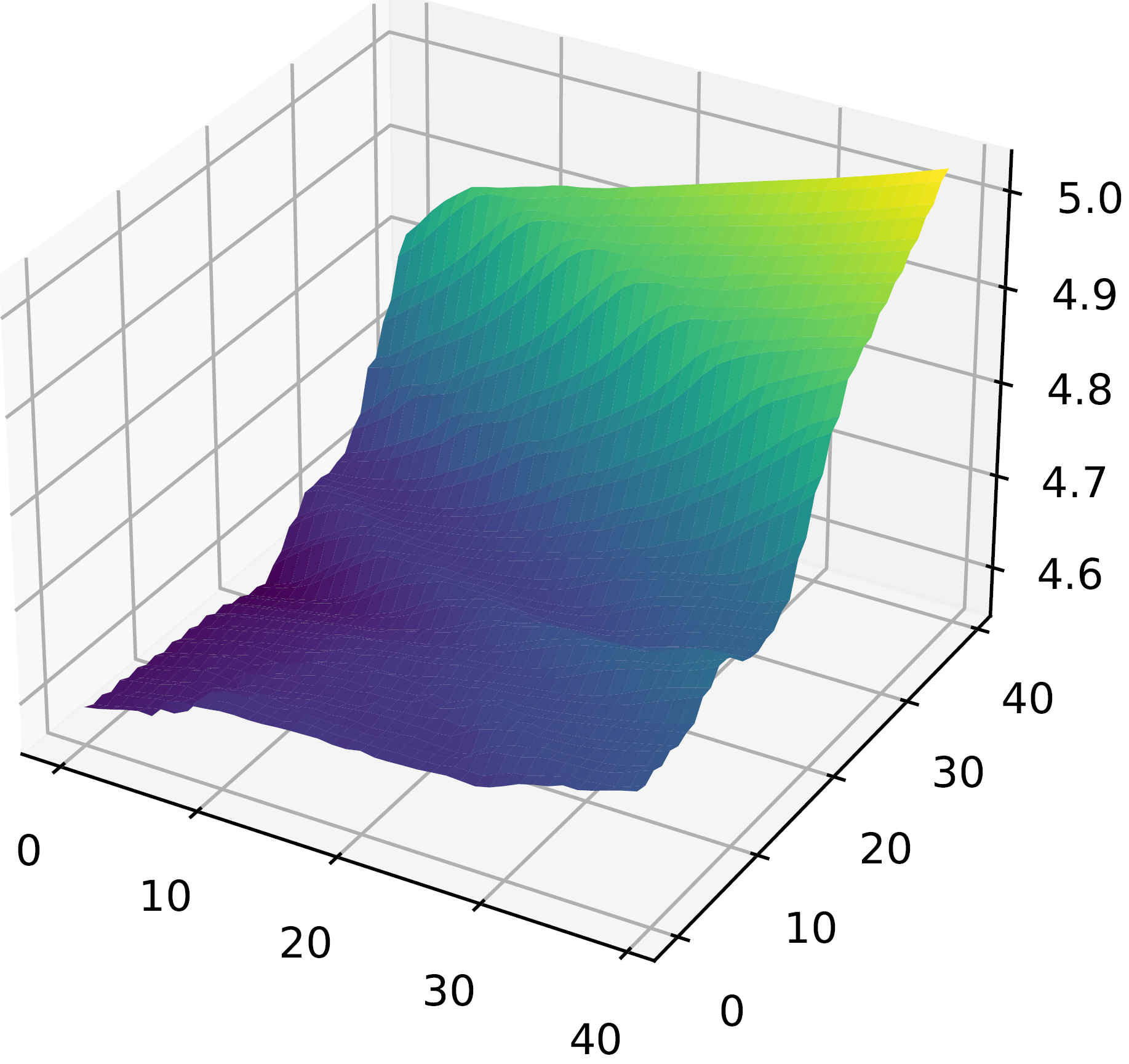} &
\includegraphics[width=0.25\textwidth]{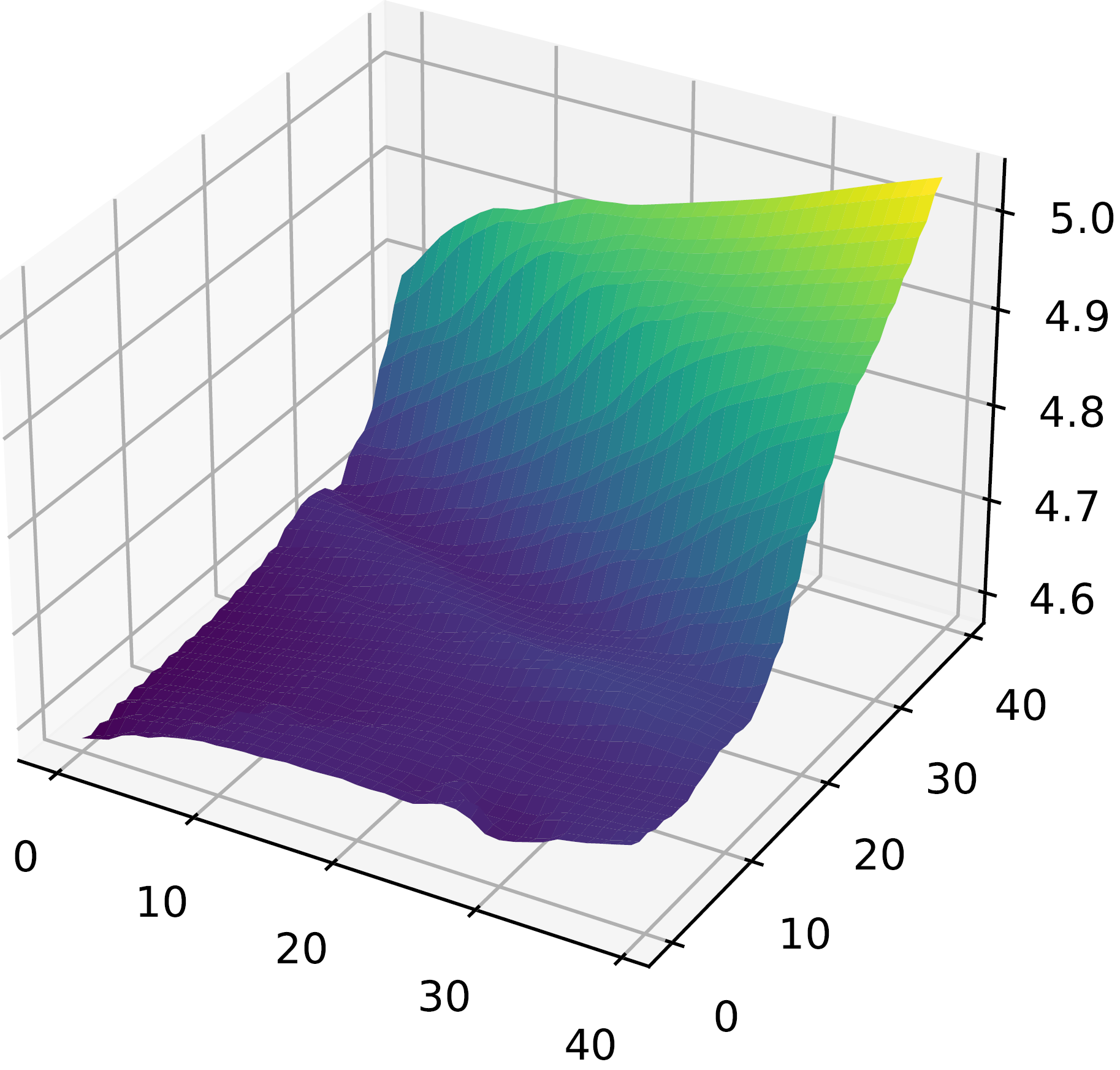} &
\includegraphics[width=0.25\textwidth]{figures/images/sclder-t19.png} \vspace{-0.65in}\\
\begin{tabular}{c}
\textsc{Ucl-Finetune}\\{ Acc: $70.80\pm 0.40$}\\~\\~\\~\\~\\~\\~\\~\\~
\end{tabular}&
\includegraphics[width=0.25\textwidth]{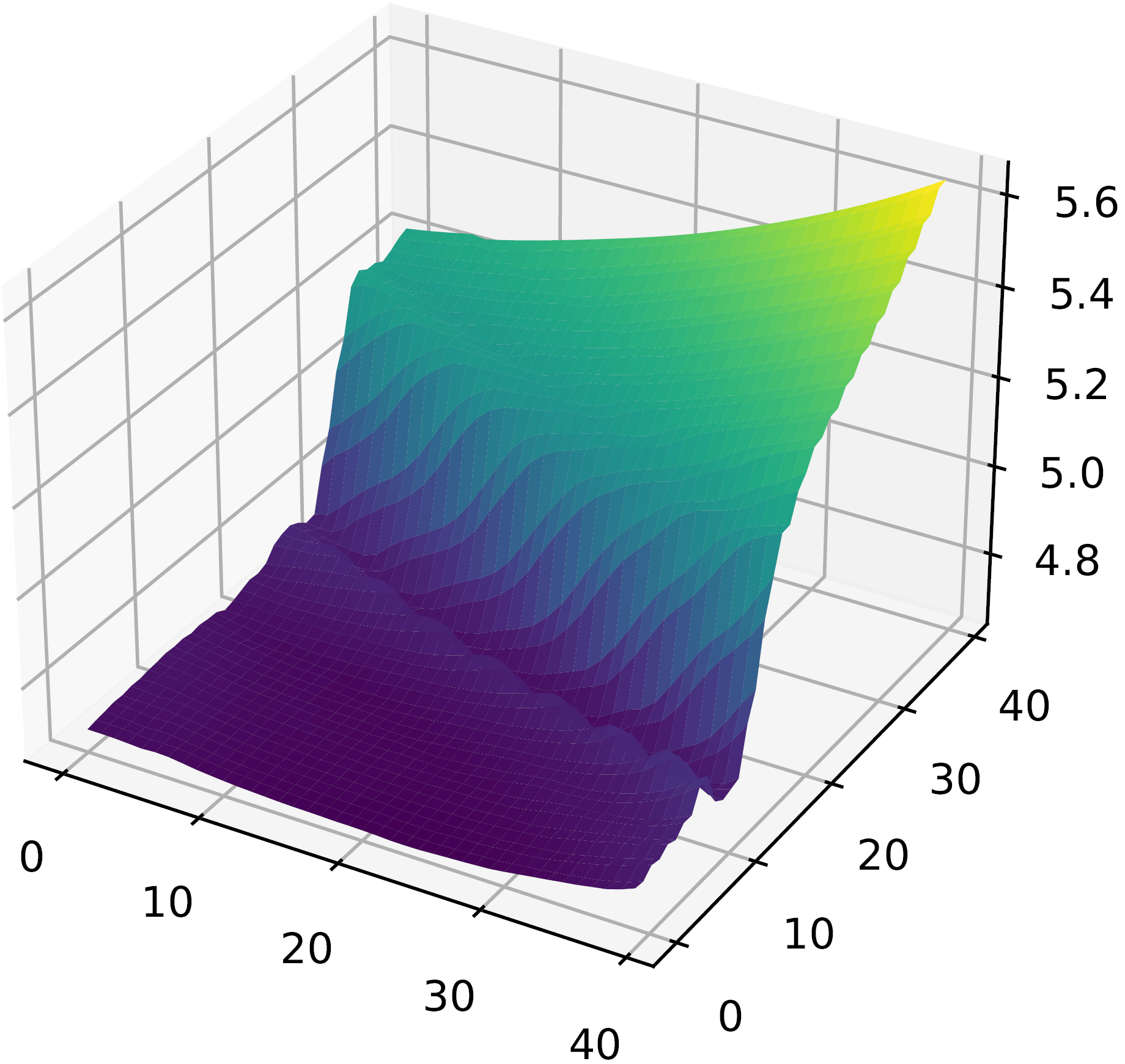} &
\includegraphics[width=0.25\textwidth]{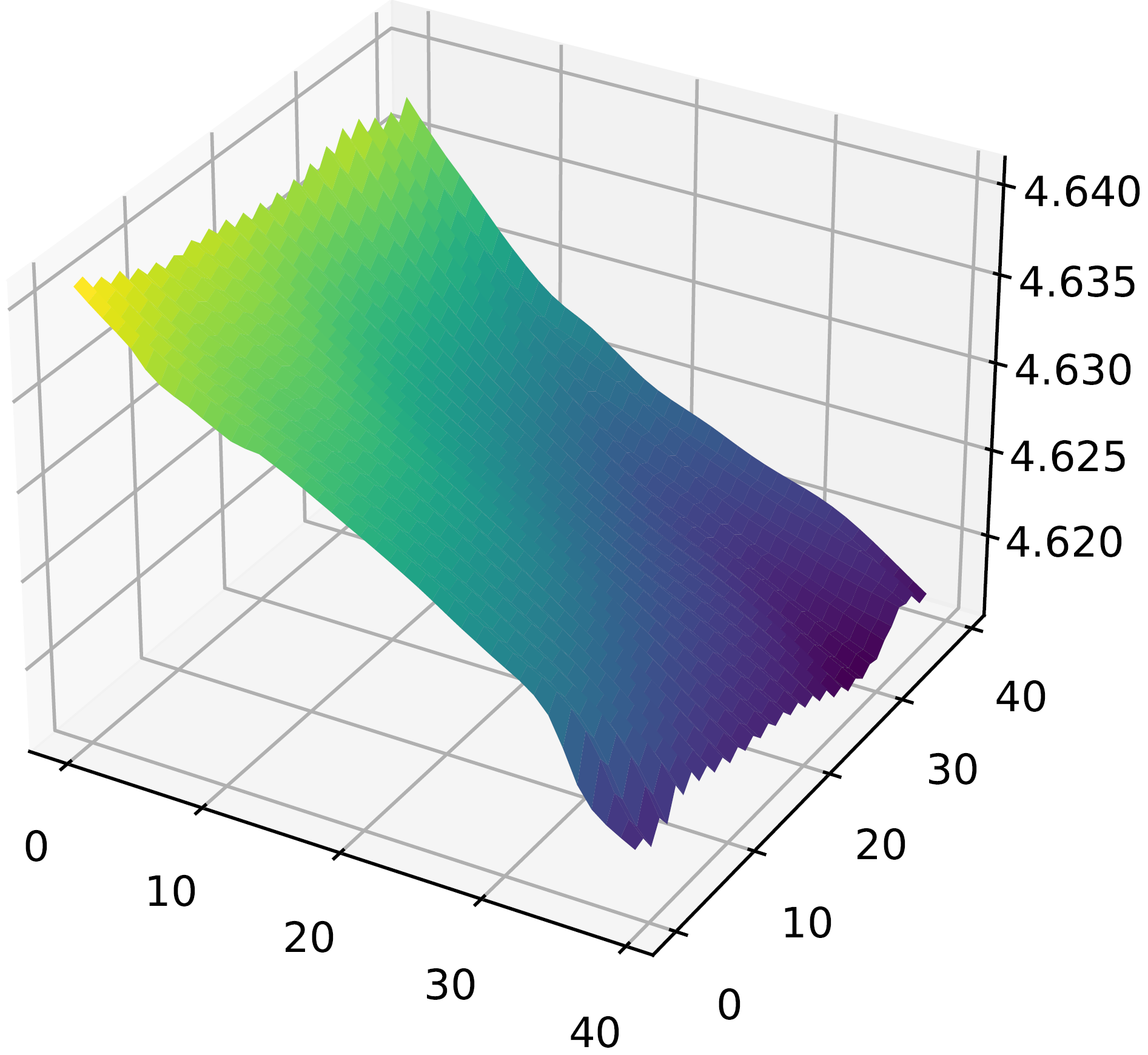} &
\includegraphics[width=0.25\textwidth]{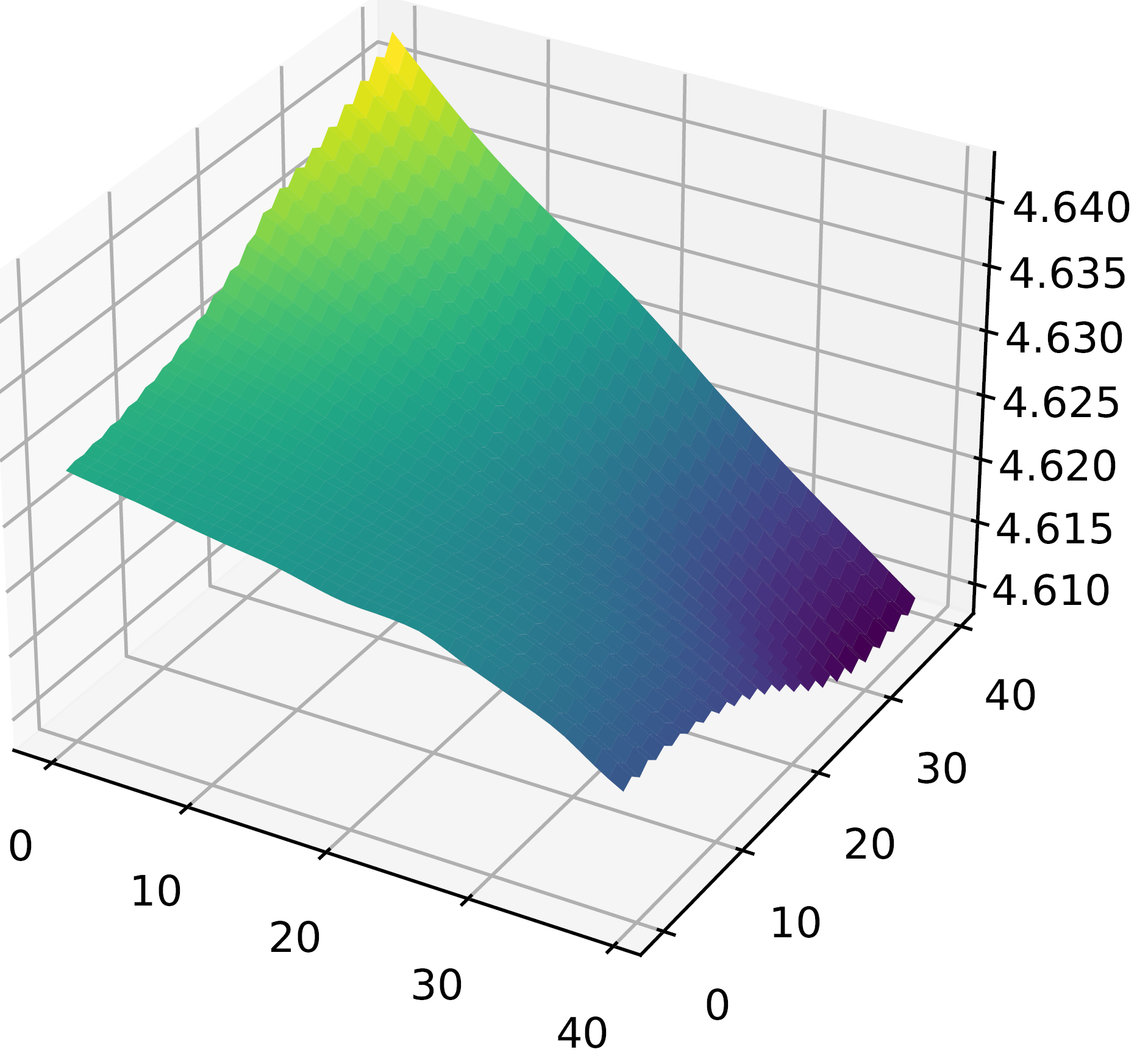} &
\includegraphics[width=0.25\textwidth]{figures/images/uclfinetune-t19.png} \vspace{-0.65in}\\
\begin{tabular}{c}
\textsc{Ucl-Si}\\{ Acc: $76.39\pm 1.59$}\\~\\~\\~\\~\\~\\~\\~\\~
\end{tabular}&
\includegraphics[width=0.25\textwidth]{figures/images/uclsi-t0.png} &
\includegraphics[width=0.25\textwidth]{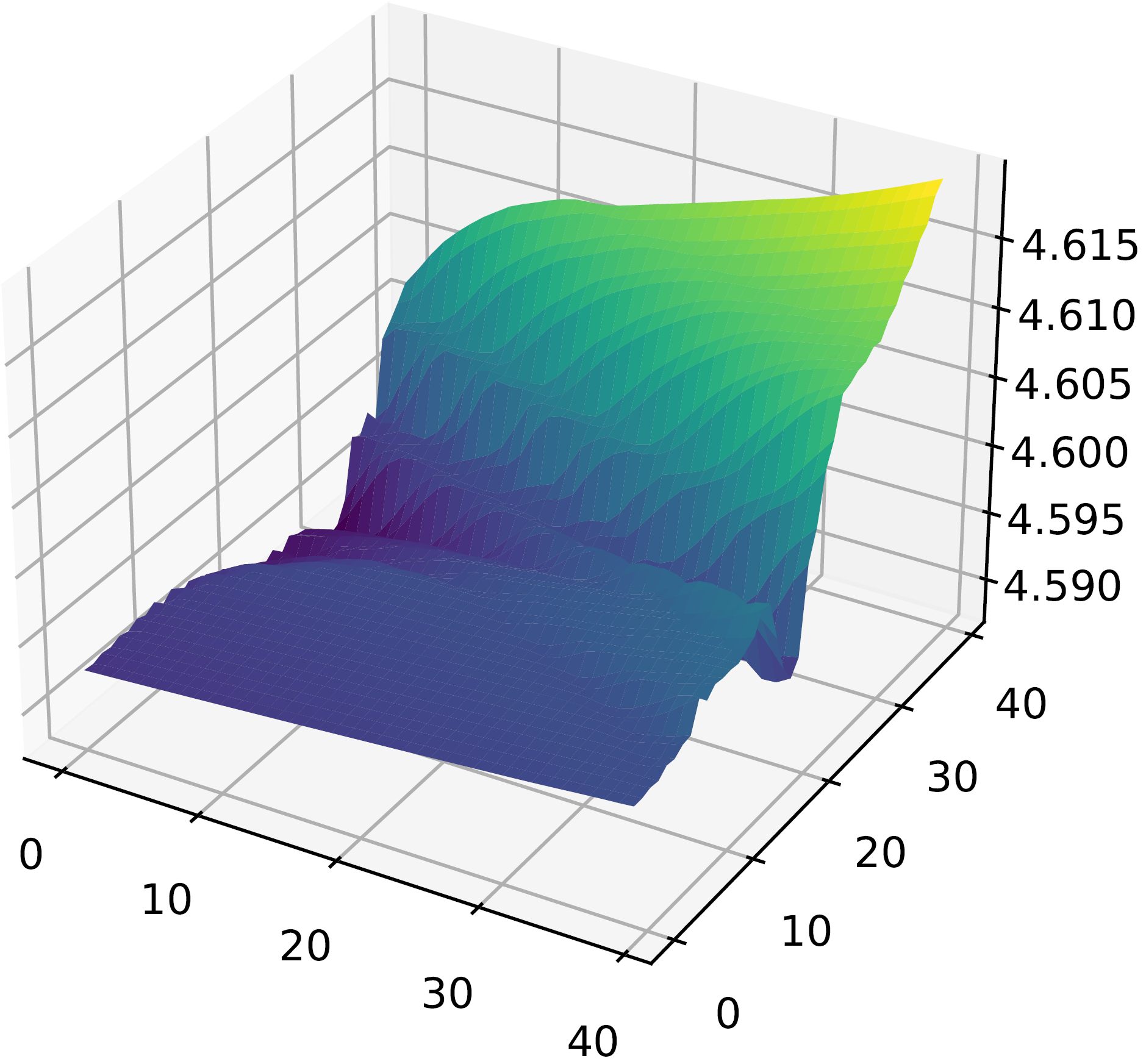} &
\includegraphics[width=0.25\textwidth]{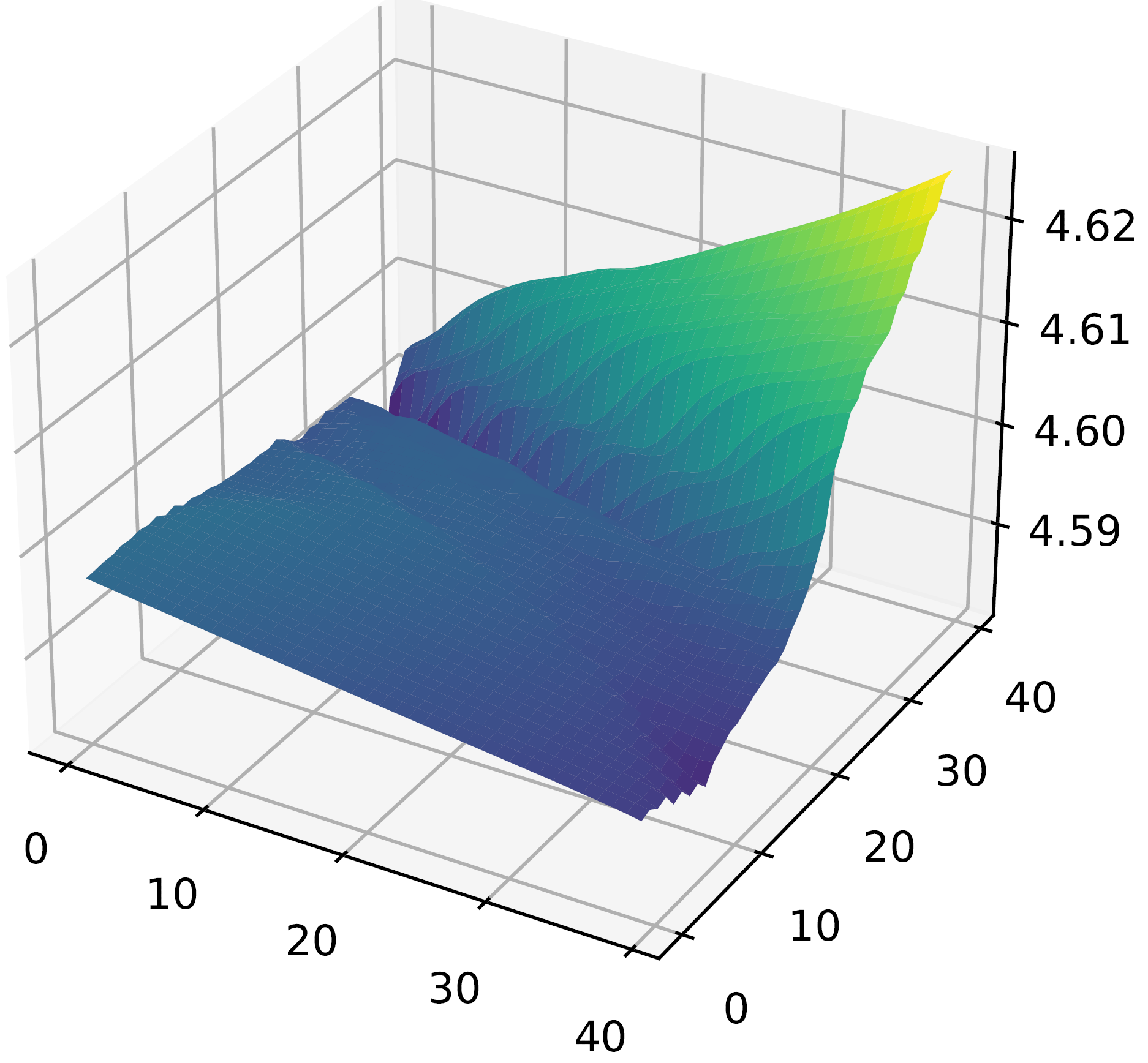} &
\includegraphics[width=0.25\textwidth]{figures/images/uclsi-t19.png} \vspace{-0.65in}\\
\begin{tabular}{c}
\textsc{Lump (Ours)}\\{ Acc: $76.60\pm 2.70$}\\~\\~\\~\\~\\~\\~\\~\\~
\end{tabular}&
\includegraphics[width=0.25\textwidth]{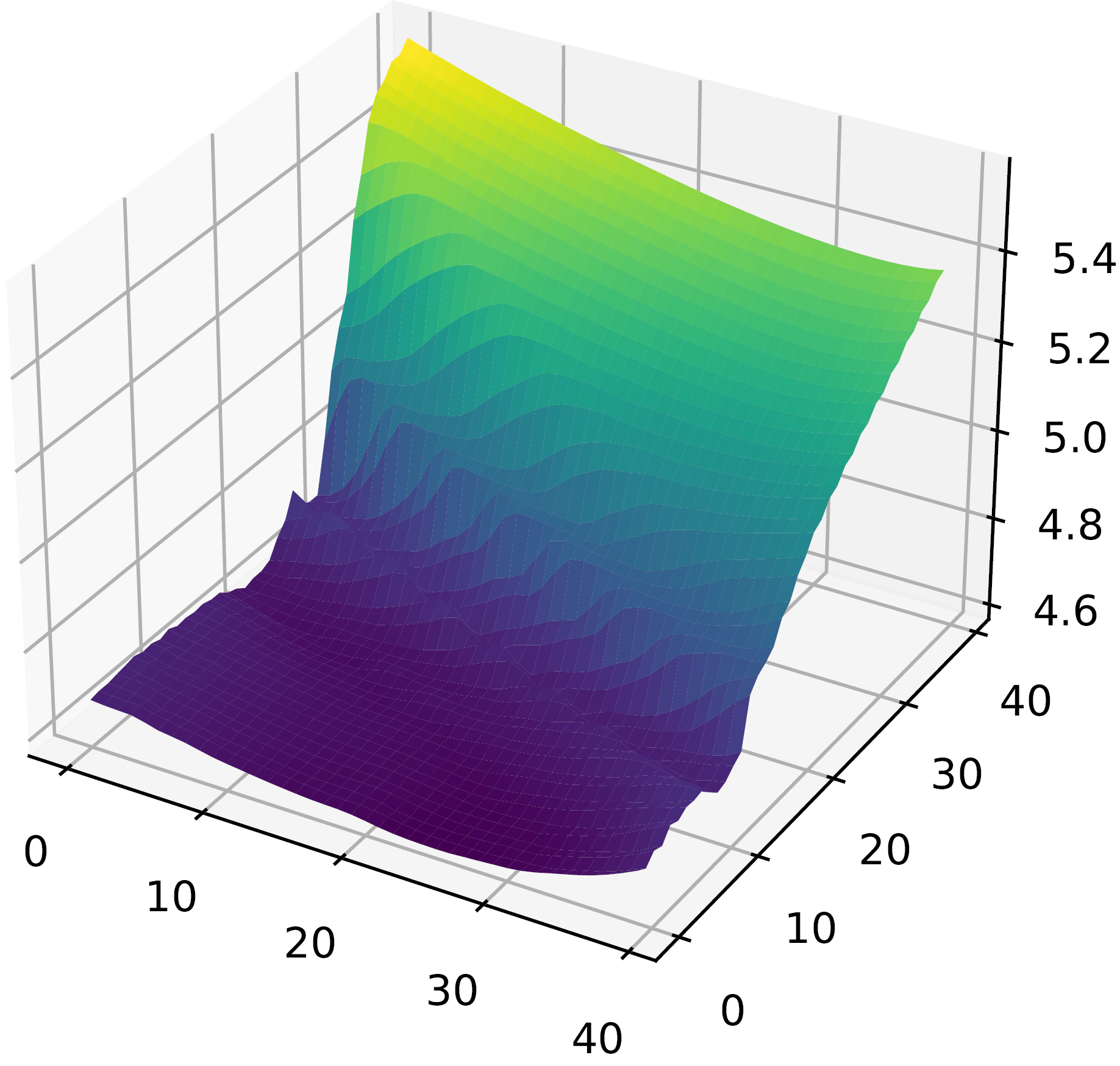} &
\includegraphics[width=0.25\textwidth]{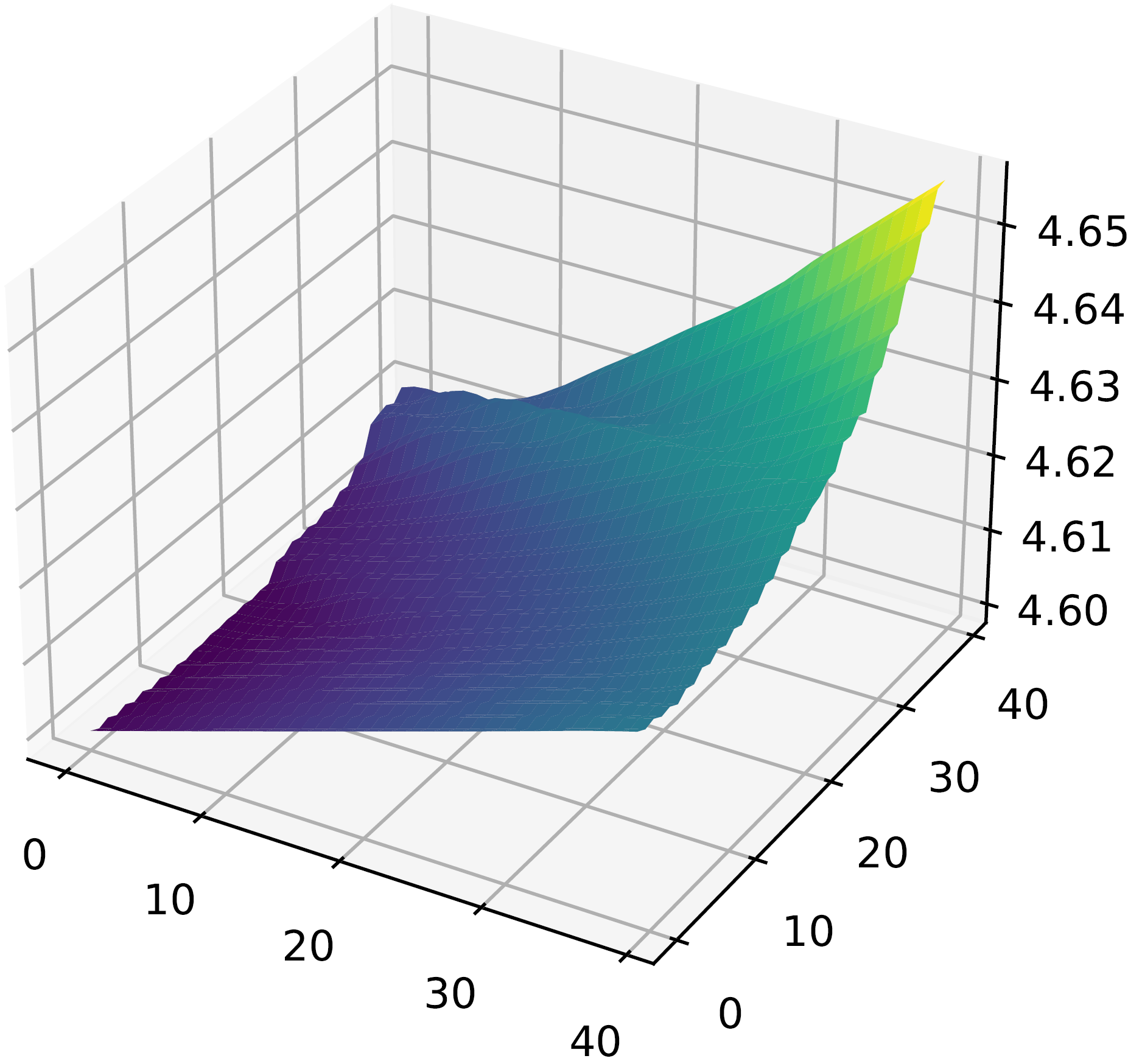} &
\includegraphics[width=0.25\textwidth]{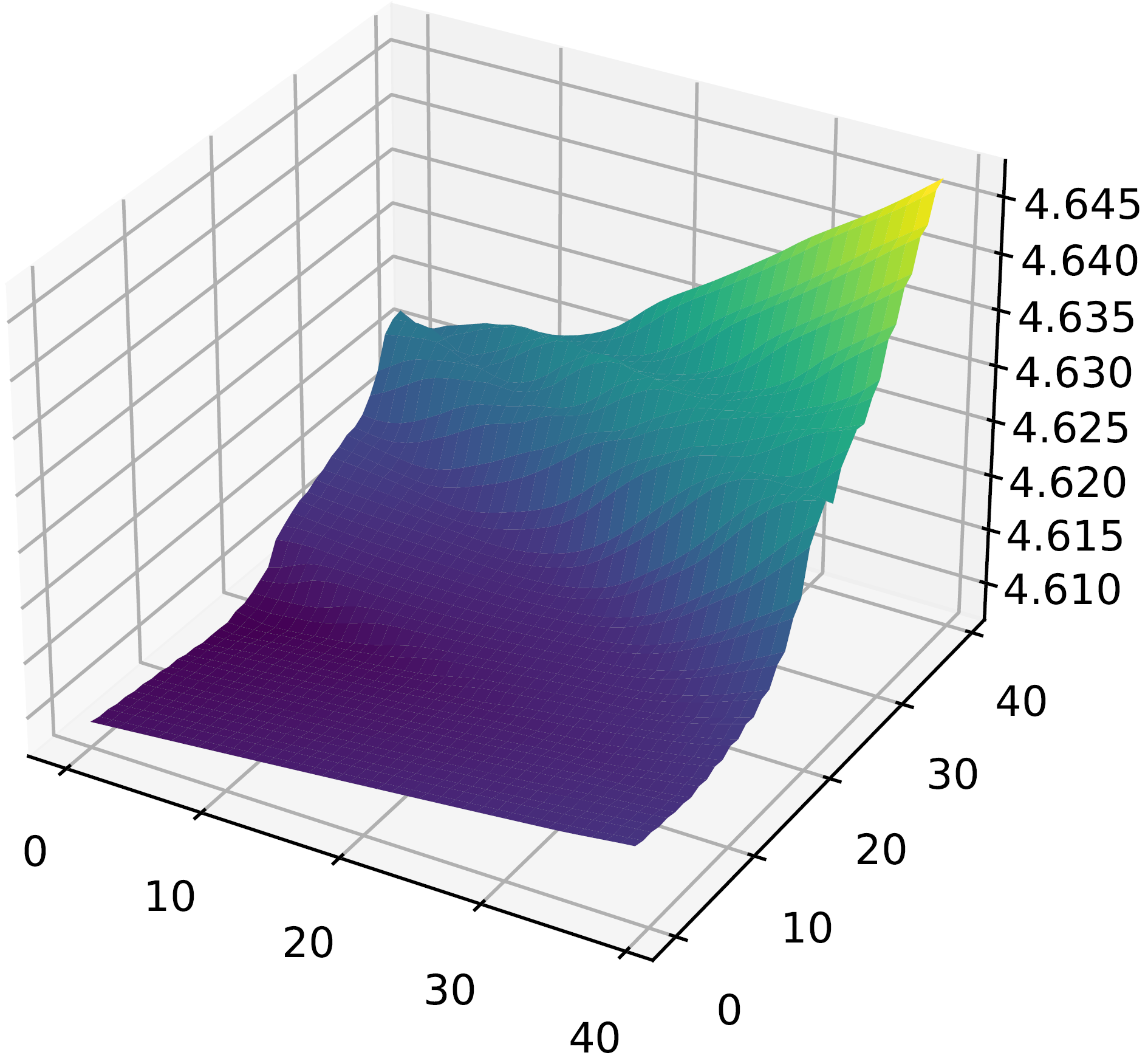} &
\includegraphics[width=0.25\textwidth]{figures/images/uclmixup-v2-t19.png} \vspace{-0.65in}\\
\end{tabular}}
\captionof{figure}{\small {\bf Loss landscape visualization} of $\mathcal{T}_0$ after the completion of training on task $\mathcal{T}_0$, $\mathcal{T}_{17}, \mathcal{T}_{18}$, and $\mathcal{T}_{19}$ for Split CIFAR-100 dataset on ResNet-18 architecture. We use Simsiam for UCL methods. \label{app:fig:loss_vis}}
\vspace{-0.15in}
\end{minipage}
\end{figure*}

\begin{figure*}[t!]
\begin{minipage}[t]{1.05\linewidth}
\resizebox{\linewidth}{!}{%
\begin{tabular}{c cccc}
\begin{tabular}{c}     
    \textsc{Raw Images}\\$\mathcal{T}_0,~\mathcal{T}_1,~\mathcal{T}_2,~\mathcal{T}_{13}$\\~\\~\\~\\~\\~\\~
\end{tabular}&
\includegraphics[width=0.20\textwidth]{figures/images/T0_imgid0_augmented_raw_cropped.png}&
\includegraphics[width=0.20\textwidth]{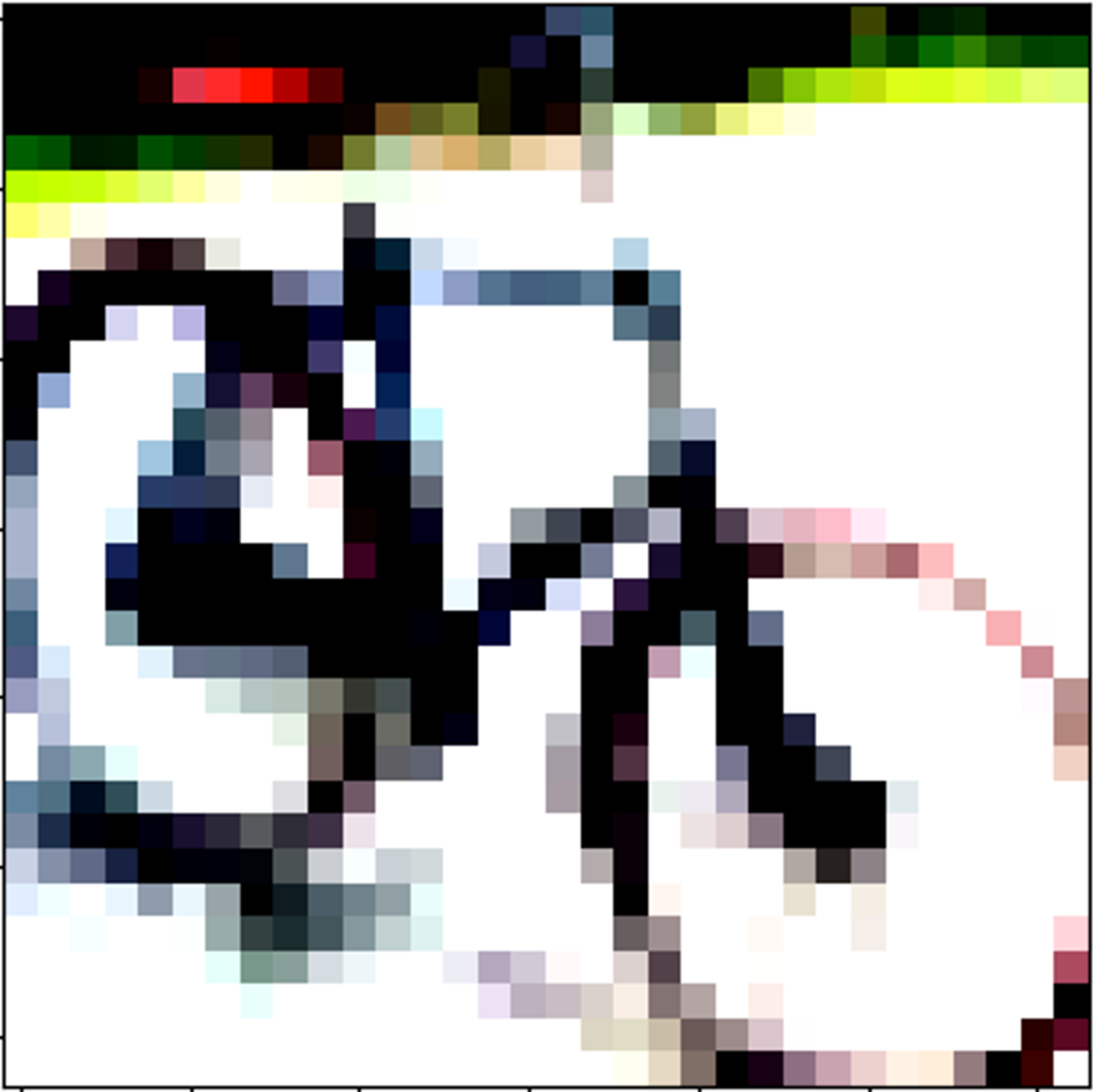}&
\includegraphics[width=0.20\textwidth]{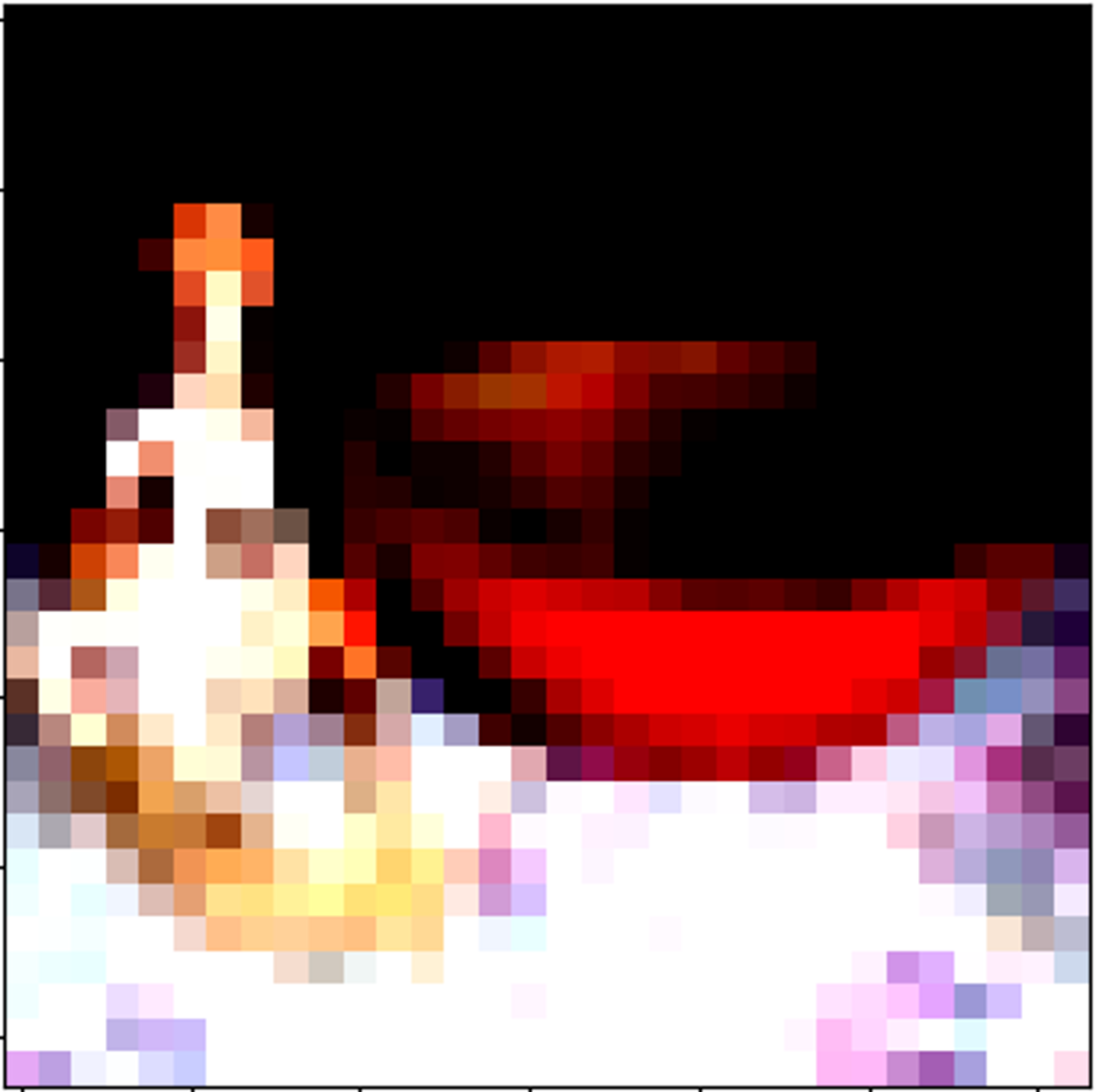}&
\includegraphics[width=0.20\textwidth]{figures/images/T13_imgid3_augmented_raw_cropped.PNG} \vspace{-0.35in}\\
\begin{tabular}{c}
\textsc{Scl-Finetune}\\{ Acc:$61.08\pm 0.04$}\\~\\~\\~\\~\\~\\~\\~\\~
\end{tabular}&
\includegraphics[width=0.25\textwidth]{figures/images/scl_finetune_3_c100__T0_imgid0_blockid1_atT19.png} &
\includegraphics[width=0.25\textwidth]{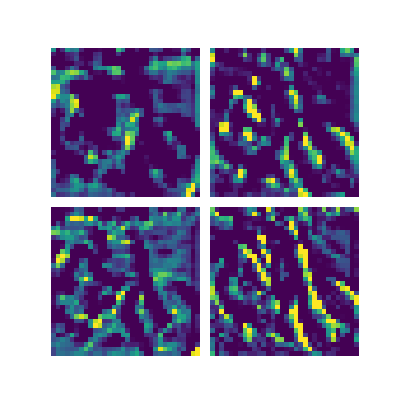} &
\includegraphics[width=0.25\textwidth]{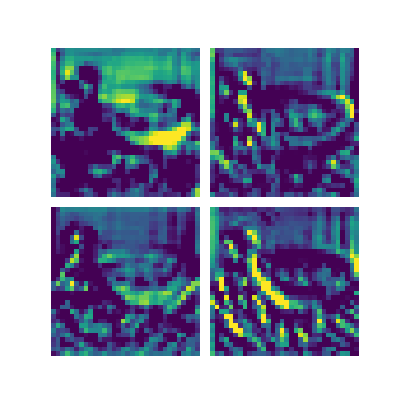} &
\includegraphics[width=0.25\textwidth]{figures/images/scl_finetune_3_c100__T13_imgid3_blockid1_atT19.png} \vspace{-0.85in}\\
\begin{tabular}{c}
\textsc{Scl-Si}\\{ Acc:$63.58\pm 0.37$}\\~\\~\\~\\~\\~\\~\\~\\~
\end{tabular}&
\includegraphics[width=0.25\textwidth]{figures/images/scl_cifar100_si_0p1__T0_imgid0_blockid1_atT19.png} &
\includegraphics[width=0.25\textwidth]{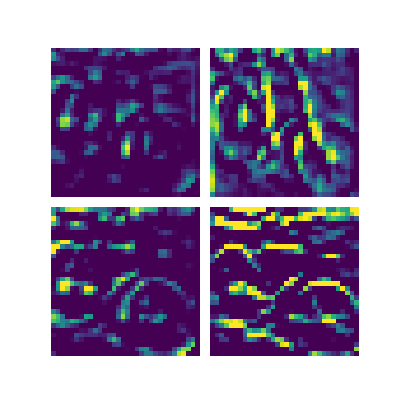} &
\includegraphics[width=0.25\textwidth]{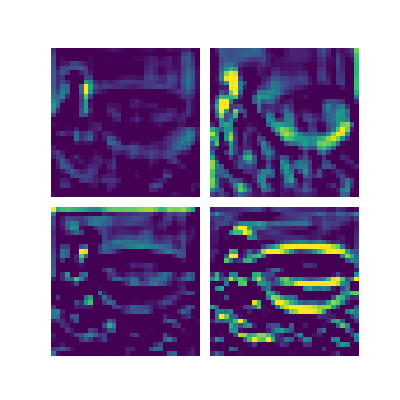} &
\includegraphics[width=0.25\textwidth]{figures/images/scl_cifar100_si_0p1__T13_imgid3_blockid1_atT19.png} \vspace{-0.85in}\\
\begin{tabular}{c}
\textsc{Scl-Gss}\\{ Acc:$70.78\pm 1.67$}\\~\\~\\~\\~\\~\\~\\~\\~
\end{tabular}&
\includegraphics[width=0.25\textwidth]{figures/images/gss_3__T0_imgid0_blockid1_atT19.png} &
\includegraphics[width=0.25\textwidth]{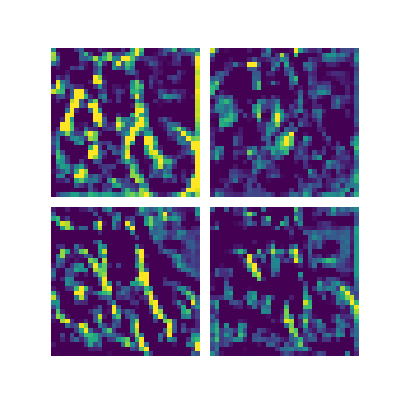} &
\includegraphics[width=0.25\textwidth]{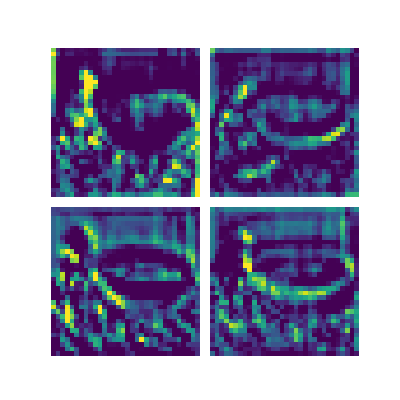} &
\includegraphics[width=0.25\textwidth]{figures/images/gss_3__T13_imgid3_blockid1_atT19.png} \vspace{-0.85in}\\
\begin{tabular}{c}
\textsc{Scl-Der}\\{ Acc:$79.52\pm 1.88$}\\~\\~\\~\\~\\~\\~\\~\\~
\end{tabular}&
\includegraphics[width=0.25\textwidth]{figures/images/der_3_c100__T0_imgid0_blockid1_atT19.png} &
\includegraphics[width=0.25\textwidth]{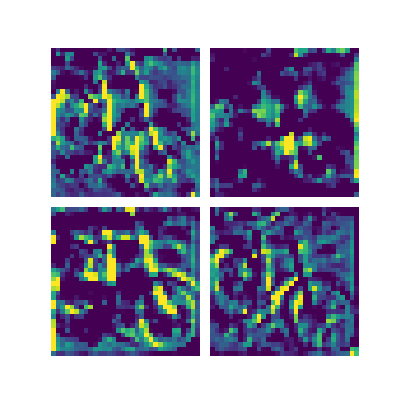} &
\includegraphics[width=0.25\textwidth]{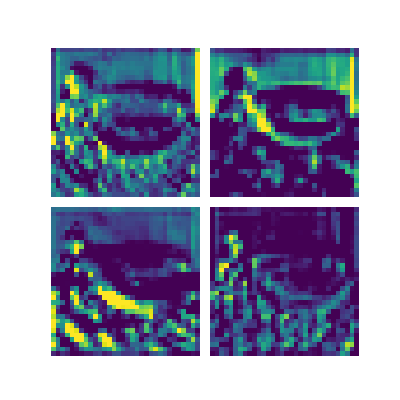} &
\includegraphics[width=0.25\textwidth]{figures/images/der_3_c100__T13_imgid3_blockid1_atT19.png} \vspace{-0.85in}\\
\begin{tabular}{c}
\textsc{Ucl-Finetune}\\{ Acc:$75.42\pm 0.78$}\\~\\~\\~\\~\\~\\~\\~\\~
\end{tabular}&
\includegraphics[width=0.25\textwidth]{figures/images/finetune_3_c100__T0_imgid0_blockid1_atT19.png} &
\includegraphics[width=0.25\textwidth]{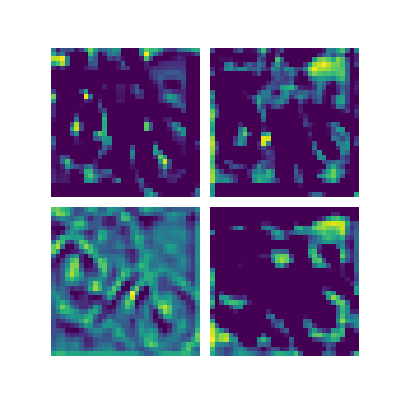} &
\includegraphics[width=0.25\textwidth]{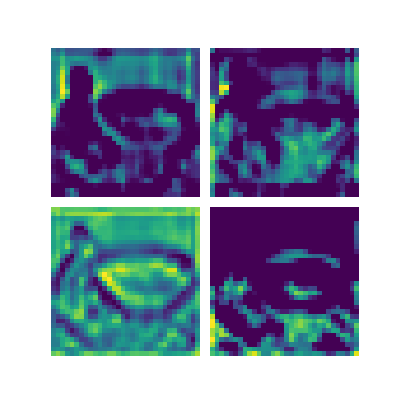} &
\includegraphics[width=0.25\textwidth]{figures/images/finetune_3_c100__T13_imgid3_blockid1_atT19.png} \vspace{-0.85in}\\
\begin{tabular}{c}
\textsc{Ucl-Si}\\{ Acc:$80.08\pm 1.30$}\\~\\~\\~\\~\\~\\~\\~\\~
\end{tabular}&\includegraphics[width=0.25\textwidth]{figures/images/ucl_cifar100_si_100p0__T0_imgid0_blockid1_atT19.png} &
\includegraphics[width=0.25\textwidth]{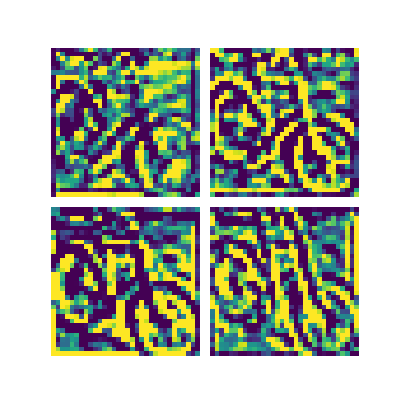} &
\includegraphics[width=0.25\textwidth]{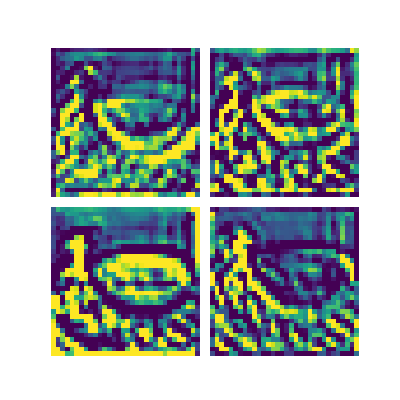} &
\includegraphics[width=0.25\textwidth]{figures/images/ucl_cifar100_si_100p0__T13_imgid3_blockid1_atT19.png} \vspace{-0.85in}\\
\begin{tabular}{c}
\textsc{Lump (Ours)}\\{ Acc:$82.30\pm 1.35$}\\~\\~\\~\\~\\~\\~\\~\\~
\end{tabular}&
\includegraphics[width=0.25\textwidth]{figures/images/ucl_cifar100_mixup_1p0__T0_imgid0_blockid1_atT19.png} &
\includegraphics[width=0.25\textwidth]{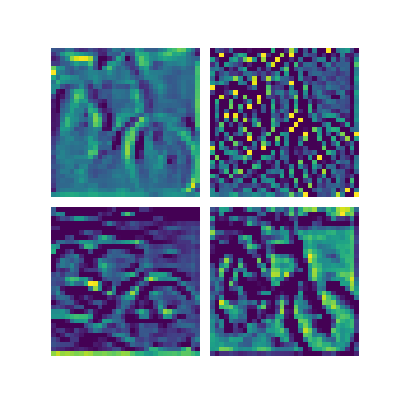} &
\includegraphics[width=0.25\textwidth]{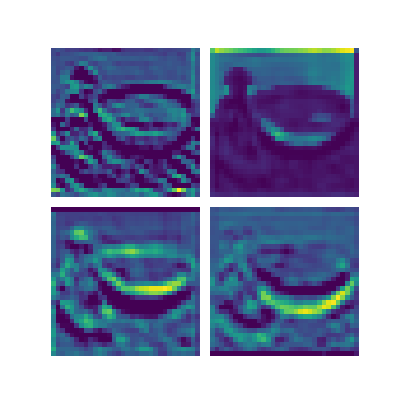} &
\includegraphics[width=0.25\textwidth]{figures/images/ucl_cifar100_mixup_1p0__T13_imgid3_blockid1_atT19.png} \vspace{-0.85in}\\
\end{tabular}}
\captionof{figure}{\small {\bf Visualization of feature maps} for the second block representations learnt by SCL and UCL strategies (with Simsiam) for Resnet-18 architecture after the completion of continual learning for Split CIFAR-100 dataset ($n=20$). The accuracy is the mean across three runs for the corresponding task. \label{fig:appendix:feature_maps_a7}}
\vspace{-0.15in}
\end{minipage}
\end{figure*}

\begin{figure*}[t!]
\begin{minipage}[t]{1.05\linewidth}
\resizebox{\linewidth}{!}{%
\begin{tabular}{c cccc}
\begin{tabular}{c}     
    \textsc{Raw Images}\\$\mathcal{T}_0,~\mathcal{T}_1,~\mathcal{T}_2,~\mathcal{T}_{13}$\\~\\~\\~\\~\\~\\~
\end{tabular}&
\includegraphics[width=0.20\textwidth]{figures/images/T0_imgid0_augmented_raw_cropped.png}&
\includegraphics[width=0.20\textwidth]{figures/images/T1_imgid0_augmented_raw_cropped.png}&
\includegraphics[width=0.20\textwidth]{figures/images/T2_imgid5_raw_cropped.PNG}&
\includegraphics[width=0.20\textwidth]{figures/images/T13_imgid3_augmented_raw_cropped.PNG} \vspace{-0.35in}\\
\begin{tabular}{c}
\textsc{Scl-Finetune}\\{ Acc:$61.08\pm 0.04$}\\~\\~\\~\\~\\~\\~\\~\\~
\end{tabular}&
\includegraphics[width=0.25\textwidth]{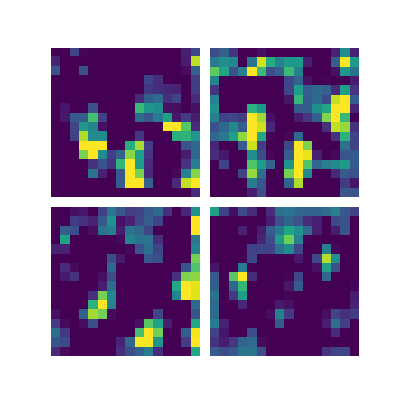} &
\includegraphics[width=0.25\textwidth]{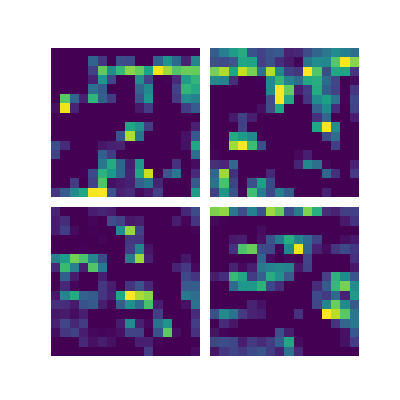} &
\includegraphics[width=0.25\textwidth]{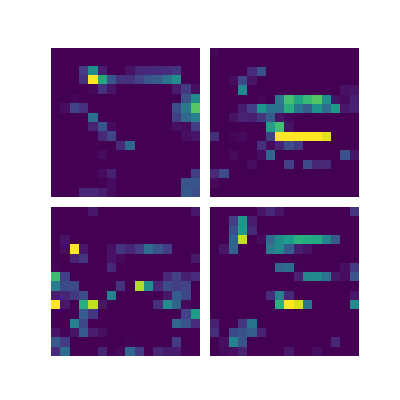} &
\includegraphics[width=0.25\textwidth]{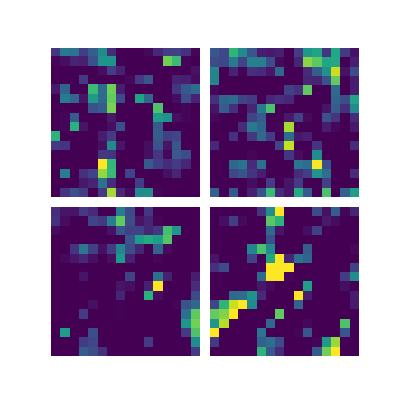}  \vspace{-0.85in}\\
\begin{tabular}{c}
\textsc{Scl-Si}\\{ Acc:$63.58\pm 0.37$}\\~\\~\\~\\~\\~\\~\\~\\~
\end{tabular}&
\includegraphics[width=0.25\textwidth]{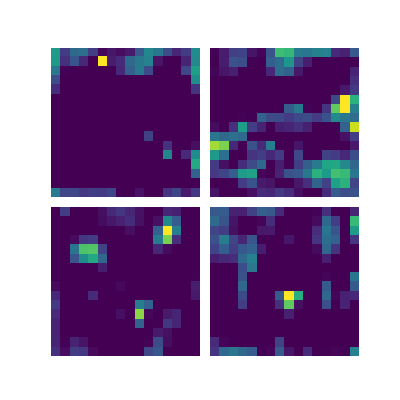} &
\includegraphics[width=0.25\textwidth]{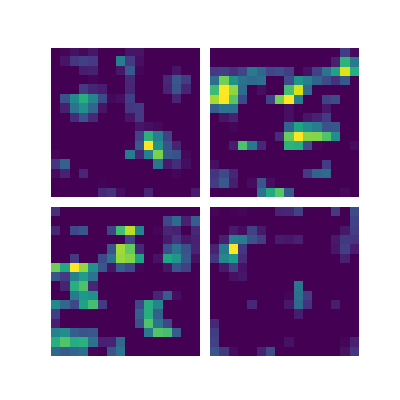} &
\includegraphics[width=0.25\textwidth]{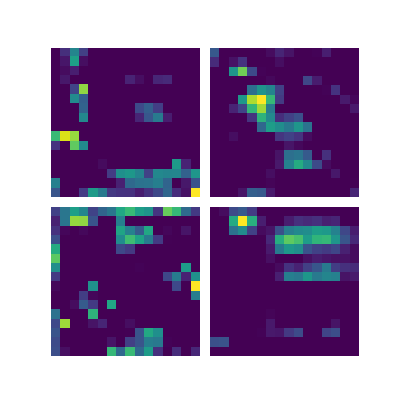} &
\includegraphics[width=0.25\textwidth]{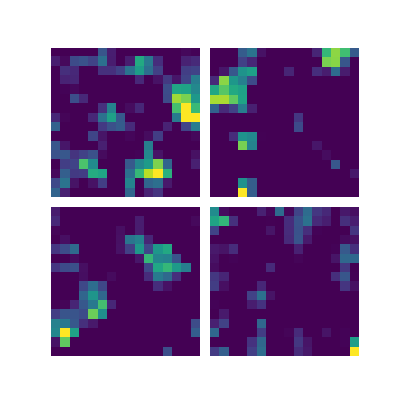}  \vspace{-0.85in}\\
\begin{tabular}{c}
\textsc{Scl-Gss}\\{ Acc:$70.78\pm 1.67$}\\~\\~\\~\\~\\~\\~\\~\\~
\end{tabular}&
\includegraphics[width=0.25\textwidth]{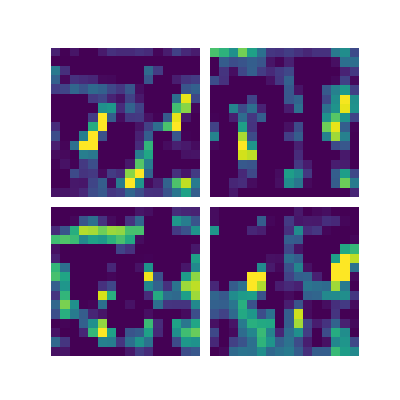} &
\includegraphics[width=0.25\textwidth]{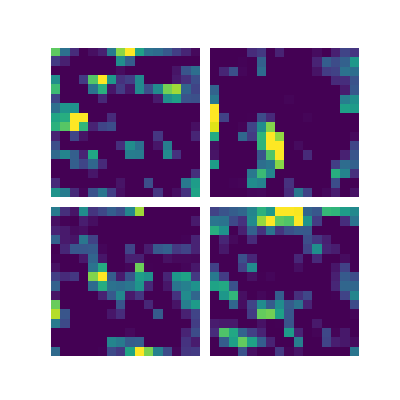} &
\includegraphics[width=0.25\textwidth]{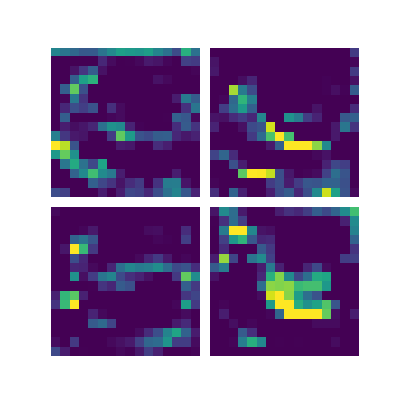} &
\includegraphics[width=0.25\textwidth]{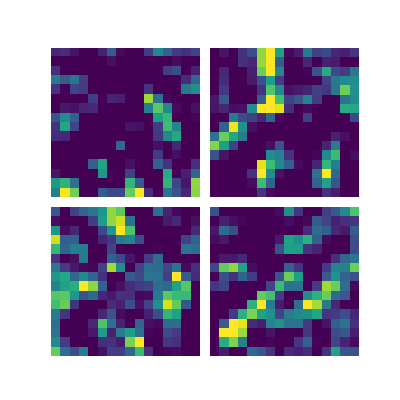}\vspace{-0.85in}\\
\begin{tabular}{c}
\textsc{Scl-Der}\\{ Acc:$79.52\pm 1.88$}\\~\\~\\~\\~\\~\\~\\~\\~
\end{tabular}&
\includegraphics[width=0.25\textwidth]{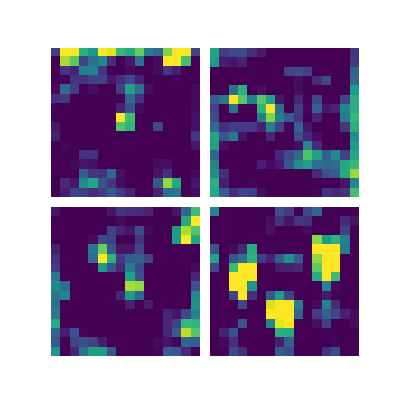} &
\includegraphics[width=0.25\textwidth]{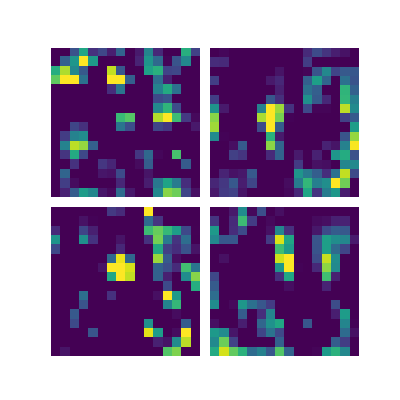} &
\includegraphics[width=0.25\textwidth]{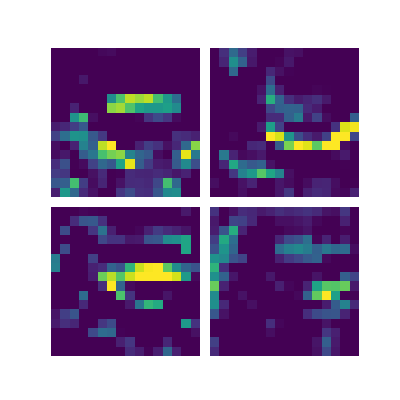} &
\includegraphics[width=0.25\textwidth]{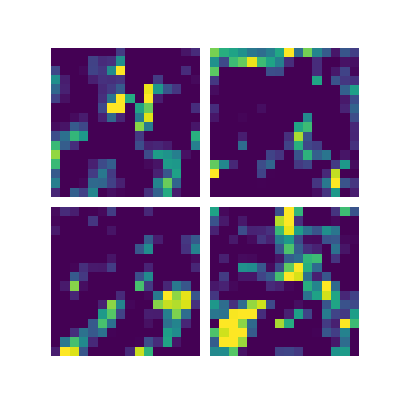}  \vspace{-0.85in}\\
\begin{tabular}{c}
\textsc{Ucl-Finetune}\\{ Acc:$75.42\pm 0.78$}\\~\\~\\~\\~\\~\\~\\~\\~
\end{tabular}&
\includegraphics[width=0.25\textwidth]{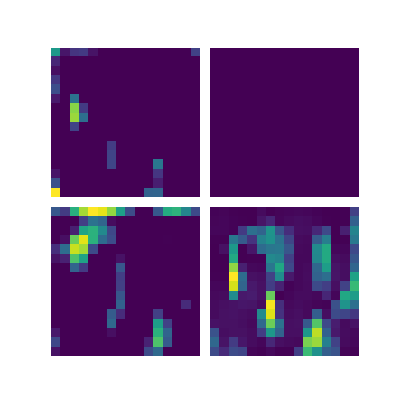} &
\includegraphics[width=0.25\textwidth]{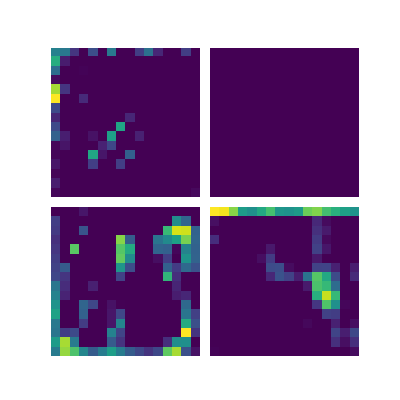} &
\includegraphics[width=0.25\textwidth]{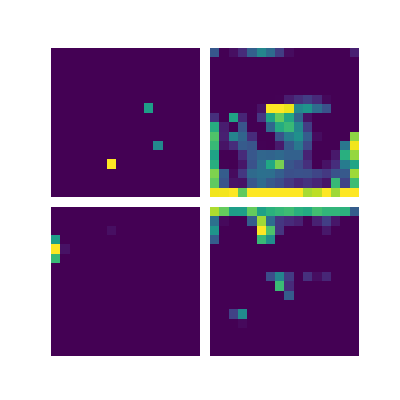} &
\includegraphics[width=0.25\textwidth]{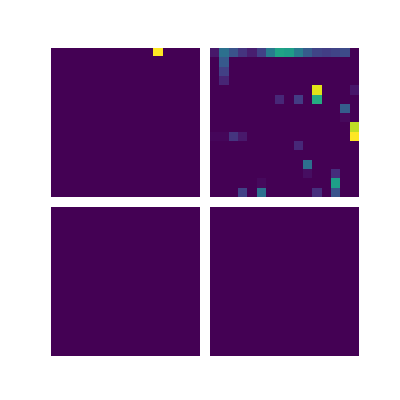} \vspace{-0.85in}\\
\begin{tabular}{c}
\textsc{Ucl-Si}\\{ Acc:$80.08\pm 1.30$}\\~\\~\\~\\~\\~\\~\\~\\~
\end{tabular}&\includegraphics[width=0.25\textwidth]{figures/rebuttal/unsup_finetune_3_c100__T0_imgid0_atT19.png} &
\includegraphics[width=0.25\textwidth]{figures/rebuttal/unsup_finetune_3_c100__T1_imgid0_atT19.png} &
\includegraphics[width=0.25\textwidth]{figures/rebuttal/unsup_finetune_3_c100__T2_imgid5_atT19.png} &
\includegraphics[width=0.25\textwidth]{figures/rebuttal/unsup_finetune_3_c100__T13_imgid3_atT19.png} \vspace{-0.85in}\\
\begin{tabular}{c}
\textsc{Lump (Ours)}\\{ Acc:$82.30\pm 1.35$}\\~\\~\\~\\~\\~\\~\\~\\~
\end{tabular}&
\includegraphics[width=0.25\textwidth]{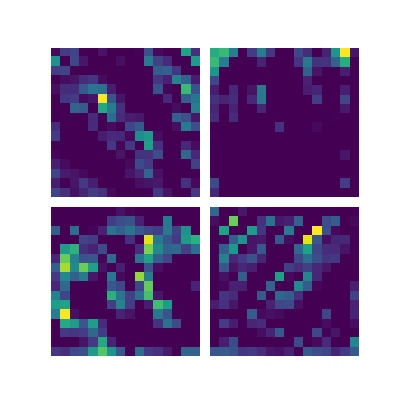} &
\includegraphics[width=0.25\textwidth]{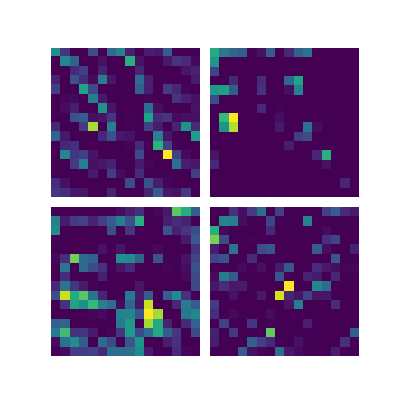} &
\includegraphics[width=0.25\textwidth]{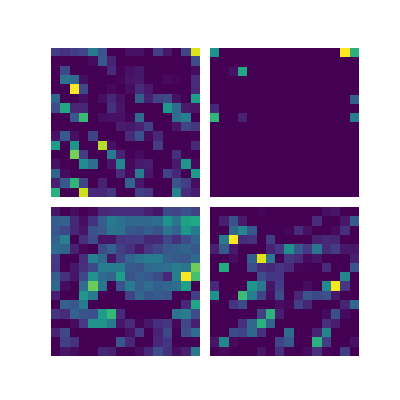} &
\includegraphics[width=0.25\textwidth]{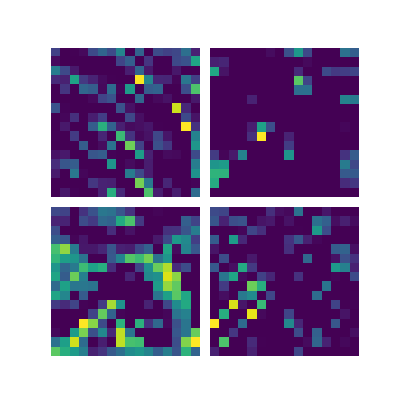} \vspace{-0.85in}\\
\end{tabular}}
\captionof{figure}{\small {\bf Visualization of feature maps} for the third block representations learnt by SCL and UCL strategies (with Simsiam) for Resnet-18 architecture after the completion of continual learning for Split CIFAR-100 dataset ($n=20$). The accuracy is the mean across three runs for the corresponding task. \label{fig:appendix:feature_maps}}
\vspace{-0.15in}
\end{minipage}
\label{fig:appendix:visualize}
\end{figure*}

\begin{figure*}[t!]
\begin{minipage}[t]{1.05\linewidth}
\resizebox{\linewidth}{!}{%
\begin{tabular}{c cccc}
\begin{tabular}{c}     
\textsc{Raw Images}\\$\mathcal{T}_0,~\mathcal{T}_1,~\mathcal{T}_2,~\mathcal{T}_{10}$\\~\\~\\~\\~\\~\\~
\end{tabular}&
\includegraphics[width=0.20\textwidth]{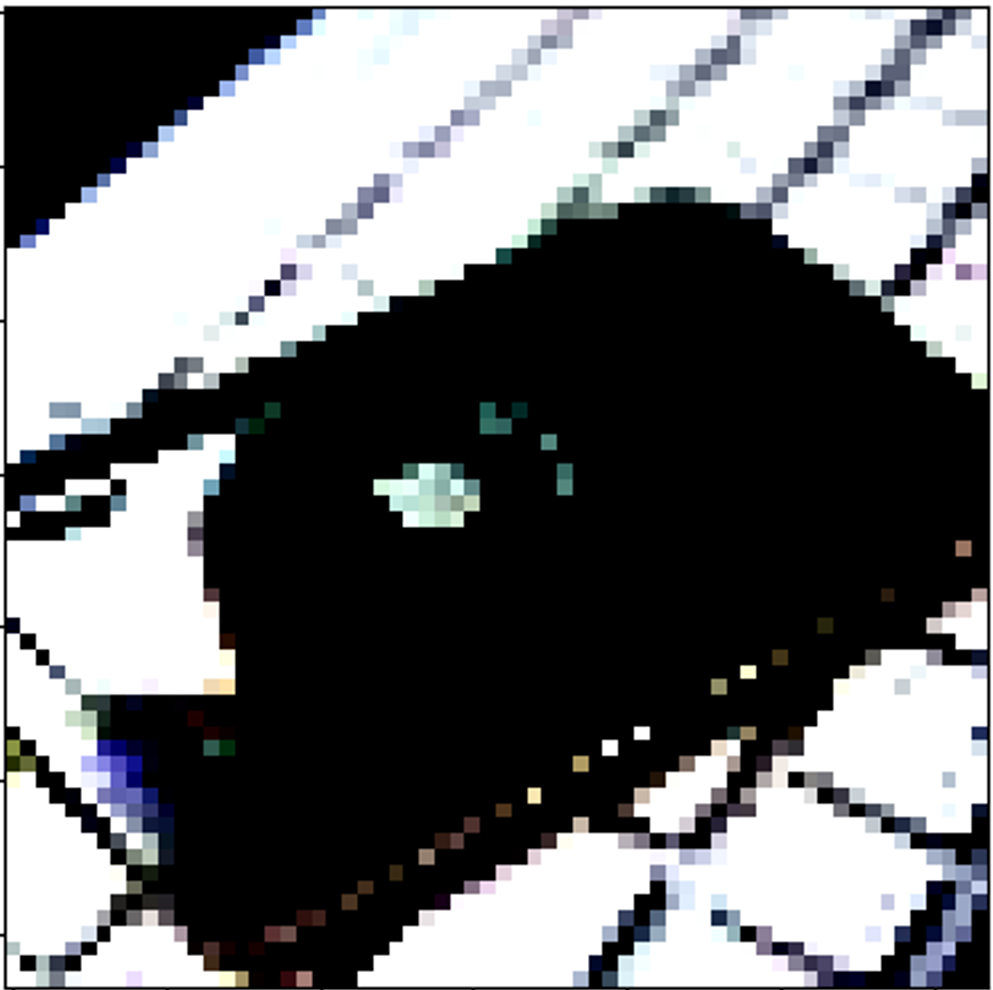}&
\includegraphics[width=0.20\textwidth]{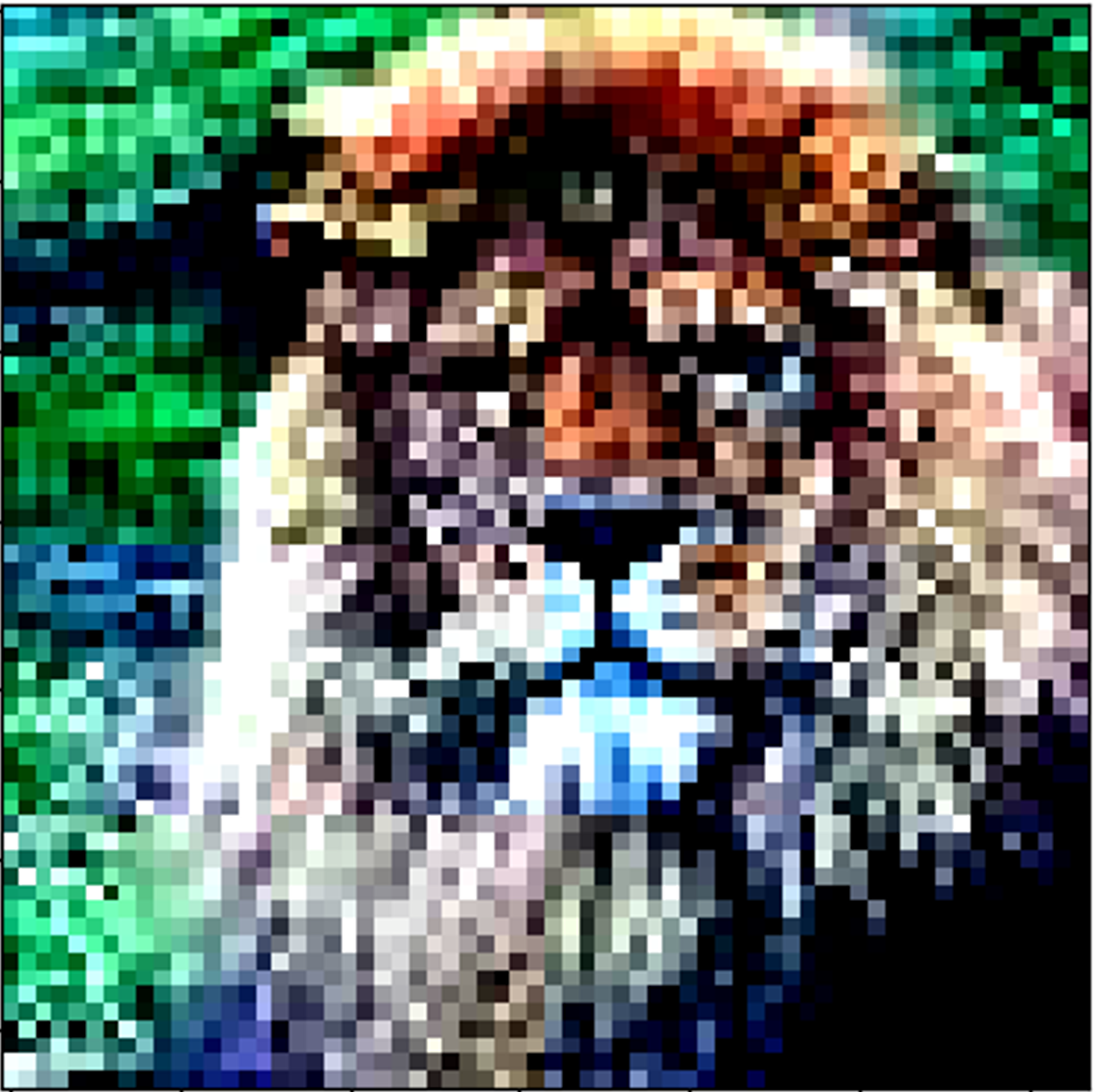}&
\includegraphics[width=0.20\textwidth]{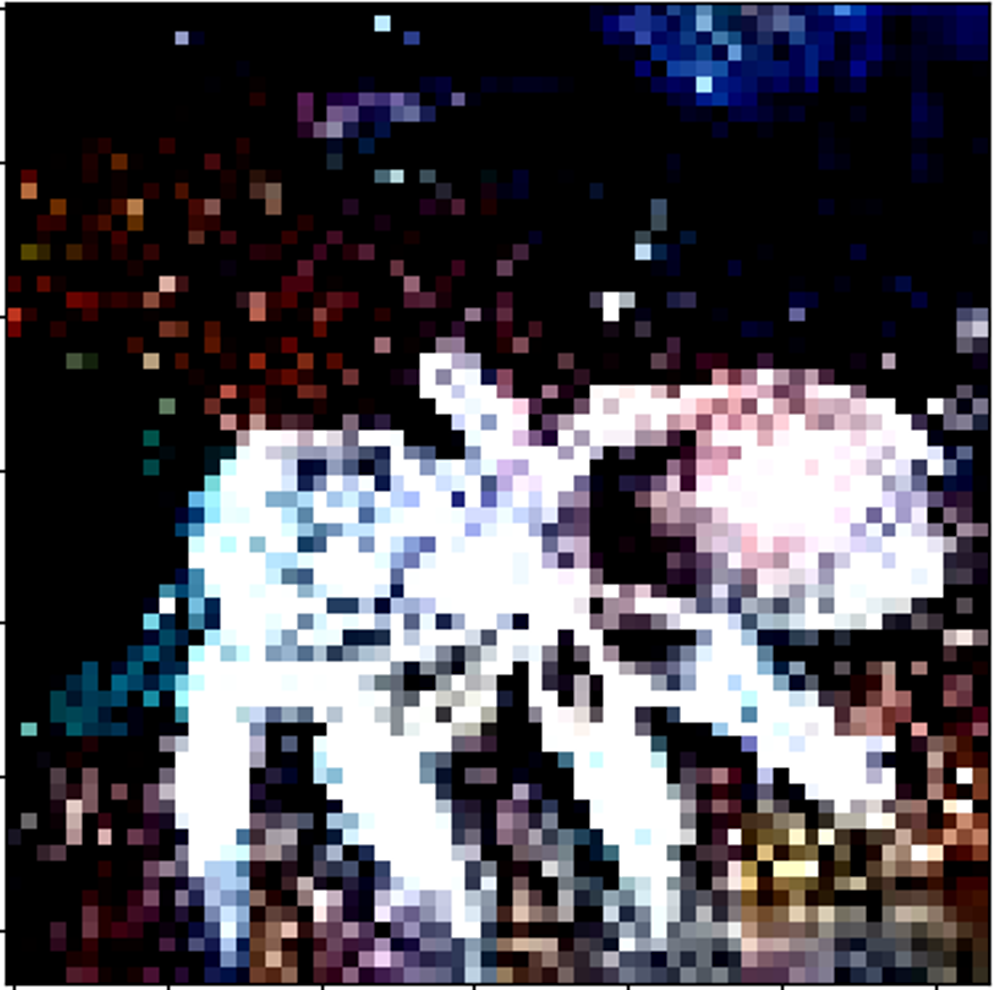}&
\includegraphics[width=0.20\textwidth]{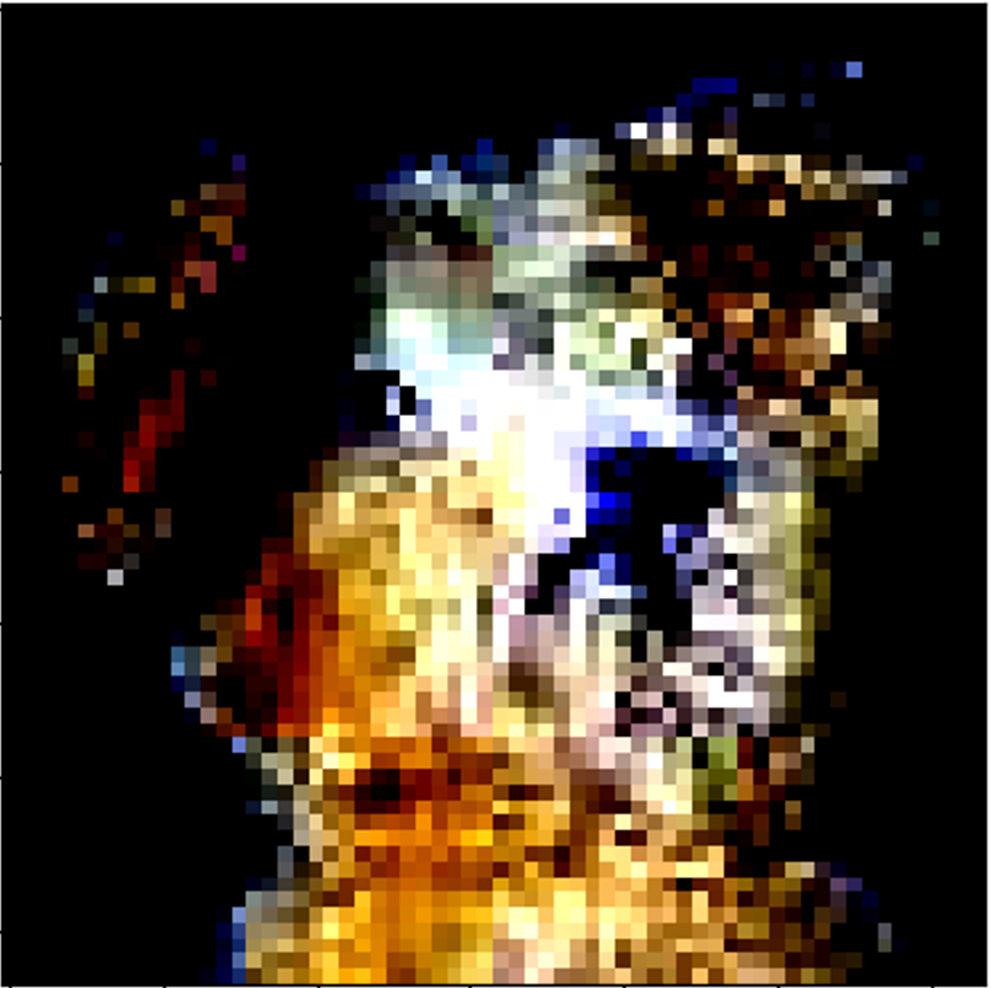} \vspace{-0.35in}\\
\begin{tabular}{c}
\textsc{Scl-Finetune}\\{ Acc:$53.10\pm 1.37$}\\~\\~\\~\\~\\~\\~\\~\\~
\end{tabular}&
\includegraphics[width=0.25\textwidth]{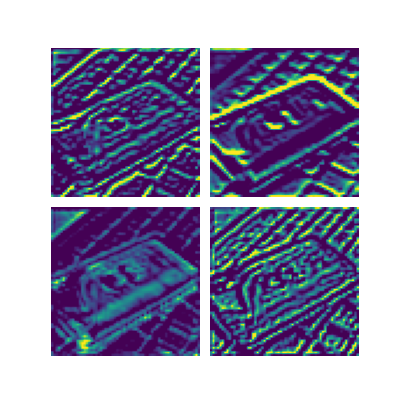} &
\includegraphics[width=0.25\textwidth]{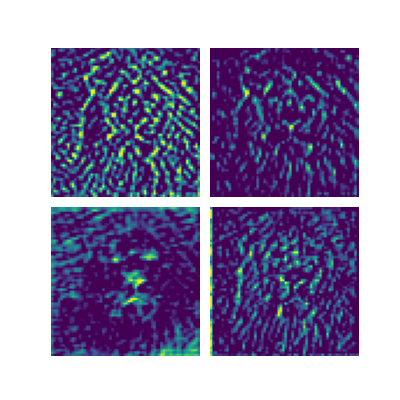} &
\includegraphics[width=0.25\textwidth]{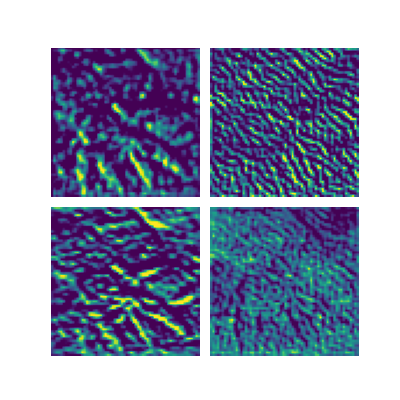} &
\includegraphics[width=0.25\textwidth]{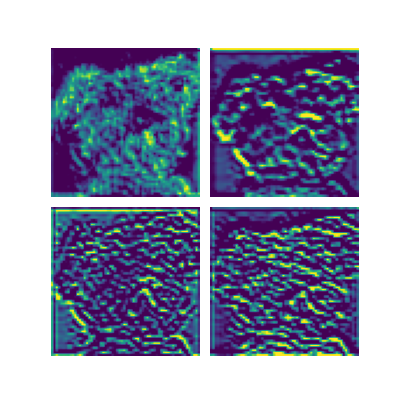} \vspace{-0.85in}\\
\begin{tabular}{c}
\textsc{Scl-Si}\\{ Acc:$44.96\pm 2.41$}\\~\\~\\~\\~\\~\\~\\~\\~
\end{tabular}&
\includegraphics[width=0.25\textwidth]{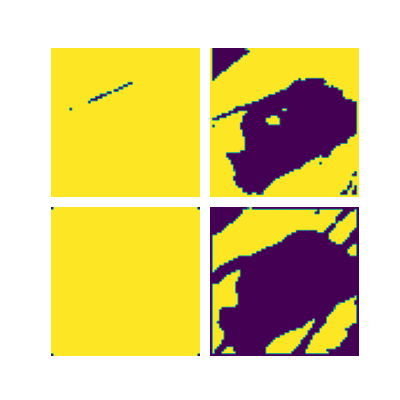} &
\includegraphics[width=0.25\textwidth]{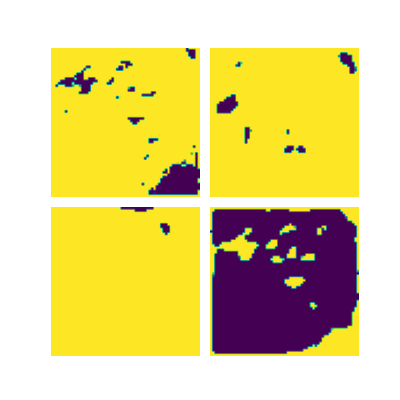} &
\includegraphics[width=0.25\textwidth]{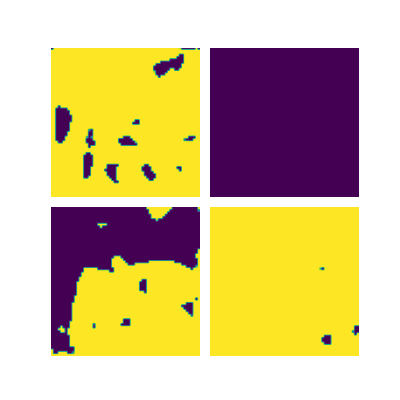} &
\includegraphics[width=0.25\textwidth]{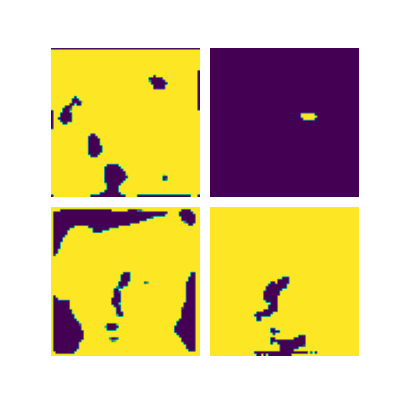} \vspace{-0.85in}\\
\begin{tabular}{c}
\textsc{Scl-Gss}\\{ Acc:$70.96\pm 0.72$}\\~\\~\\~\\~\\~\\~\\~\\~
\end{tabular}&
\includegraphics[width=0.25\textwidth]{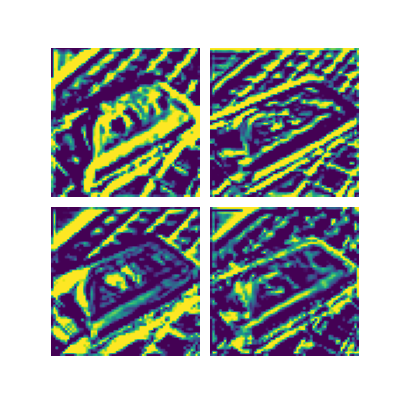} &
\includegraphics[width=0.25\textwidth]{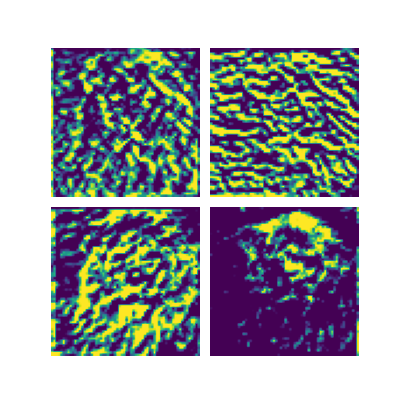} &
\includegraphics[width=0.25\textwidth]{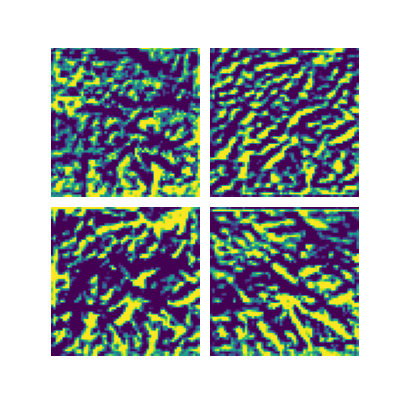} &
\includegraphics[width=0.25\textwidth]{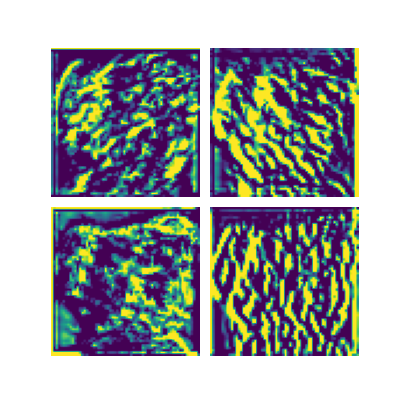} \vspace{-0.85in}\\
\begin{tabular}{c}
\textsc{Scl-Der}\\{ Acc:$68.03\pm 0.85$}\\~\\~\\~\\~\\~\\~\\~\\~
\end{tabular}&
\includegraphics[width=0.25\textwidth]{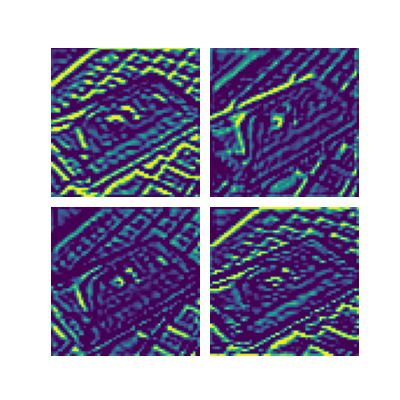} &
\includegraphics[width=0.25\textwidth]{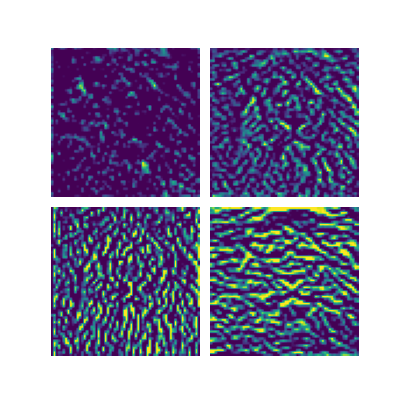} &
\includegraphics[width=0.25\textwidth]{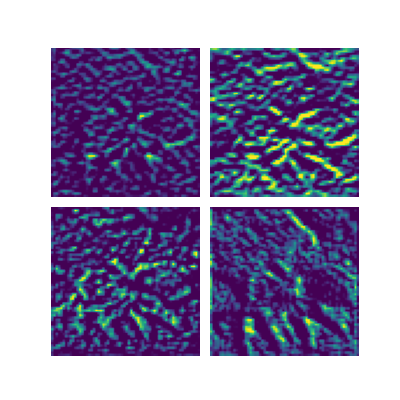} &
\includegraphics[width=0.25\textwidth]{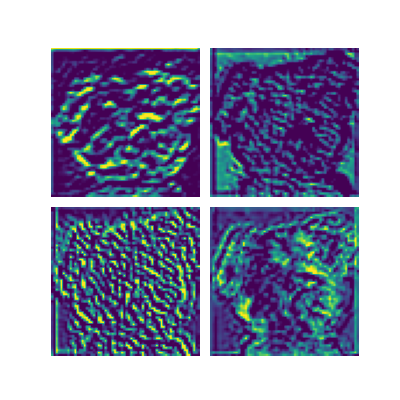} \vspace{-0.85in}\\
\begin{tabular}{c}
\textsc{Ucl-Finetune}\\{ Acc:$71.07\pm 0.20$}\\~\\~\\~\\~\\~\\~\\~\\~
\end{tabular}&
\includegraphics[width=0.25\textwidth]{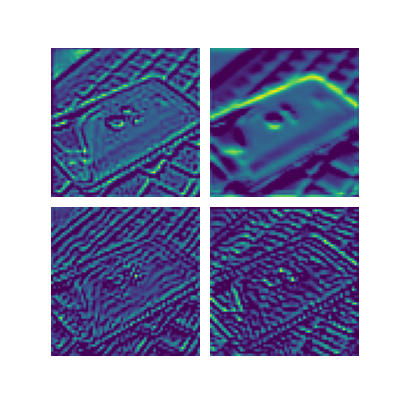} &
\includegraphics[width=0.25\textwidth]{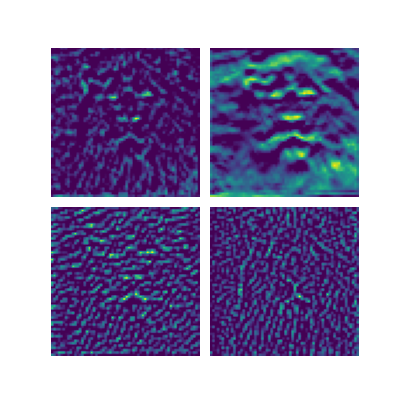} &
\includegraphics[width=0.25\textwidth]{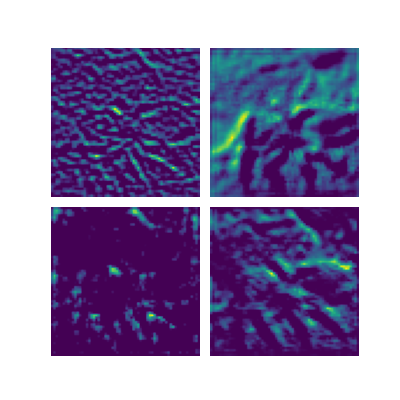} &
\includegraphics[width=0.25\textwidth]{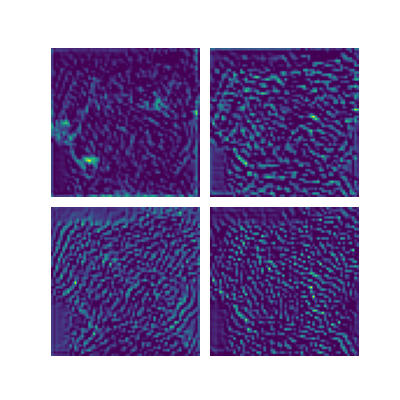} \vspace{-0.85in}\\
\begin{tabular}{c}
\textsc{Ucl-Si}\\{ Acc:$72.34\pm 0.42$}\\~\\~\\~\\~\\~\\~\\~\\~
\end{tabular}&\includegraphics[width=0.25\textwidth]{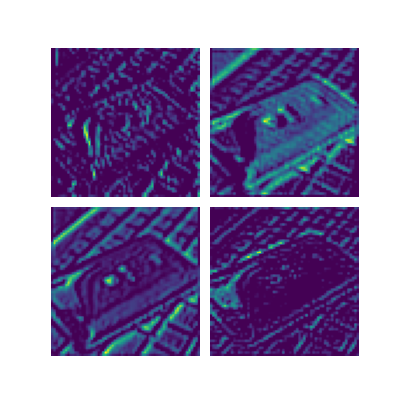} &
\includegraphics[width=0.25\textwidth]{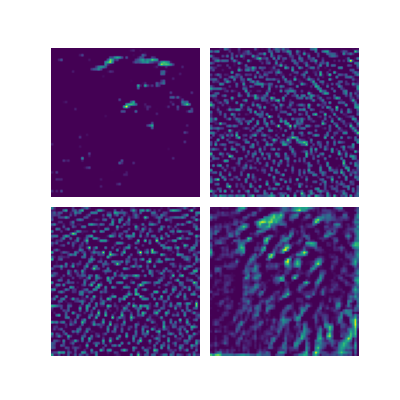} &
\includegraphics[width=0.25\textwidth]{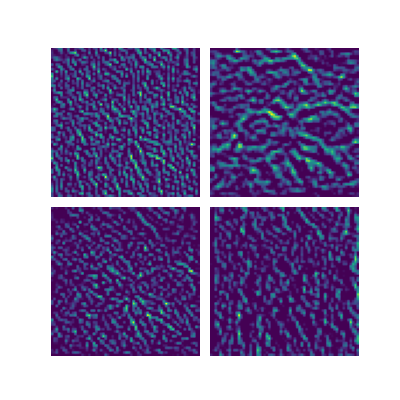} &
\includegraphics[width=0.25\textwidth]{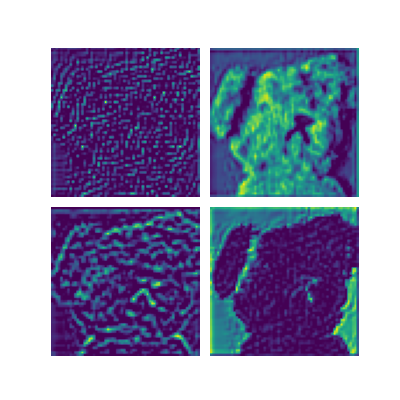} \vspace{-0.85in}\\
\begin{tabular}{c}
\textsc{Lump (Ours)}\\{ Acc:$76.66\pm 2.39$}\\~\\~\\~\\~\\~\\~\\~\\~
\end{tabular}&
\includegraphics[width=0.25\textwidth]{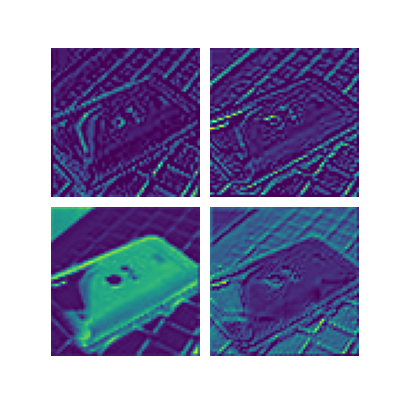} &
\includegraphics[width=0.25\textwidth]{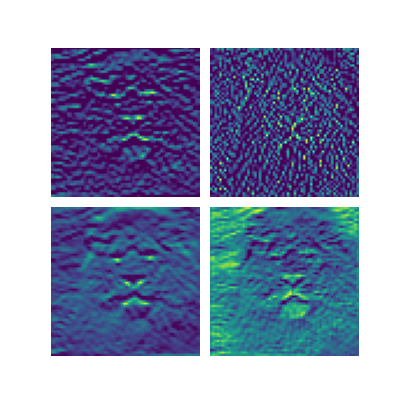} &
\includegraphics[width=0.25\textwidth]{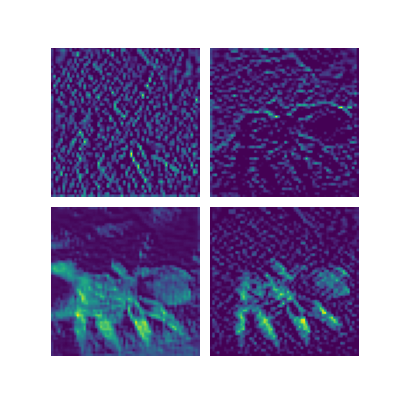} &
\includegraphics[width=0.25\textwidth]{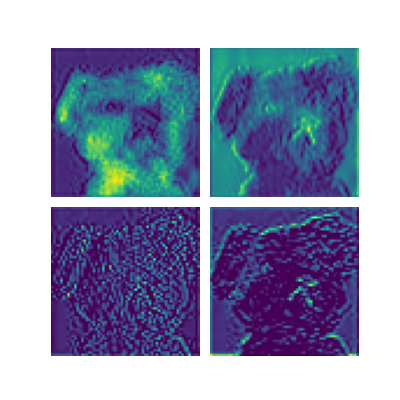} \vspace{-0.85in}\\
\end{tabular}}
\captionof{figure}{\small {\bf Visualization of feature maps} for the second block representations learnt by SCL and UCL strategies (with Simsiam) for Resnet-18 architecture after the completion of continual learning for Split Tiny-ImageNet dataset ($n=20$). The accuracy is the mean across three runs for the corresponding task. \label{fig:appendix:feature_maps_tiny}}
\vspace{-0.15in}
\end{minipage}
\end{figure*}

We provide additional loss landscape on Split CIFAR-100 in \cref{app:fig:loss_vis} and \cref{fig:appendix:feature_maps_a7}, \cref{fig:appendix:feature_maps} show the second and third block feature visualizations on Split CIFAR-100 respectively. \cref{fig:appendix:feature_maps_tiny} shows the feature visualizations for Split Tiny-ImageNet on ResNet-18 architecture.


\end{document}